\documentclass[twocolumn]{article}

\usepackage{subcaption}
\usepackage{multirow}

\title{Two Deep Learning Solutions for Automatic Blurring of Faces in Videos}

\author{Roman Plaud\\
       {\small \'Ecole des Ponts, ParisTech, France (\texttt{plaud.roman@gmail.com})} \\
        \and
        Jose-Luis Lisani\\
        {\small Universitat de les Illes Balears, Spain (\texttt{joseluis.lisani@uib.es})}\\
}

\usepackage[font={sf,footnotesize}]{caption}

\usepackage{hyperref,graphicx,amsmath,amssymb,amsthm}
\usepackage{url}

\usepackage{titlecaps}
\Addlcwords{a an on and for the of in to with from which as his their at or those by its et ses des de du la en les aux le dans zu l0}

\graphicspath{{./figures/}}

\date{}

\begin{document}

\maketitle

\begin{abstract}
The widespread use of cameras in everyday life situations generates a vast amount of data that may contain 
sensitive information about the people and vehicles moving in front of them (location, license plates, physical
characteristics, etc). In particular, people's faces are recorded by surveillance cameras in public spaces.
In order to ensure the privacy of individuals, face blurring techniques can be applied to the collected videos.
In this paper we present two deep-learning based options to tackle the problem.
First, a direct approach, consisting of a classical object detector (based on the YOLO architecture)
trained to detect faces, which are subsequently blurred. Second, an indirect approach, in which a Unet-like segmentation network is trained to output a version of the input image in which all the faces have been blurred.
\end{abstract}


\section{Introduction}\label{sec:intro}

The emergence of laws that enforce the privacy of individuals 
appearing in images and videos taken at public spaces
has led to the development of methods to anonymize visual data.
In this paper we address the problem of face anonymization, in particular
we seek to investigate methods to automatically blur people's faces.

We focus on the use of deep-learning techniques, and we investigate how
two popular architectures can help to tackle the problem.

First we take a direct approach: detect the faces and then blur them in a post-processing step.
This can be achieved using an object detector and we choose in our tests the
well-known YOLO architecture, since it is fast and accurate and a version specifically
trained to detect faces is publicly available\footnote{\url{https://github.com/deepcam-cn/yolov5-face}}.  
Section~\ref{sec:yolo} gives details of the architecture, the training set and the post-processing 
of the detections to obtain the final results.

We also investigate the use of another approach in which faces are not directly detected, but they
are directly blurred using a segmentation network. A recently proposed Unet-based network called
DeOldify\footnote{\url{https://github.com/jantic/DeOldify}} has shown outstanding results for the colorization of gray images. An open source implementation of the method, together with a rigorous analysis of its steps, was
presented in~\cite{deoldifyIPOL}. Even if the network is not designed to detect the objects or textures present
in an image, it has the ability to ascertain which is the appropriate color for each pixel of the images.
We wanted to check if this architecture, when trained with pairs of original and processed images (in which the faces had been blurred), was able to directly produce the blurred results, without the need of a detector
network. Details about DeOldify and its training are provided in Section~\ref{sec:unet}.

Experiments comparing the results obtained with both approaches are displayed in Section~\ref{sec:experiments}
and some conclusions are exposed in Section~\ref{sec:conclusions}.

\section{Face Blurring using YOLO}\label{sec:yolo}
In this section, we explore the world of face detection and the utilization of YOLOv5Face \cite{yolo}, a powerful face detection model, to accomplish this task. Our objective is to detect faces within images and subsequently apply a blur to these detected faces. To do so we rely on \cite{yolo}. We will explain the architecture of the YOLOv5Face model as well as its and training methodology. Then we will focus on the inference methodology employed to detect and blur faces effectively. \\
YOLOv5Face is a model that draws its inspiration from the popular architecture known as YOLO (You Only Look Once). YOLO, developed by Joseph Redmon et al.\ in 2016 \cite{yolo_original}, revolutionized object detection algorithms with its real-time performance. YOLO introduced the concept of a single-stage object detector, which significantly reduced the computational complexity compared to traditional two-stage models like Faster R-CNN. YOLOv5Face builds upon this foundation, utilizing YOLO's efficient architecture while specializing in facial detection tasks. By leveraging YOLO's speed and accuracy, YOLOv5Face offers a highly effective and practical solution for detecting faces in various applications.

\subsection{Architecture}
\begin{figure}[!h]
    \centering
    \includegraphics[width=\linewidth]{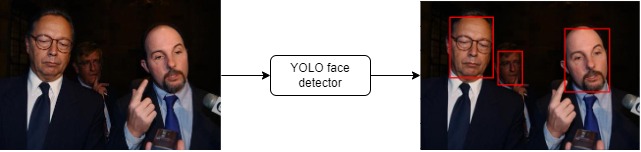}
    \caption{YOLOv5Face}
    \label{fig:yolo}
\end{figure}

YOLOv5Face is a convolutional neural network (CNN) which takes as input an image and returns coordinates of faces in that image. It returns also pixels values of 5 landmarks of detected faces, which we do not use in this study. These face coordinates are encoded as boxes, namely it returns pixel coordinates of the top left and bottom right points of the boxes. We show in Figure~\ref{fig:yolo} some results of YOLOv5Face.
\\

\begin{figure}[!h]
    \centering
    \includegraphics[width=\linewidth]{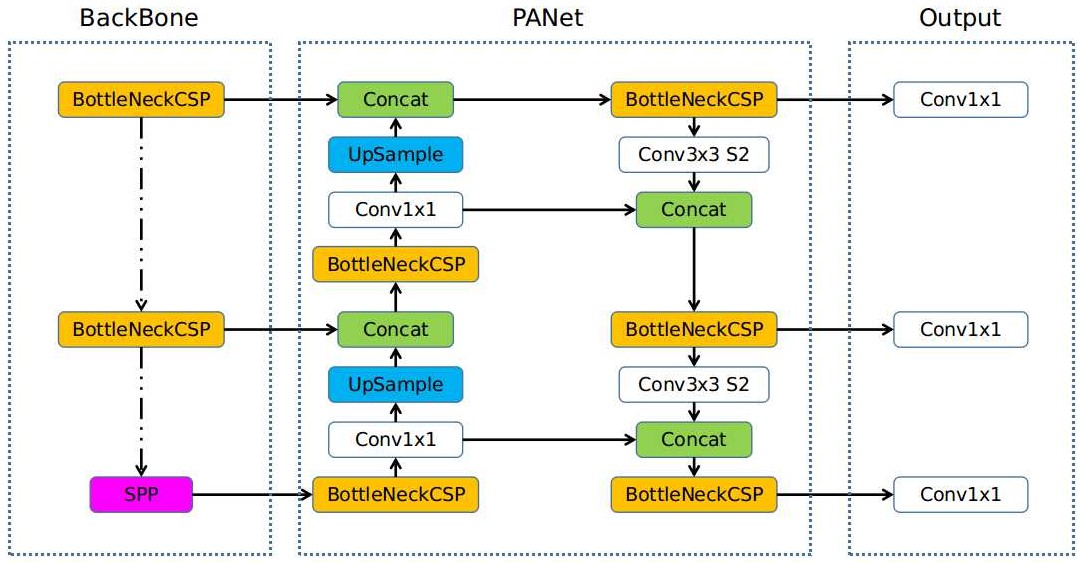}
    \caption{YOLOv5 architecture}
    \label{fig:yolov5}
\end{figure}

As it is extensively detailed in \cite{yolo}, we do not delve into details regarding the YOLOv5Face architecture. In a nutshell, it consists of the backbone, neck, and head whose architecture rely on YOLOv5 \cite{yolov5} as depicted in Figure~\ref{fig:yolov5}. There exist different sizes of architecture from Nano to X-Large whose number of parameters range from 1.73M to 141.1M. To speed up the experiments, our implementation is based on the Nano model which is the tiniest one and has 1.73M parameters.
\subsection{Dataset and Training}
All models were trained on the WIDER dataset \cite{wider}. Regarding the loss function, as for many YOLO models, it is a combination of losses: objectness score, class probability score, and bounding box regression score. In addition to these losses, a landmark regression loss was added to perform the 5 landmark regression. A SGD optimizer was used and the initial learning rate was set to 1e-2 (and a final learning rate of 1e-5). The weight decay was set to 5e-3. A
momentum of 0.8 was used in the first three warming-up epochs.
After that, the momentum was changed to 0.937. The training
run 250 epochs with a batch size of 64.

\subsection{Inference methodology}\label{sec:inf_meth_yolo}
\begin{figure}[!h]
    \centering
    \includegraphics[width=\linewidth]{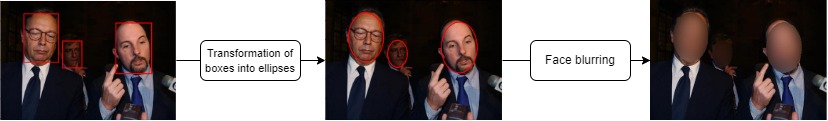}
    \caption{Yolo inference methodology}
    \label{fig:yolo_}
\end{figure}

The YOLOv5Face is only a face detector so that we need postprocessing steps to blur faces. It is important to remark that, when the method is applied to a video, we operate at a frame level, without taking into account any temporal information. Let's consider an image, we detail below how inference is performed and sum it up in Figure~\ref{fig:yolo_}.

\begin{enumerate}
    \item \textbf{Face detection}. As explained, YOLOv5Face detects faces and returns a list of boxes coordinates of faces (see Figure~\ref{fig:yolo}). The input image can be resized to speed up detection.
    \item \textbf{Transformation of boxes into ellipses}. To have a better rendering, we transform the detected boxes into ellipses. To do so, we consider the ellipses of angle $0$ and whose major and minor radius correspond to half the width and height of the box. We generate a mask where white pixels correspond to the areas in the image in which the network has detected a face.
    \item \textbf{Face blurring}. Our face blurring technique aims at being independent of the size of the faces in the images. In fact, we do not want to apply Gaussian blur with the same standard deviation to two faces in two different images. In fact, applying a Gaussian blur with the same standard deviation would result in a blurring which would be dependent on the size of the faces. We then choose, at a frame level, the standard deviation of the blurring kernel as a function of the minimum dimension of the face detected in that frame. We then blur the whole frame with the selected standard deviation. Then we replace all the pixels of the previously created mask by its blurred version.
\end{enumerate}

\textbf{Remark}. We could apply a Gaussian blur whose standard deviation is a function of each detected face dimension. But in the case of overlapping faces, the rendering is not satisfactory, creating edges that are undesirable.

Some results are displayed in Section \ref{sec:experiments}.

\section{Face Blurring using DeOldify}\label{sec:unet}

To perform the face blurring task, we heavily rely on the work in \cite{deoldifyIPOL}. We detail in this section our motivations. 
First, we did not want to train a face detector, as there are already a great amount of them trained on a greater amount of data \cite{yolo}. With a face detector we would have a second step that would consist in blurring all the faces detected, as explained in Section~\ref{sec:yolo}. What we wanted is to directly operate on the image and automatically perform blurring, getting rid of the detection step. \\
To do so, the Unet-like architecture described in \cite{deoldifyIPOL} seem well fitted. Notably, this model has demonstrated success in colorizing images, particularly skin tones and faces. This indicates the model's capability to perform two tasks: skin/face identification and its colorization.  Consequently, we were motivated to select the same architecture for our approach.

\subsection{Architecture}
\begin{figure}[!h]
    \centering
    \includegraphics[width=\linewidth]{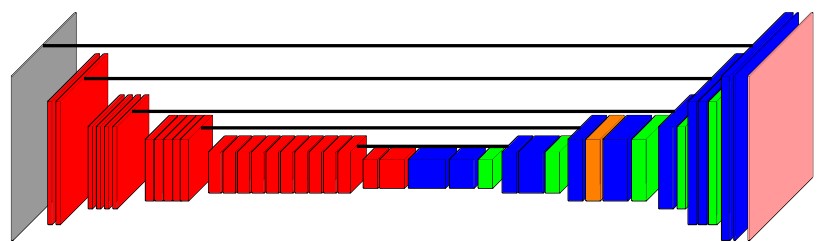}
    \caption{DeOldify architecture (from \cite{deoldifyIPOL}). In red: pretrained ResNet, in blue: convolutional blocks, in green: upsample layers, in
orange: self-attention layer, and in pink: sigmoid layer. The black lines stand for the skip connections}
    \label{fig:unet}
\end{figure}

We keep the architecture used in \cite{deoldifyIPOL}, that is a Unet architecture \cite{unet}, composed of an encoder whose weights are initialized with ResNet50 \cite{resnet} checkpoint trained on ImageNet \cite{imagenet}. These encoder weights are \textbf{freezed} during training. The decoder is a standard Unet decoder, except from the fact that a self-attention layer is added as displayed in Figure~\ref{fig:unet}. We also conducted experiments getting rid of the self-attention layer (see Section~\ref{sec:experiments}).

\subsection{Dataset for training}

\subsubsection{Datasets overview}

In this section, we detail how we perform the dataset construction. We use two different datasets:
\begin{enumerate}
    \item Face Detection Data Set and Benchmark (FDDB) \cite{fddb} 
    \item WIDER FACE: A Face Detection Benchmark (WIDER) \cite{wider}.
\end{enumerate}
Both datasets are face detection datasets, which means that each image comes with annotations that provide information about the localization of every human face in the image.\\
The FDDB dataset contains annotations for 5171 faces in a set of 2845 images. For each image the face annotations are encrypted in an ellipse fashion $\left(r_a, r_b, \theta, x, y\right)$ where:
\begin{itemize}
    \item $\left(r_a, r_b\right)$ are the major and minor axis radius of the ellipse (in pixels)
    \item $\theta$ is the orientation angle of the ellipse
    \item $\left( x, y \right)$ are the pixel coordinates of the center of the ellipse.
\end{itemize}
The WIDER dataset contains 32,203 images and label 393,703 faces. For each image the face annotations are encrypted in as rectangle fashion $\left(x, y, w, h\right)$ where:
\begin{itemize}
    \item $\left(x, y\right)$ are the pixel coordinates of the top left corner.
    \item $\left(w, h \right)$ are the width and the height of the rectangle (in pixels)
\end{itemize}
One observation that immediately stands out is that the FDDB face annotations appear to be more precise, as they are encrypted as ellipses. However, it is worth noting that the WIDER dataset is significantly larger in size.

\subsubsection{Construction methodology}\label{sec:cons_dat}

As previously mentioned, the input object for this task should be the raw image, while the desired output is the same image with blurred faces. To achieve this, we must combine the image with its corresponding annotations in order to generate the blurred version. In order to achieve this we proceed exactly as in Section~\ref{sec:inf_meth_yolo}.\\

\begin{figure}[!h]
    \centering
    \begin{subfigure}[b]{0.45\linewidth}
        \centering
        \includegraphics[width=0.8\linewidth]{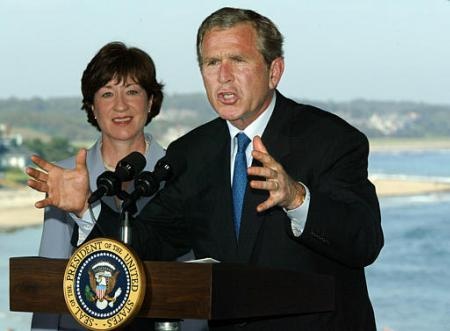}
        \label{fig:subfig1}
    \end{subfigure}
    \begin{subfigure}[b]{0.45\linewidth}
        \centering
        \includegraphics[width=0.8\linewidth]{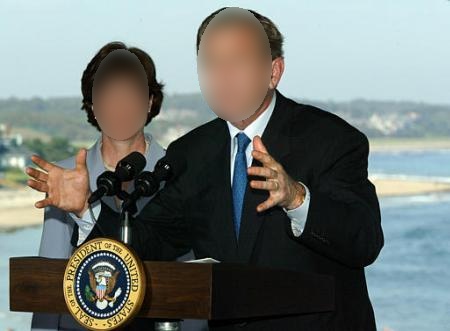}
        \label{fig:subfig2}
    \end{subfigure}
    \begin{subfigure}[b]{0.45\linewidth}
        \centering
        \includegraphics[width=0.8\linewidth]{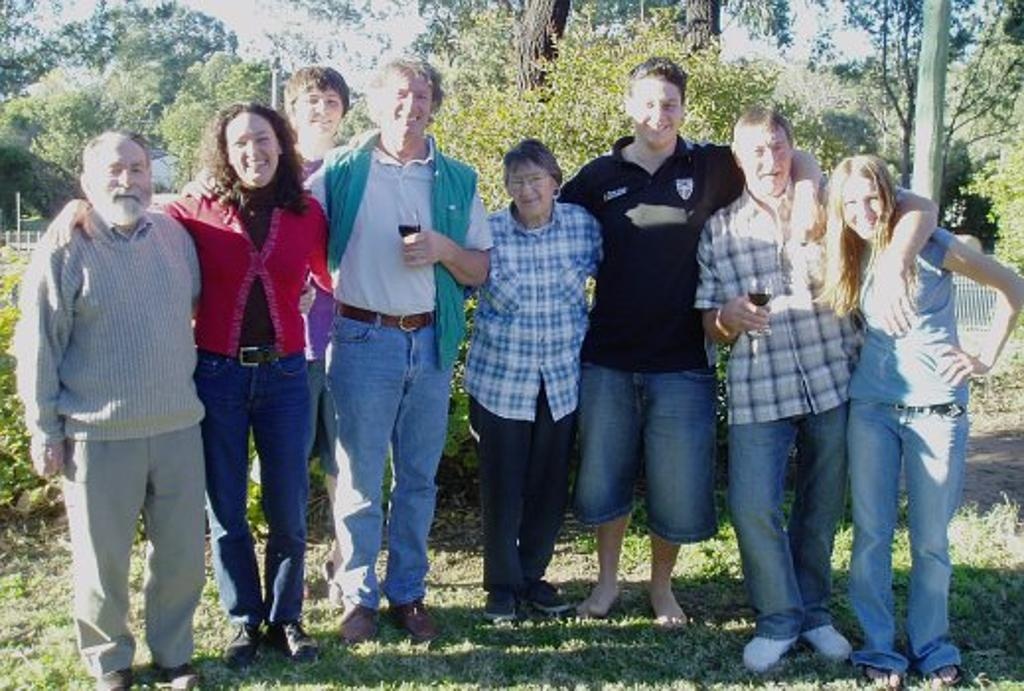}
        \label{fig:subfig3}
    \end{subfigure}
    \begin{subfigure}[b]{0.45\linewidth}
        \centering
        \includegraphics[width=0.8\linewidth]{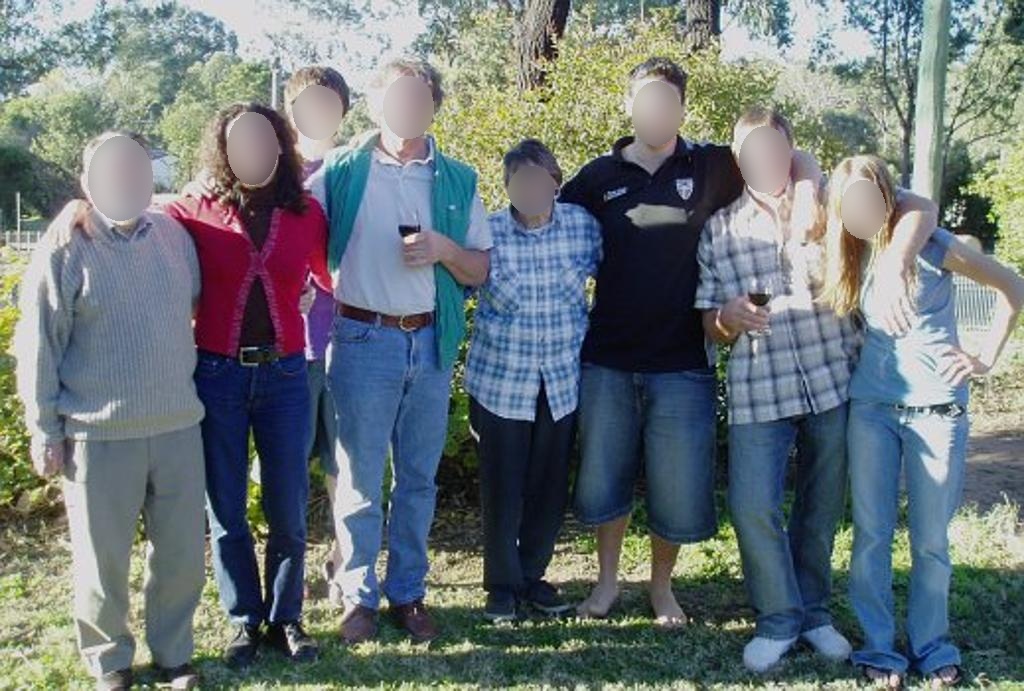}
        \label{fig:subfig4}
    \end{subfigure}
    \caption{Correspondence inputs-targets for an image of FDDB dataset (up) and WIDER dataset (down).}
    \label{fig:1}
\end{figure}

For the FDDB dataset annotations come in an ellipse fashion so that we perform directly \textbf{step 3} (Face Blurring). For the WIDER dataset annotations come as boxes, so that we perform \textbf{step 2} (Transformation of boxes into ellipses) and \textbf{step 3} (Face Blurring).\\
It is worth noting that there is no clear ground truth in our problem as several blurs can perform as efficiently the task we want to implement. We display in Figure~\ref{fig:1} an example of both inputs and targets for both datasets.

We then build a train/validation split consisting of 15148 training images pairs and 3803 validation images pairs. As the WIDER dataset came with a train/validation split we keep that split and create one only for the FBBD dataset.

\subsection{Training}
Training is performed in a supervised fashion with the couple of input-targets described in the previous section. In this section we detail the choice of loss function and of training procedure.
\subsubsection{Loss function}
In the original paper \cite{deoldifyIPOL}, training was performed using a feature loss inspired by \cite{vggloss}. It corresponds to computing a weighted average L1 loss of intermediate outputs of a VGG network. As we consider that the VGG features extracted from a face and its blurred counterpart are intuitively similar, we did not think relevant to use this loss. Instead we focus on more classical losses: L1 loss and Mean Square Error (MSE) for which we recall the definition below.
\begin{itemize}
    \item L1 loss : $l_1\left(\textbf{u}, \textbf{v}\right) = \frac{1}{\left|\textbf{u}\right|}\underset{\textbf{i}\in\textbf{u}}{\sum}\left| \textbf{u}\left(\textbf{i}\right) - \textbf{v}\left(\textbf{i}\right)\right|$
    \item MSE : $MSE\left(\textbf{u}, \textbf{v}\right) = \frac{1}{\left|\textbf{u}\right|}\underset{\textbf{i}\in\textbf{u}}{\sum}\left( \textbf{u}\left(\textbf{i}\right) - \textbf{v}\left(\textbf{i}\right)\right)^2$
\end{itemize}

Where $u$ and $v$ are, respectively, the output of the network and the ground truth. $i$ represent a pixel and $\left|\textbf{u}\right|$ the number of pixels of $u$.

\subsubsection{Training Procedure}
To train the network, we follow the procedure in \cite{deoldifyIPOL}  which proposes a progressive training that begins by learning on smaller image sizes and to progressively increase the size of images until we reach computation limitations. At each step,
\begin{itemize}
    \item The model is initialized with the previous step final model (randomly if Step 1).
    \item The batch size is selected to match  computation resources.
    \item The learning rate is selected with regards to the batch size.
    \item The model is trained for 20 epochs on the constructed dataset.
\end{itemize}
We sum up in the Table~\ref{tab:training} the training hyperparameters of each step of training.

\begin{table}[!h]
\centering
\resizebox{\linewidth}{!}{
\begin{tabular}{c|c|c|c}
     & Image size & Batch Size & learning rate init.\\
    \hline
    Step 1 & $64\times64$ & $80$ & $10^{-3}$  \\
    \hline
    Step 2 & $128\times128$ & $20$ & $10^{-4}$  \\
    \hline
    Step 3 & $192\times192$ & $8$ & $5\times10^{-5}$  \\
\end{tabular}
}
\caption{Training details.}
\label{tab:training}
\end{table}

Training is performed using an AdamW optimizer coupled with a exponential decay learning rate scheduler of multiplicative factor of $0.8$. We use a GPU NVIDIA GeForce RTX 2080 Ti of 11GB.

\subsection{Inference methodology}\label{sec:inf_meth}
\begin{figure}[!h]
    \centering
    \includegraphics[width=\linewidth]{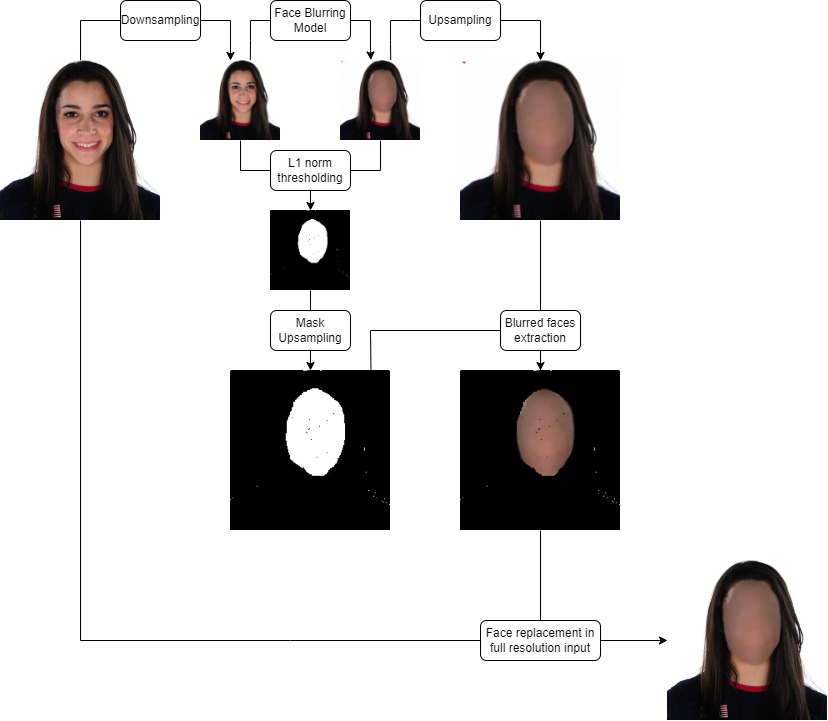}
    \caption{Inference methodology.}
    \label{fig:schema}
\end{figure}

Upon completing the training process, our objective is to develop a methodology that performs face blurring in an input image of arbitrary dimensions. As in the case of YOLO, when we want to blur a video we operate at a frame level, getting rid of any temporal information. \\
A naive strategy would be to directly give the raw image as input to the network but the higher the image resolution the higher the computation time  will be. If we do not control the input image resolution we can not control the computation time. Then, we develop an inference methodology which we detail below and which is summed up in Figure~\ref{fig:schema}. 

\begin{itemize}
    \item \textbf{Downsampling}. The input image is downsampled to a given size. In practice we use either $192\times192$, $256\times256$ or $512\times512$.
    \item \textbf{Model forward}. The resized image is passed through the network.
    \item \textbf{Upsampling}. The blurred image is resized to its original input size.
    \item \textbf{L1 norm threshold}. We compute the difference between the value at each pixel in the downsampled input image and the downsampled blurred image. By thresholding these differences a binary mask containing the face is obtained.
    \item \textbf{Mask Upsampling}. The mask is resized to obtain a new mask with the dimensions of the original input.
    \item \textbf{Blurred faces extraction}. We combine the resized mask and the resized upsampled output to extract only faces from the output.
    \item \textbf{Face replacement in the full resolution input}. We combine the full resolution input with the blurred faces extracted at the previous step to construct the full resolution image with blurred faces.
\end{itemize}

What is left to determine is the choice of hyperparameters in that methodology, namely the dimension on downsizing, as well as the level of thresholding for mask extraction.
\begin{itemize}
    \item \textbf{Choice of thresholding}. This choice is totally empirical. After a careful study, we observe that a wide range of thresholds give almost the same results. In fact, following the code implementation in \cite{deoldifyIPOL}, the difference between the values of the input and the output of the network range from -3 to 3 so that if the model blurs a face the difference in the pixel is significantly different from $0$ (on average a pixel blurred has a difference of $0.4$ with respect to the original value). We then set the L1 threshold at $0.1$.
    \item \textbf{Downsampling size}. Regarding the size of downsampling, this hyperparameter is crucial. In fact, during training, the bigger the size of the face in the image, the bigger is the difference between input and target and the easier it is for the network to learn to detect and blur the face. Therefore, bigger faces tend to be much more easily detected and blurred. At inference time, what we observe is that if we downsample at the size of the last step of the network training ($192\times192$) it fails to detect small faces that do not cover much pixels (see Section~\ref{sec:experiments} for examples of the phenomenon). For this reason, it is interesting to set a higher downsampling size ($512\times512$ for example), so that small faces cover more pixels and consequently are more easily detected by the network. Of course, the counterpart is that inference time in then higher. This tradeoff will be discussed in Section~~\ref{sec:experiments}.
\end{itemize}

\section{Experiments}\label{sec:experiments}
\subsection{Evaluation Methodology}
In the face blurring context, evaluation can not be done completely quantitatively. In fact, for a given image, there is a large amount of different blurs that can complete the task. Still, we can compare methods using two criteria:
\begin{enumerate}
    \item \textbf{Visual evaluation and face counting}. We evaluate visually if face blurring is correctly done through a straight-forward criteria. Correctly blurred means we cannot recognize the original person in the output image. For each test set image, we count all correctly blurred faces.

    \item \textbf{Computation time}. This is a quantitative metric, which allows us to compare the inference time computation between models.
\end{enumerate}

To perform evaluation we construct a test set of 6 images which features several specific particularities, namely:
\begin{itemize}
    \item Variations in face distance to the camera.
    \item Variations in orientation of faces (faces not facing the camera).
    \item Variations in number of faces.
    \item Variations in the face itself (partially masked, wearing glasses or hats).
\end{itemize}

In this section we will test the influence on the results of several hyperparameters of the Unet network. We will extensively study the downsampling dimension for inference, the choice of the loss and the relevance of the self-attention layer. We will also conduct some experiments using the YOLOv5Face face detector.

\subsection{Experiments using DeOldify Unet}

\subsubsection{Influence of the downsampling dimension for inference}
In this section, we study the influence of the downsampling dimension using the Unet method. We recall that the downsampling dimension corresponds to the size to which the image is resized before passing it through the face blurring model. The last stage of training of the model uses input images of size $192\times192$, we then could think the network is more fitted to be inferred at this size. But practically and as we said in Section~\ref{sec:inf_meth}, bigger faces tend to be more easily detected and blurred, as they cover more pixels during training, which facilitates detection. That observation motivates the study of the influence of the downsampling dimension. \\
In this section, all results correspond to models that were trained with the L1 loss.\\

\begin{figure}[h]
    \centering
    \includegraphics[width=0.24\linewidth]{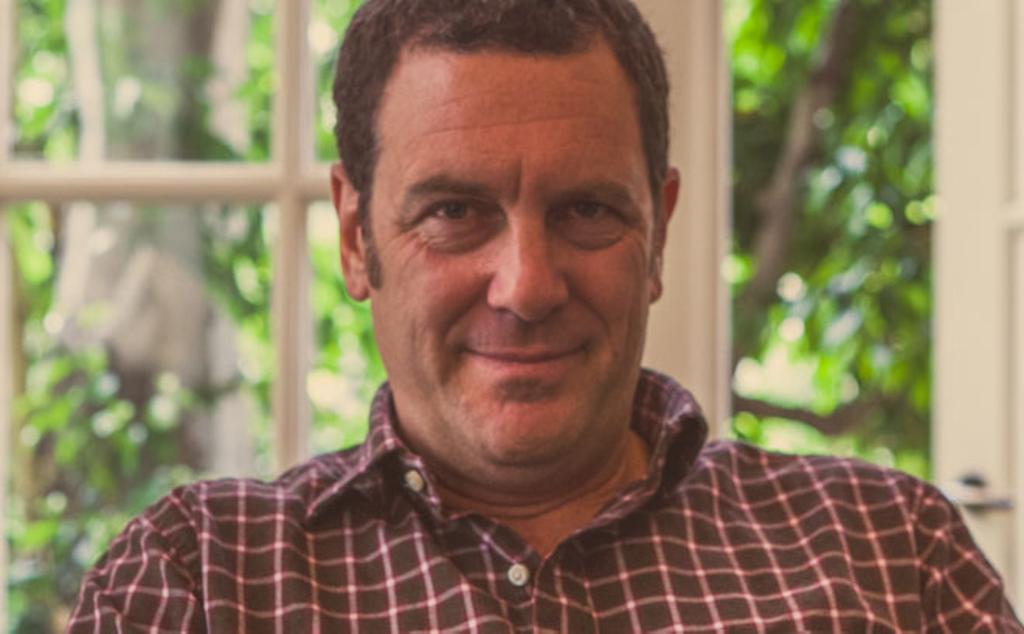}
    \includegraphics[width=0.24\linewidth]{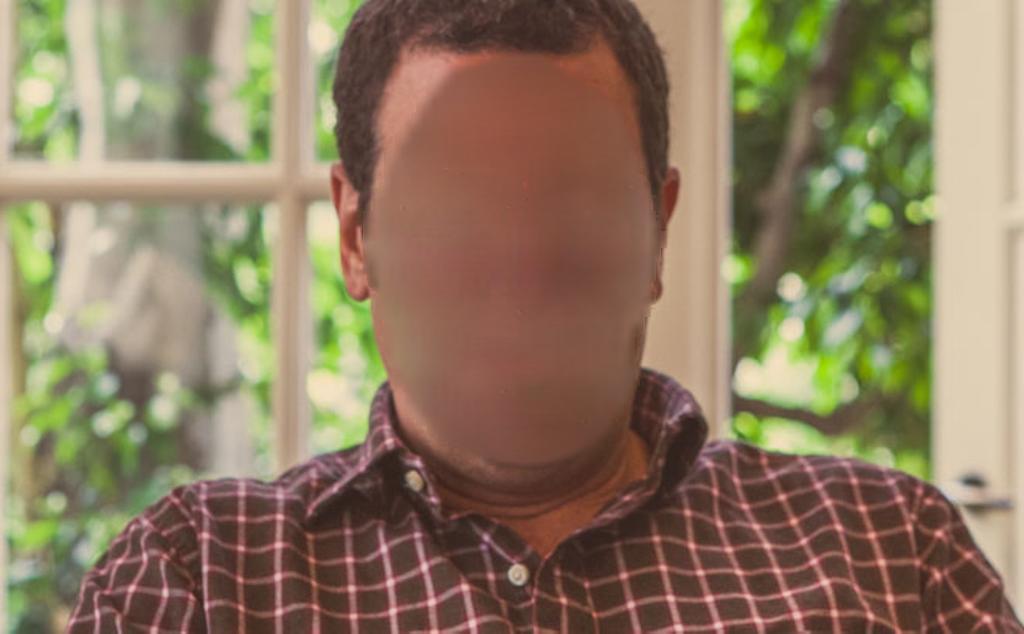}
    \includegraphics[width=0.24\linewidth]{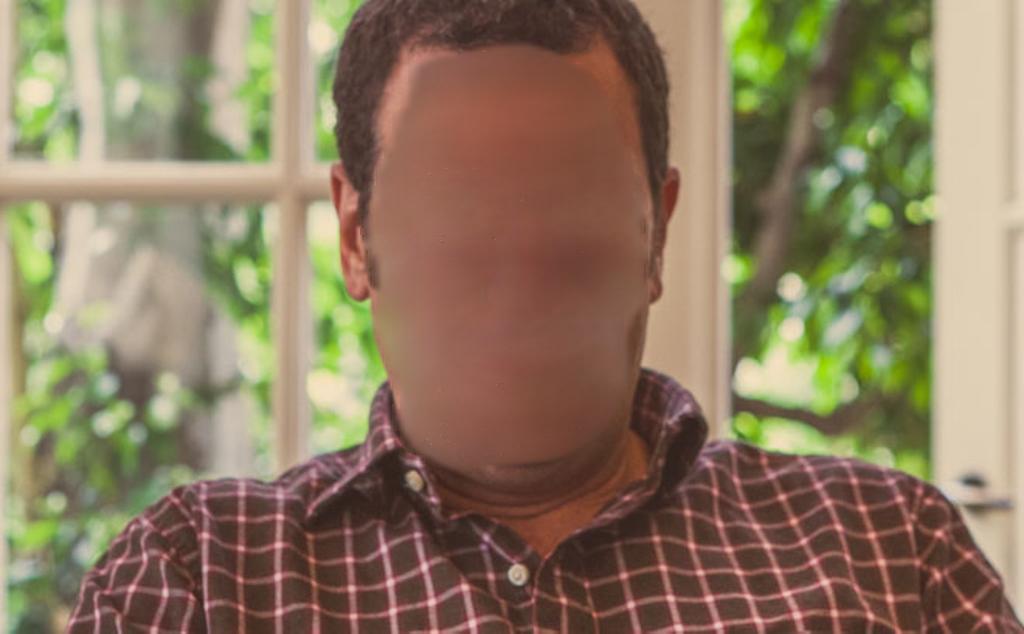}
    \includegraphics[width=0.24\linewidth]{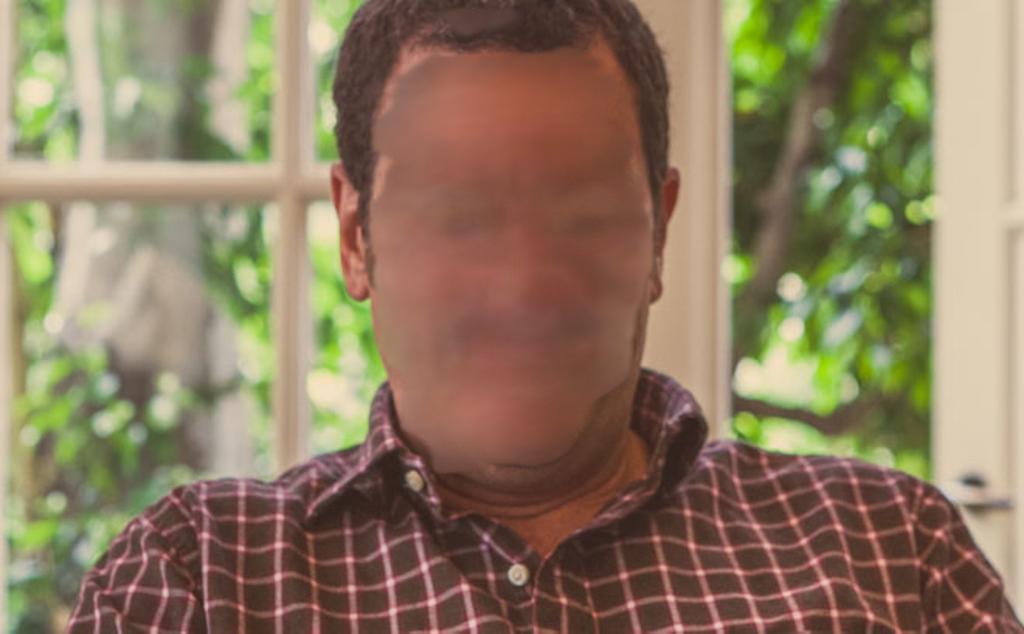}\\
    
    \includegraphics[width=0.24\linewidth]{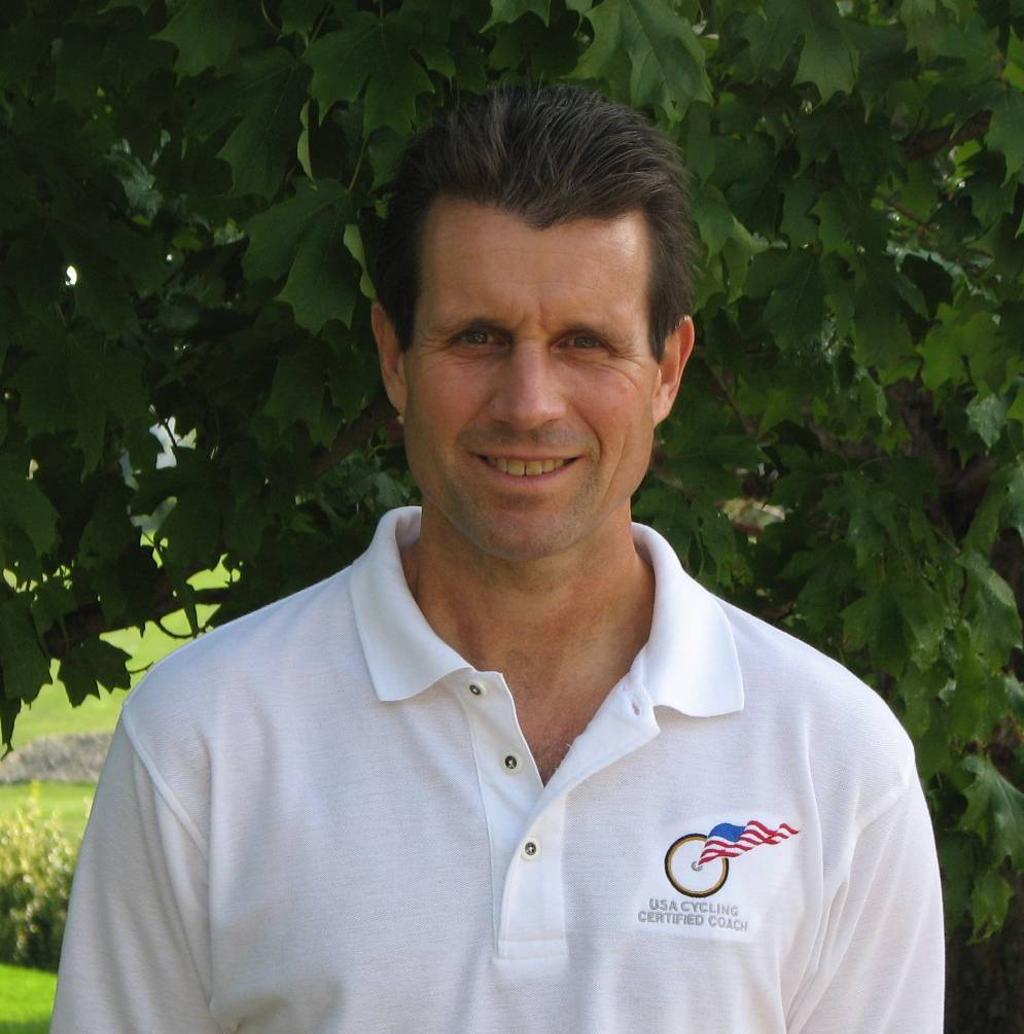}
    \includegraphics[width=0.24\linewidth]{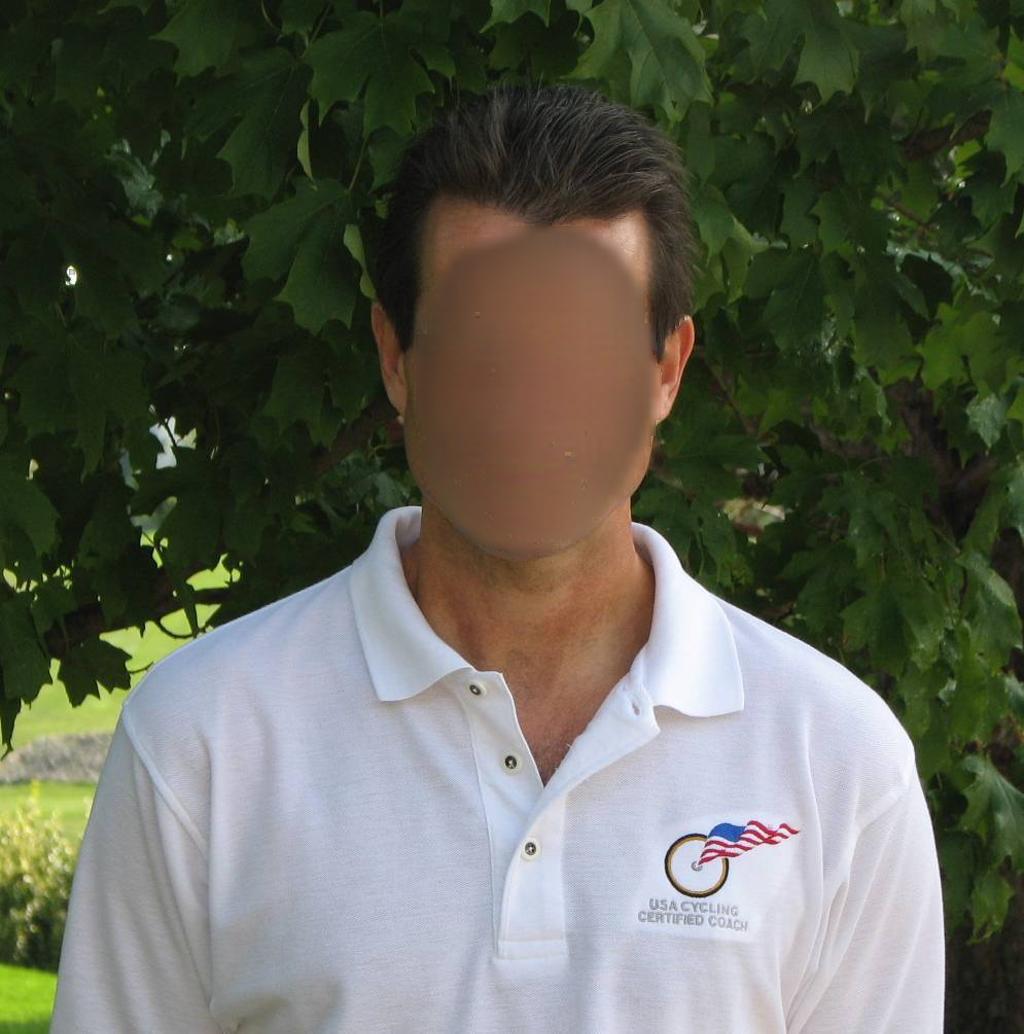}
    \includegraphics[width=0.24\linewidth]{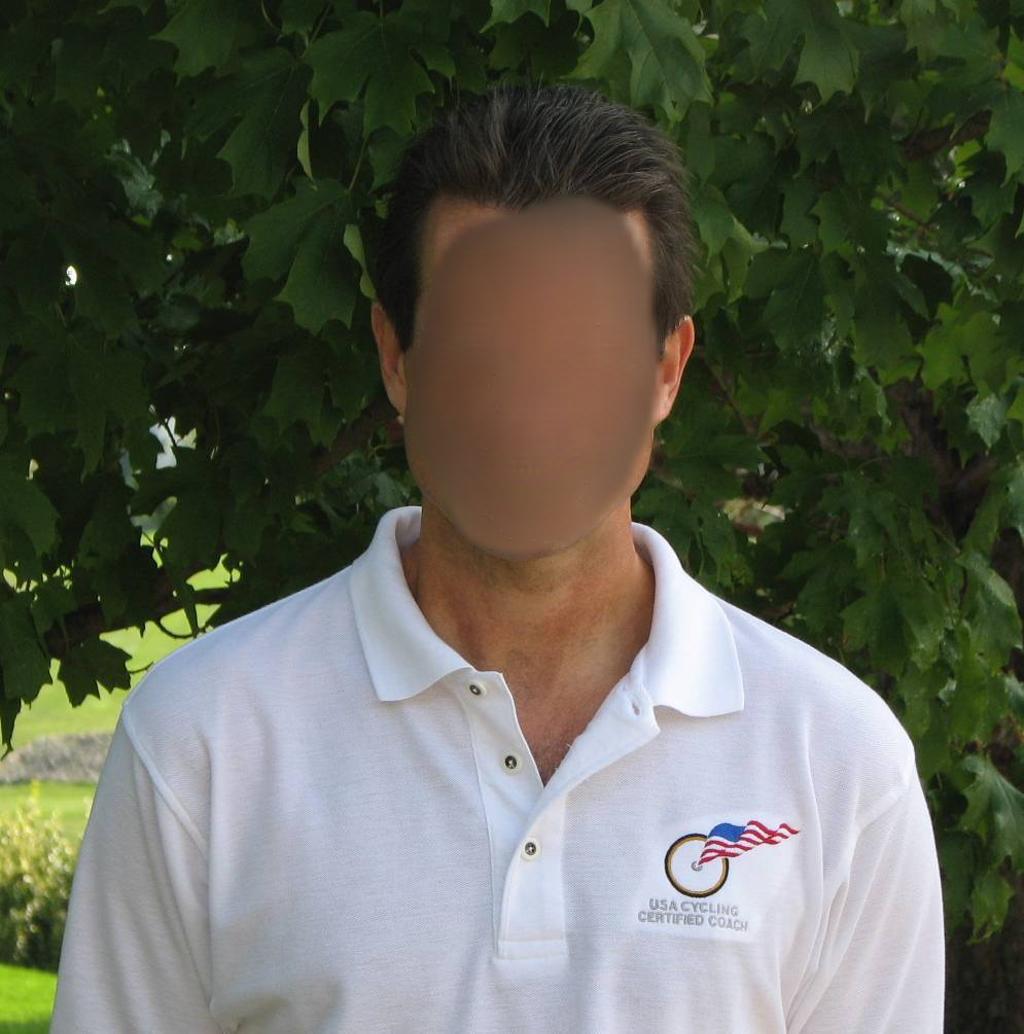}
    \includegraphics[width=0.24\linewidth]{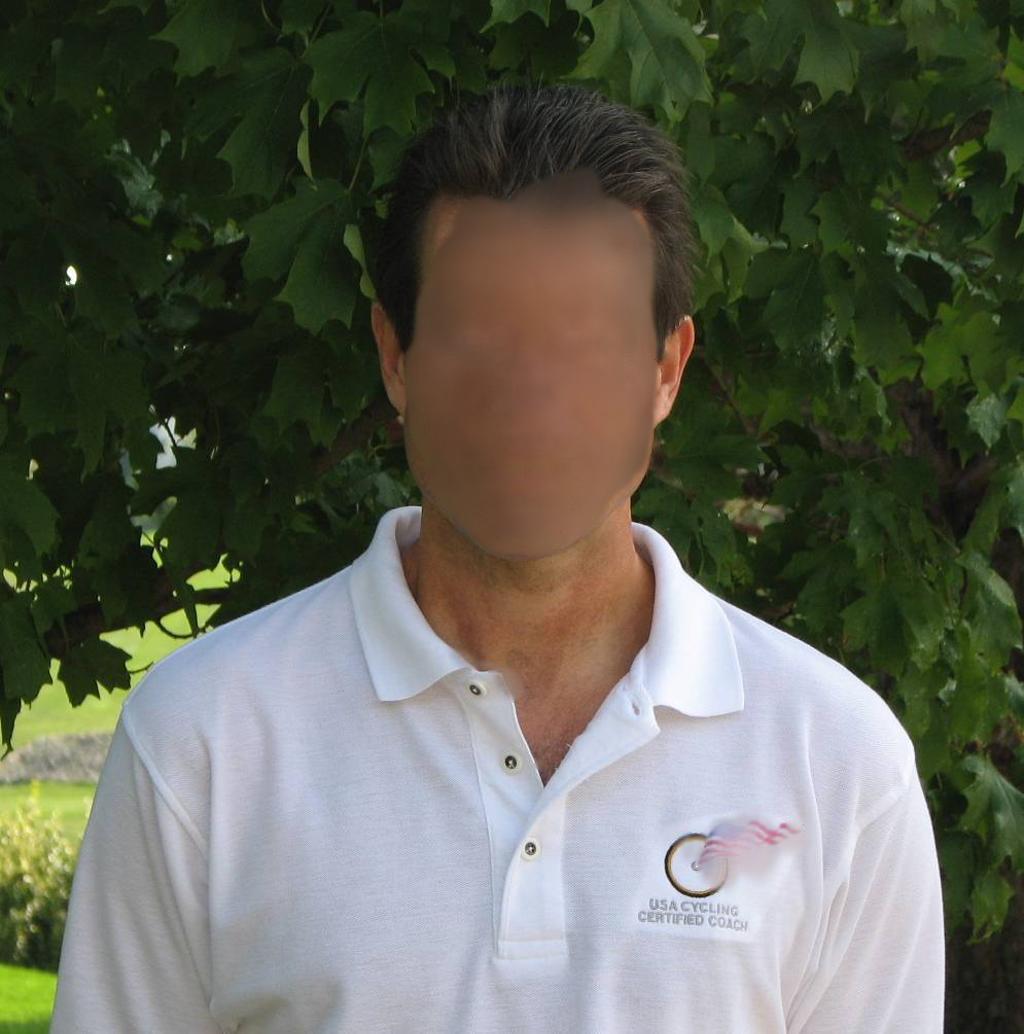}\\
    
    \includegraphics[width=0.24\linewidth]{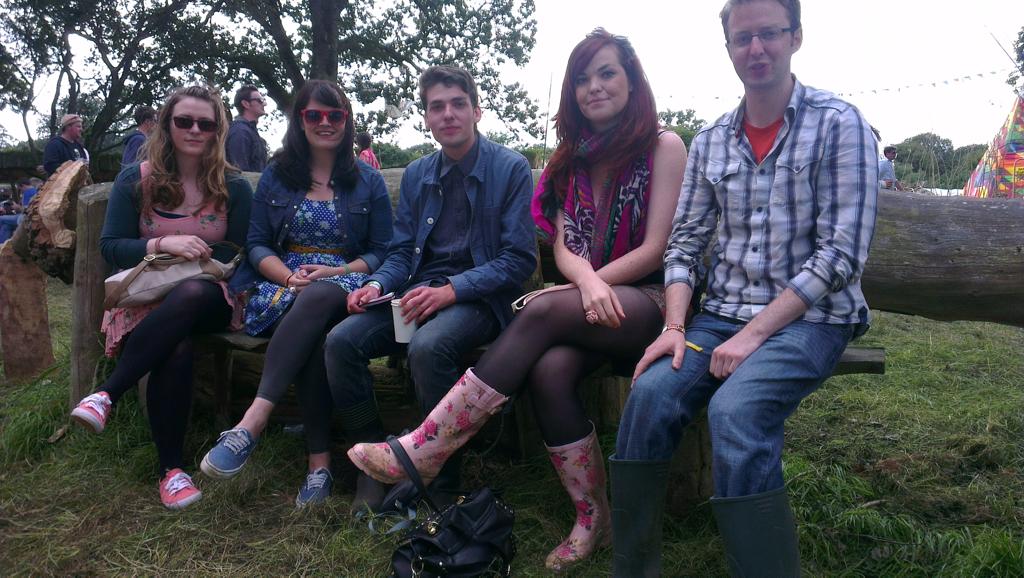}
    \includegraphics[width=0.24\linewidth]{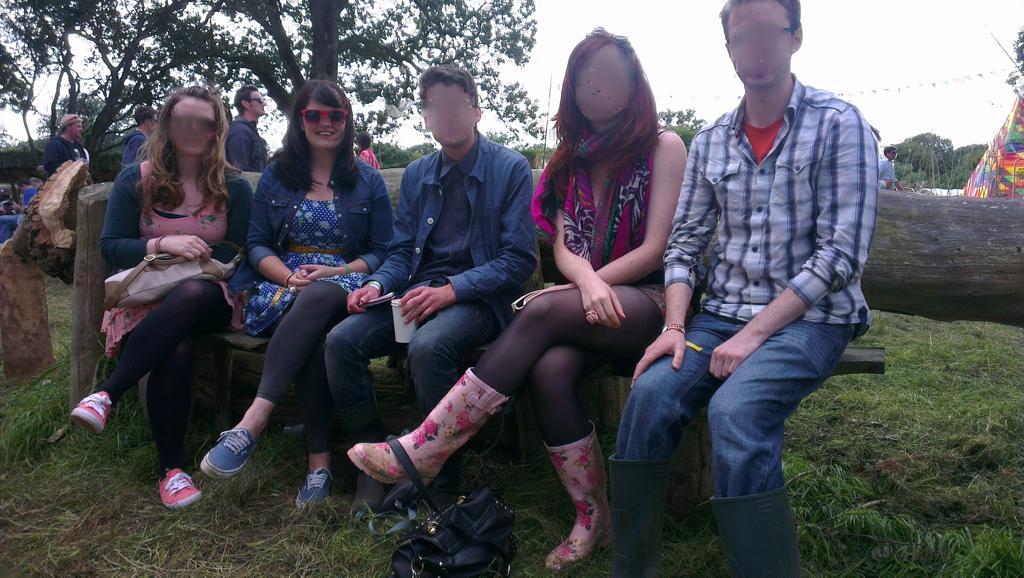}
    \includegraphics[width=0.24\linewidth]{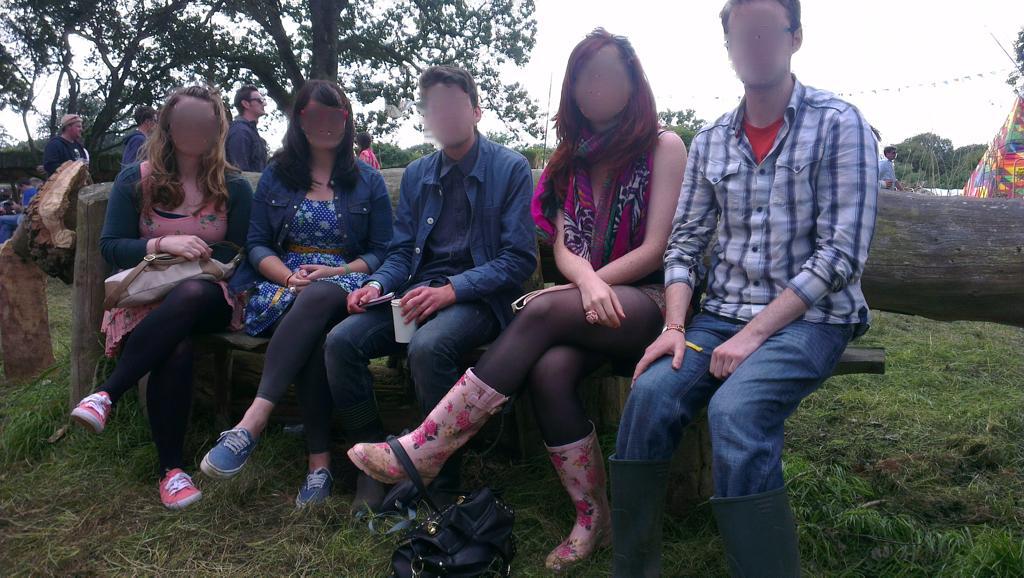}
    \includegraphics[width=0.24\linewidth]{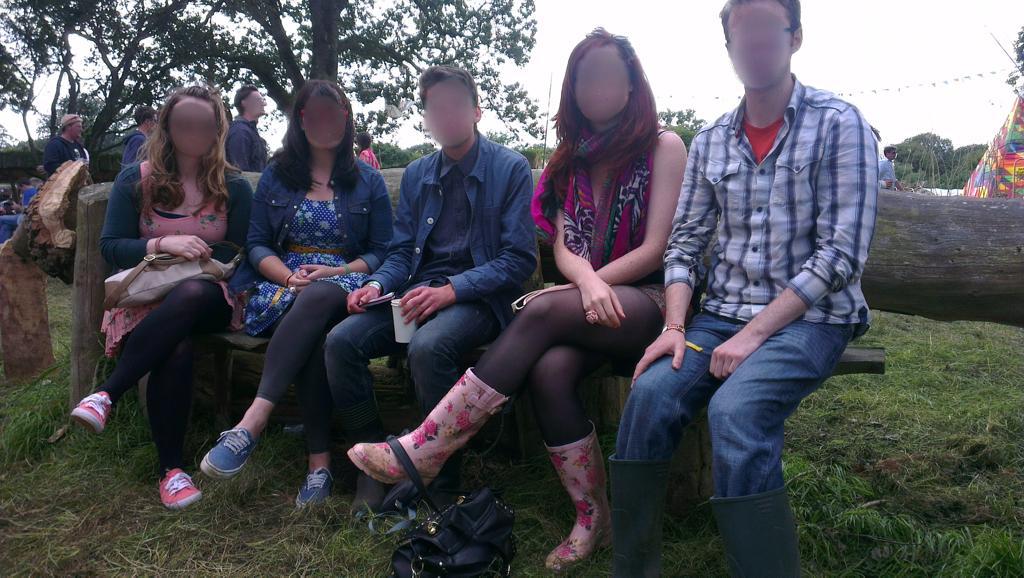}\\

    \includegraphics[width=0.24\linewidth]{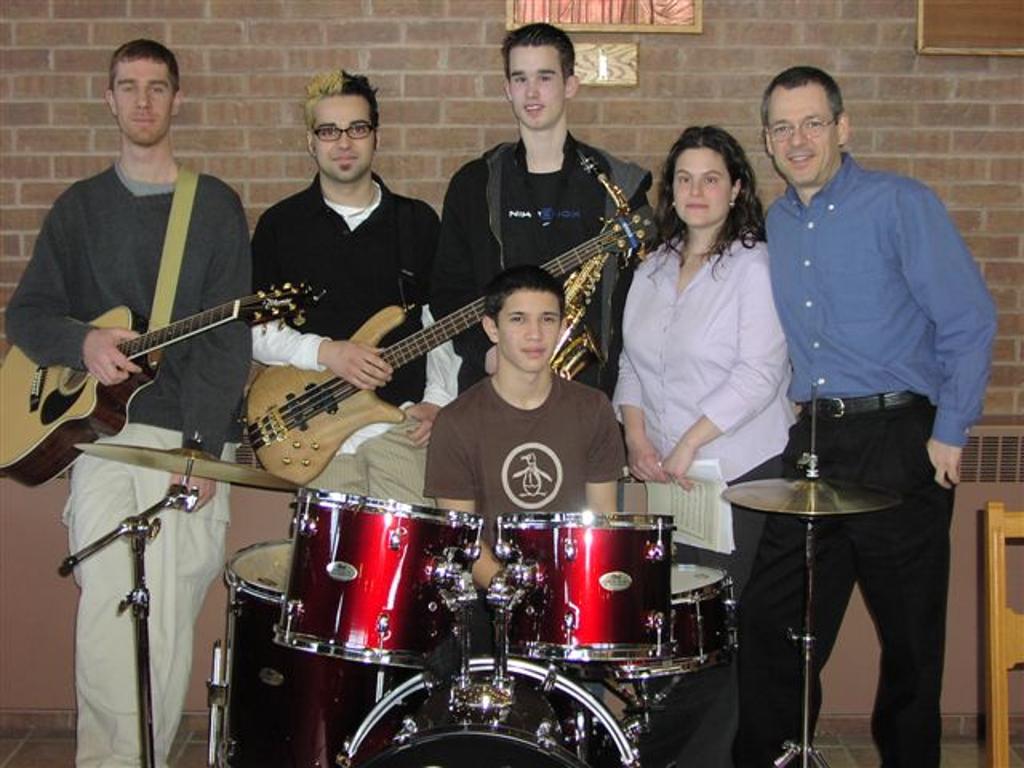}
    \includegraphics[width=0.24\linewidth]{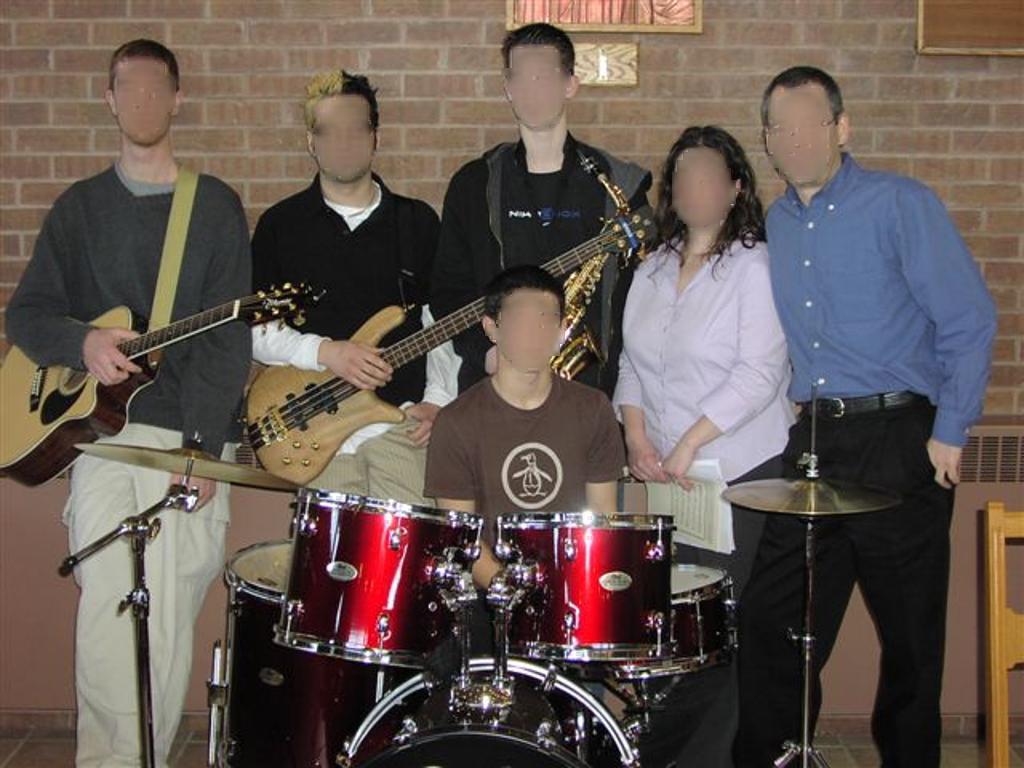}
    \includegraphics[width=0.24\linewidth]{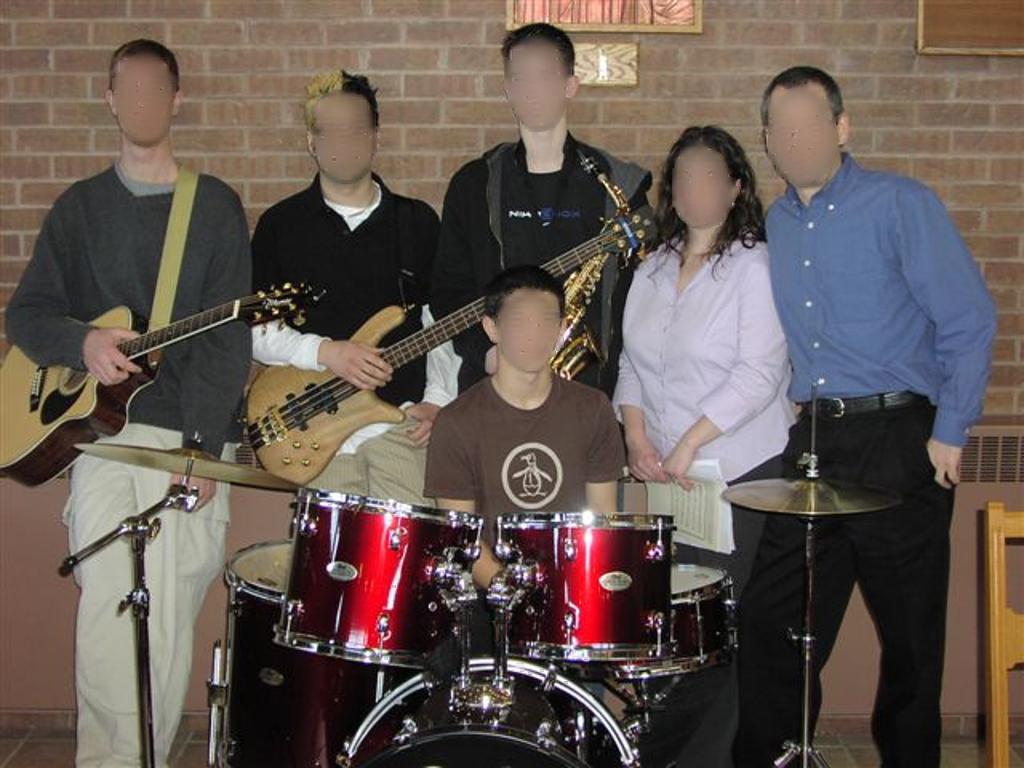}
    \includegraphics[width=0.24\linewidth]{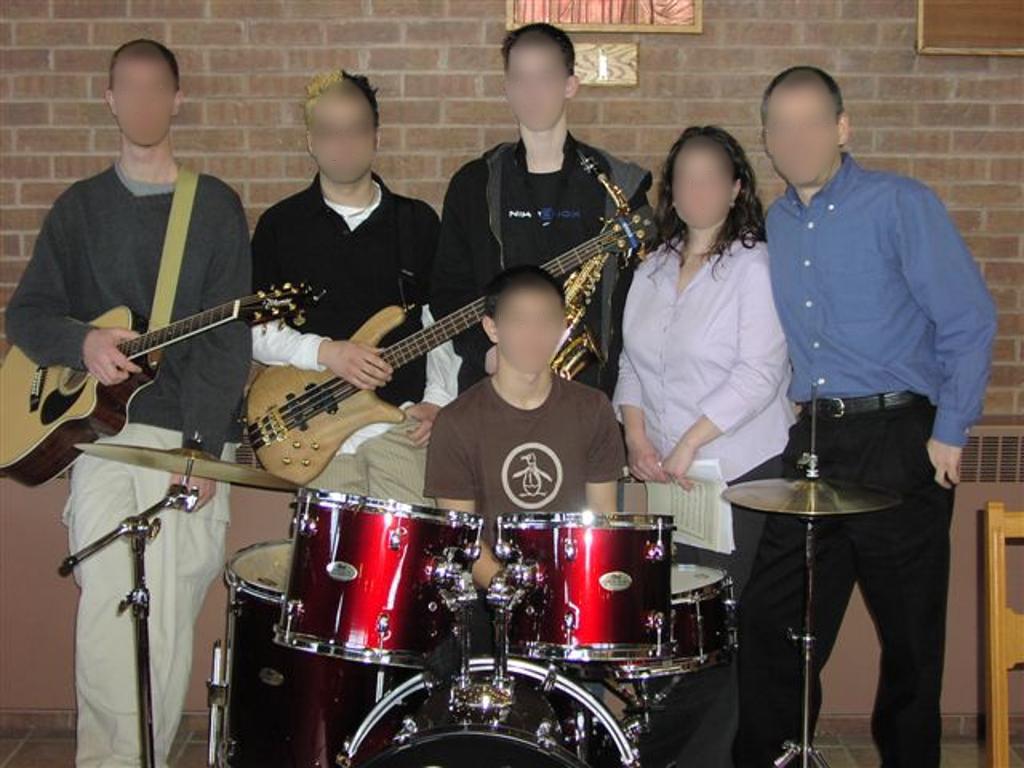} \\

    \includegraphics[width=0.24\linewidth]{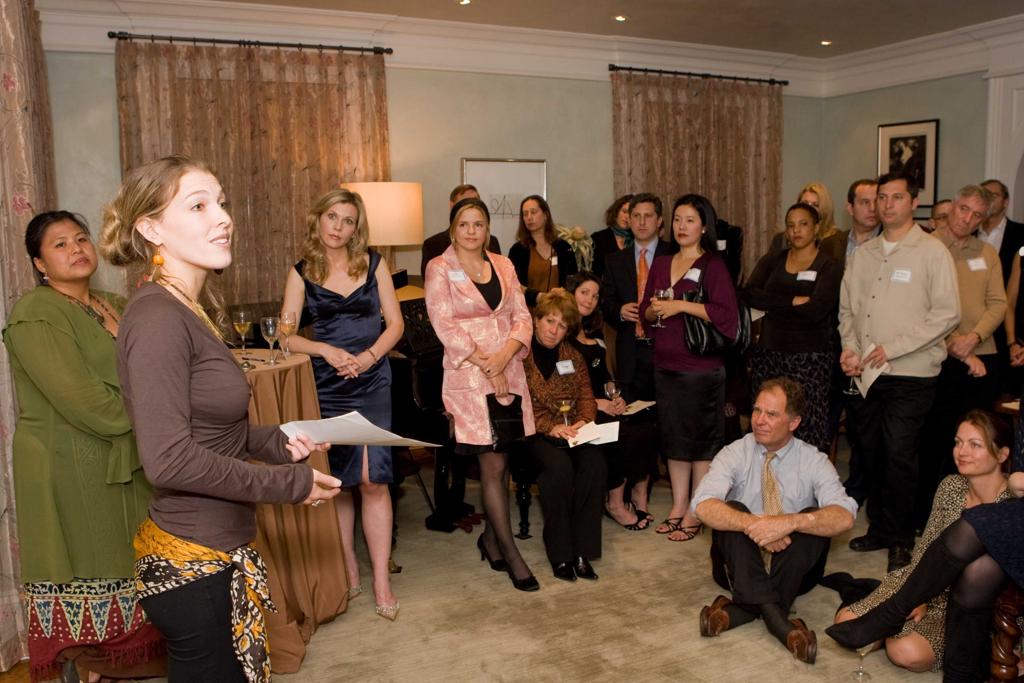}
    \includegraphics[width=0.24\linewidth]{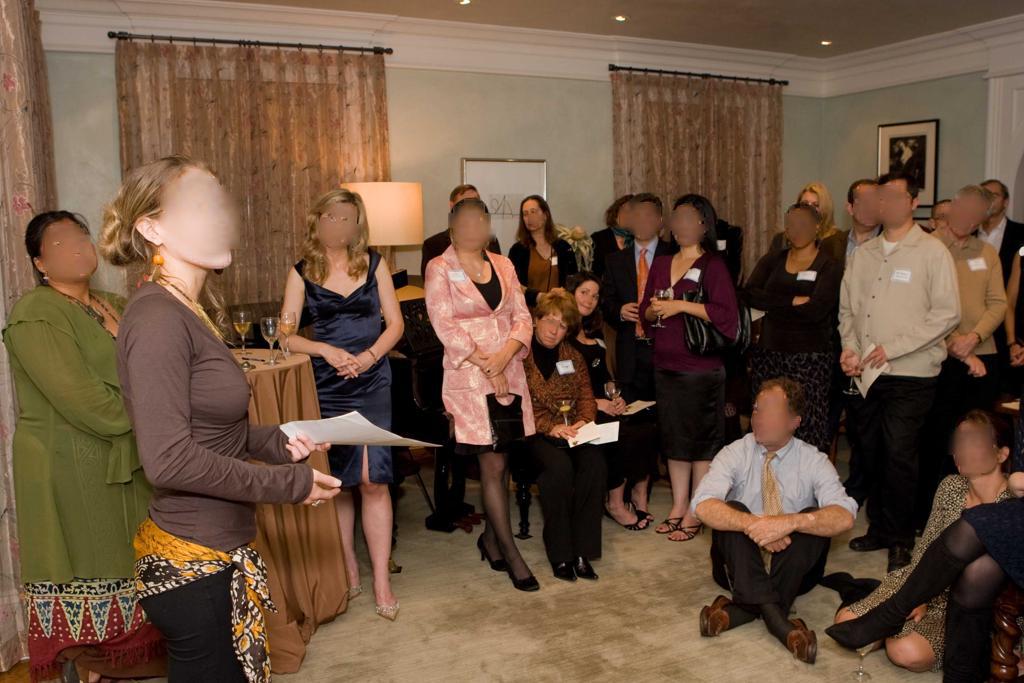}
    \includegraphics[width=0.24\linewidth]{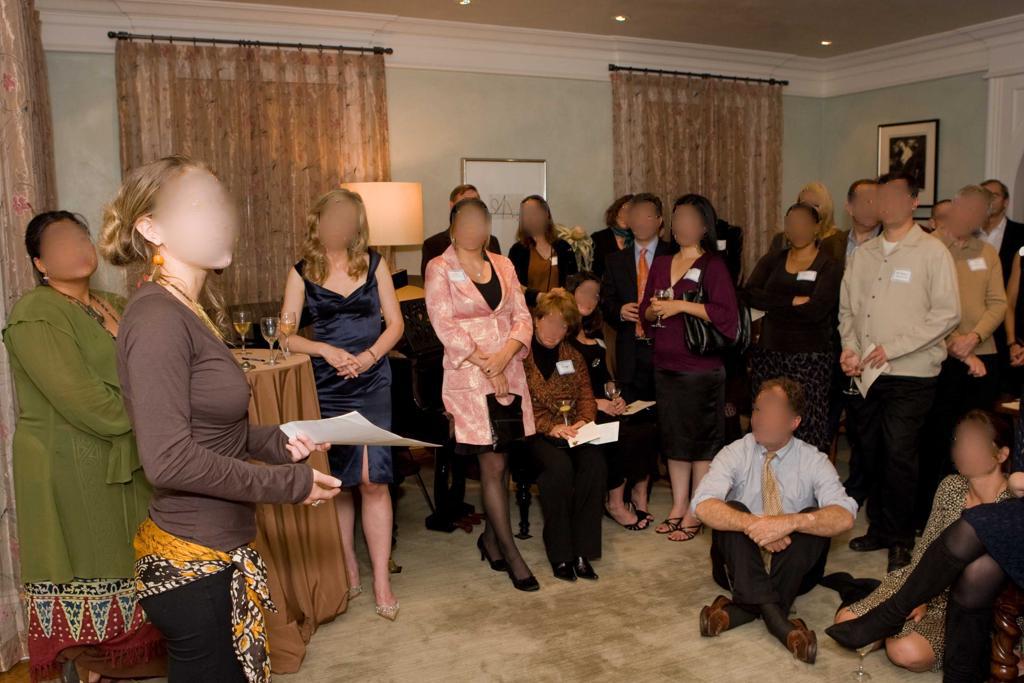}
    \includegraphics[width=0.24\linewidth]{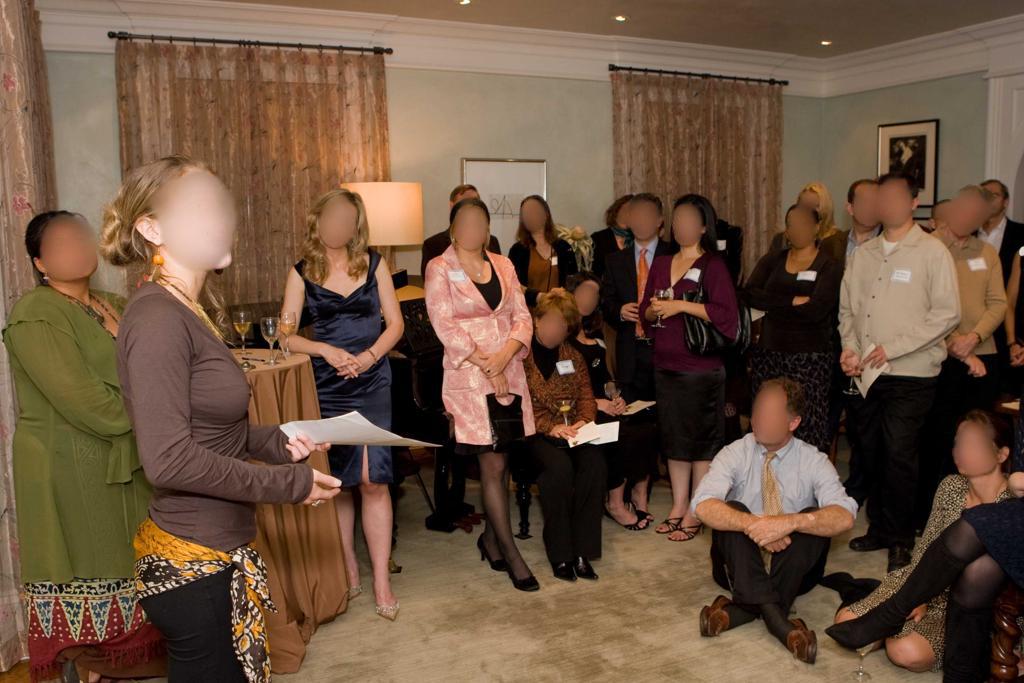}\\
    
    \includegraphics[width=0.24\linewidth]{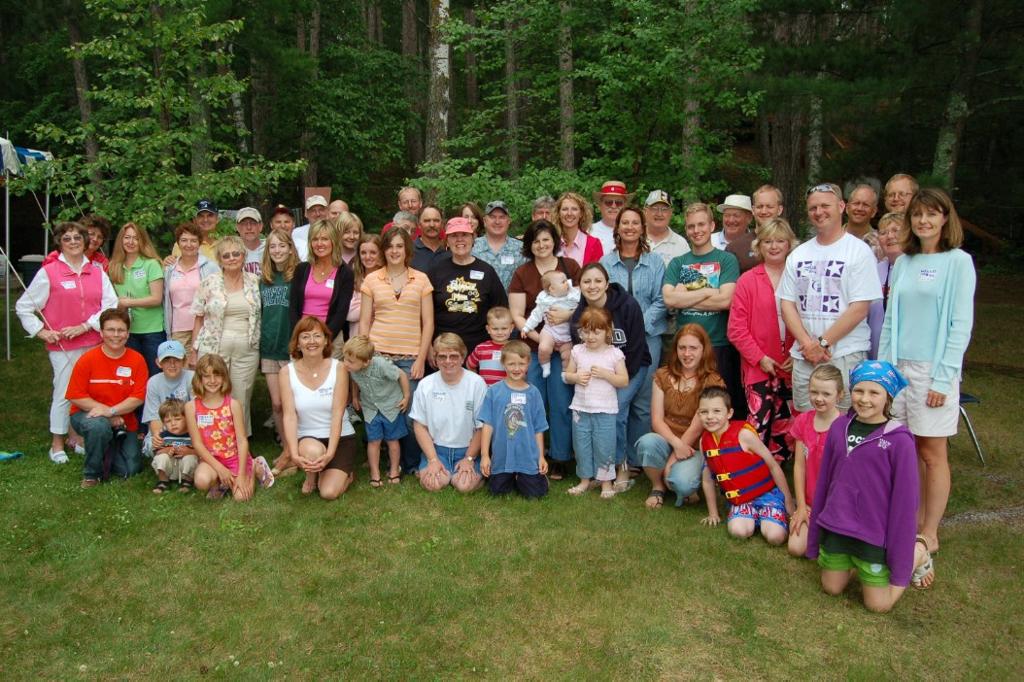}
    \includegraphics[width=0.24\linewidth]{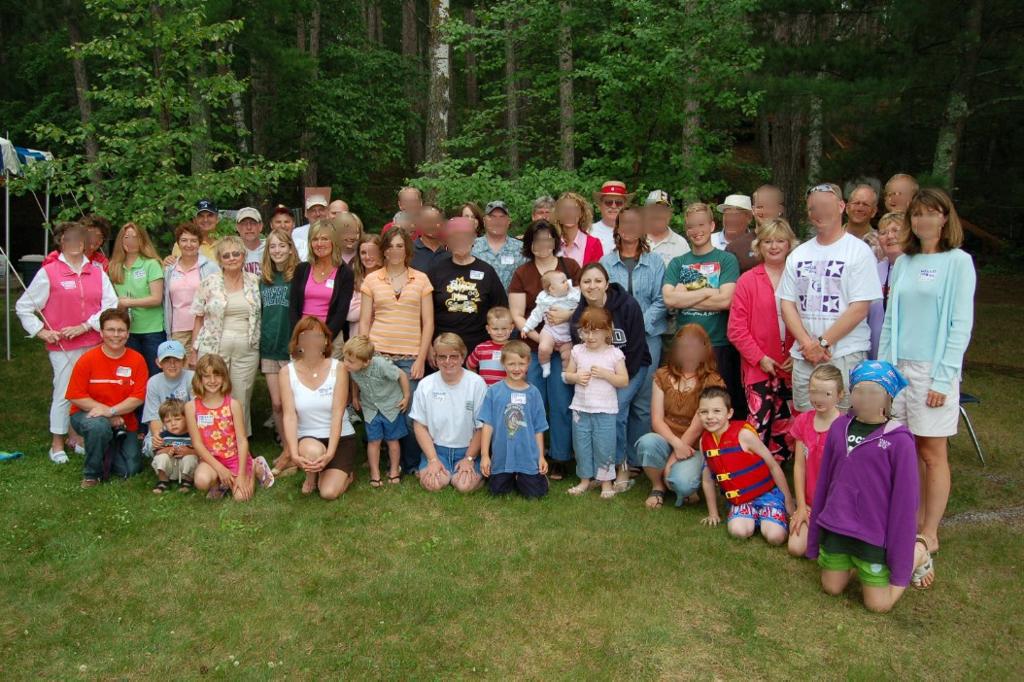}
    \includegraphics[width=0.24\linewidth]{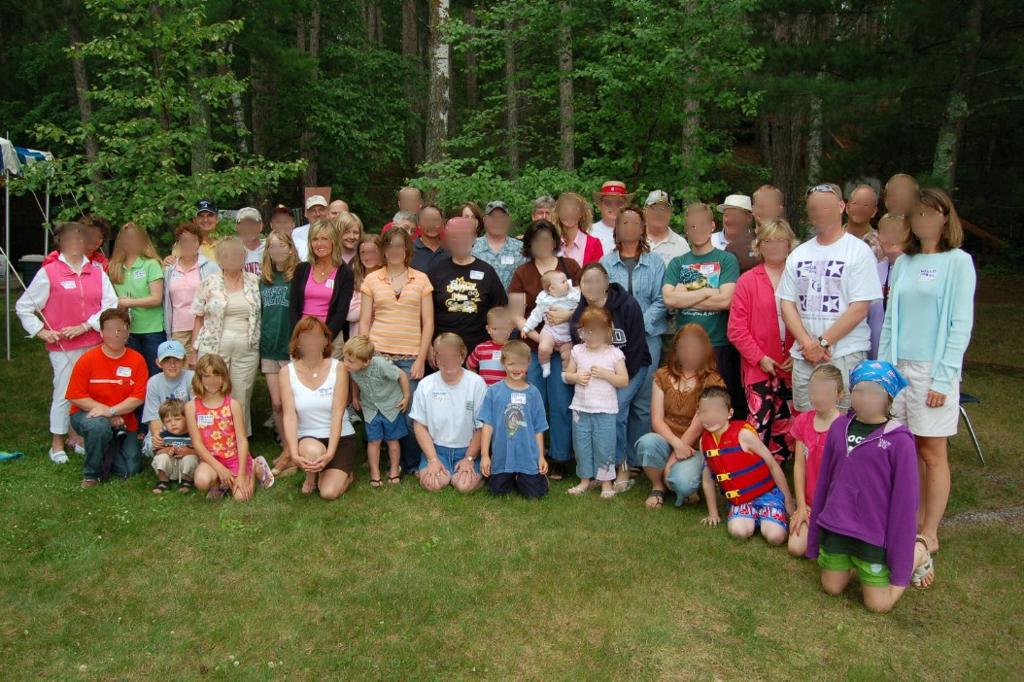}
    \includegraphics[width=0.24\linewidth]{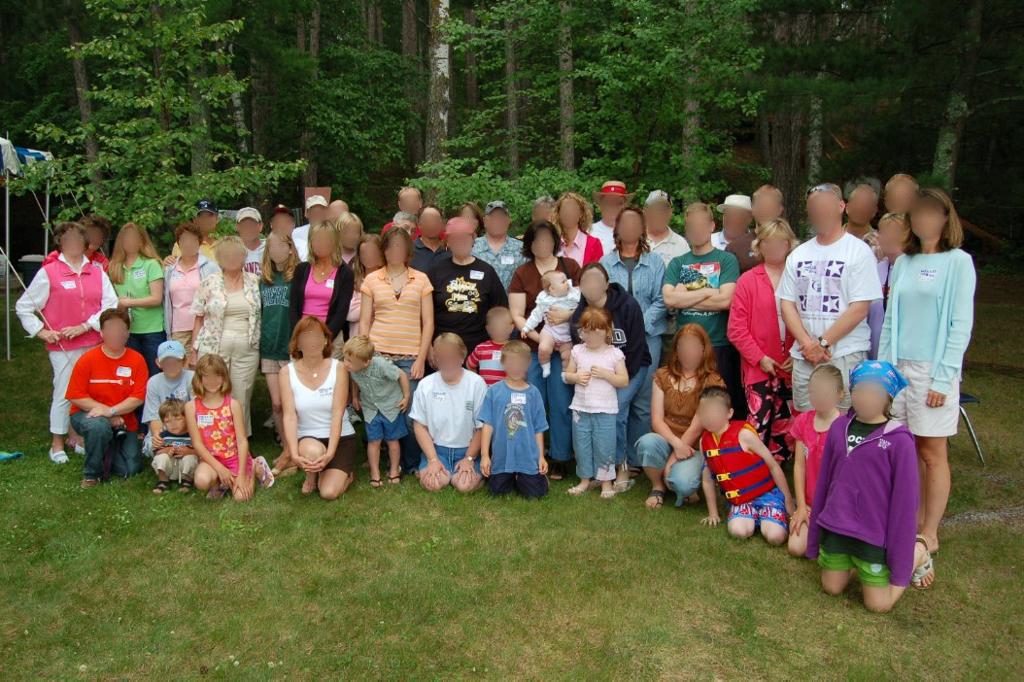}

    \caption{Test set results. Original image (left), result for $192\times192$ inference dimension (center left), result for $256\times256$ inference dimension (center right), result for $512\times512$ inference dimension (right)}
    \label{fig:dimension}
\end{figure}

\noindent
\textbf{Visual evaluation}\\
We show in Figure~\ref{fig:dimension} results obtained for three downsampling dimensions from left to right : $192\times192$, $256\times256$ and $512\times512$

We observe several phenomena.
\begin{itemize}
    \item When a face is big, meaning it covers a great percentage of the whole image (as in the two first images of Figure~\ref{fig:dimension}), small downsampling dimensions seem more fitted. In fact, we see in the first image, that for dimensions $192\times192$ and $256\times256$ faces are completely blurred whereas we can distinguish some face features for dimension $512\times512$ (although, the face still remains unrecognizable). Then, on such images, small downsampling dimension seem better fitted.
    \item When faces are smaller, the option that uses dimensions $192\times192$ starts to struggle a lot to blur faces. First, it does not succeed in blurring the face of a person with glasses (third image) and also misses a lot of faces in the two last images. On the other hand, the option with $512\times512$ dimensions almost never misses a face, and completely blurs it. The last image is blatant, the $192\times192$ option misses more than half of the faces, $256\times256$ misses a quarter of them, whereas $512\times512$ only misses the three smallest faces.
\end{itemize}
Table~\ref{tab:dimension} displays the count of correctly blurred faces in each of the images used in the tests. 

\begin{table}[htbp]
  \centering
\resizebox{\linewidth}{!}{
  \begin{tabular}{c|c|c| c |c|c}
    &\multirow{2}{*}{Number of faces} & \multicolumn{3}{|c|}{Number of faces blurred} & \multirow{2}{*}{Best visual result}\\
    & & $192\times192$ & $256\times256$ & $512\times512$ & \\
    \hline\hline
    image 1  & 1 & \textbf{1} & \textbf{1} & \textbf{1}& $192\times192$\\
    \hline
    image 2 & 1 &  \textbf{1} & \textbf{1} & \textbf{1} &$192\times192$ \\
    \hline\hline
    image 3 & 5 &  4& \textbf{5} & \textbf{5}  &$512\times512$ \\
    \hline
    image 4 & 6 & \textbf{6} & \textbf{6} & \textbf{6} & $512\times512$\textbf{}\\
    \hline\hline
    image 5 & 18 & 13 & \textbf{18} & \textbf{18} & $512\times512$\\
    \hline
    image 6 & 51 & 20 & 38 & \textbf{48} & \textbf{}$512\times512$
  \end{tabular}
}
  \caption{Face counting results (the best result for each image is in bold).}
  \label{tab:dimension}
\end{table}

We see that, depending on the input image, the optimal downsampling size can be different. In the code and online demo associated with this publication (see below), one can vary the input image size.\\
Still, overall the $512\times512$ downsampling size seem better in most cases, surpassing all other dimensions when it comes to blur small faces and still performs face blurring of big faces, even if it is visually a little worse than with smaller dimensions. However, a higher downsampling size comes also with a higher computation time. That is what we are going to briefly discuss in the next section.\\

\noindent
\textbf{Computation Time}.\\
We perform inference using a GPU (NVIDIA GeForce RTX 2080 Ti with 11GB VRAM) in all our tests.
We conduct three experiments:
\begin{itemize}
    \item One experiment assuming that input images have already been downsampled to the desired input size of the network (either 192 x 192, 256 x 256 or 512 x 512). In this case the downsampling, binary mask extraction and upsampling steps described in Figure~\ref{fig:schema} do not need to be performed.
    \item One experiment assuming that input images are of size $1024\times1024$ (we follow the algorithm described in Figure~\ref{fig:schema})
    \item One experiment assuming that input images are of size $2048\times2048$ (we follow the algorithm described in Figure~\ref{fig:schema}).
\end{itemize}

We calculate the computation times for $100$ images and average them for each experiment. Results are displayed in Table~\ref{tab:fps} and expressed in frames per second.

\begin{table}[htbp]
  \centering
\resizebox{\linewidth}{!}{
  \begin{tabular}{c|c|c|c}
   & $192\times192$ & $256\times256$ & $512\times512$ \\
  \hline\hline
  No resizing needed & $28.4$ & $22.0$ & $9.6$ \\
  \hline
  Input size of $1024\times1024$ & $9.2$ & $8.3$ & $5.4$\\
  \hline
  Input size of $2048\times2048$ & $3.0$ & $2.8$ & $2.4$
  \end{tabular}
}
  \caption{Number of frames per second processed  by the algorithm in function of the downsampling size.}
  \label{tab:fps}
\end{table}

We observe that all the resizing steps (see Figure~\ref{fig:schema}) greatly deteriorate the algorithm performance. To avoid this phenomenon, we could take advantage of multiprocessing using several cores. We did not implement this idea, but as images are processed independently, we think it could greatly improve performances. We also observed that the bigger the input image is, the more similar are all computation times.

\subsubsection{Influence of the loss function}
As in the previous section, we use a visual evaluation of the results to assess the influence of the loss function. We compare two models, one which was trained using $MSE$ loss and another using L1 loss. For both models, we use $512\times512$ as inference downsampling size.

\begin{figure}[h]
    \centering
    \includegraphics[width=0.24\linewidth]{figures/test_set/original_images/image_0.jpg}
    \includegraphics[width=0.24\linewidth]{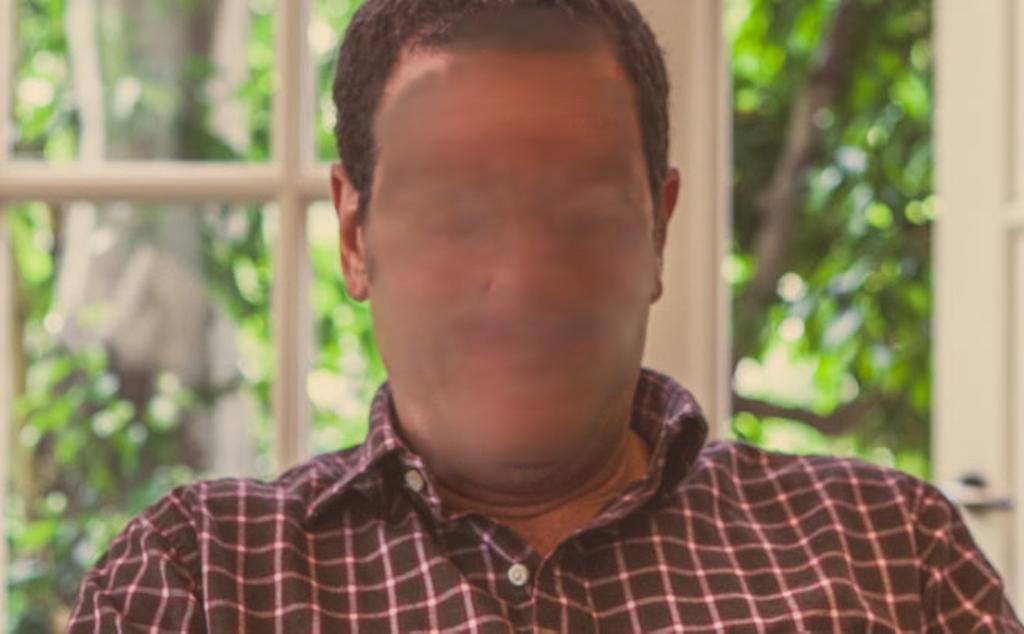}
    \includegraphics[width=0.24\linewidth]{figures/test_set/l1_512/image_0.jpg}\\
    
    \includegraphics[width=0.24\linewidth]{figures/test_set/original_images/image_25.jpg}
    \includegraphics[width=0.24\linewidth]{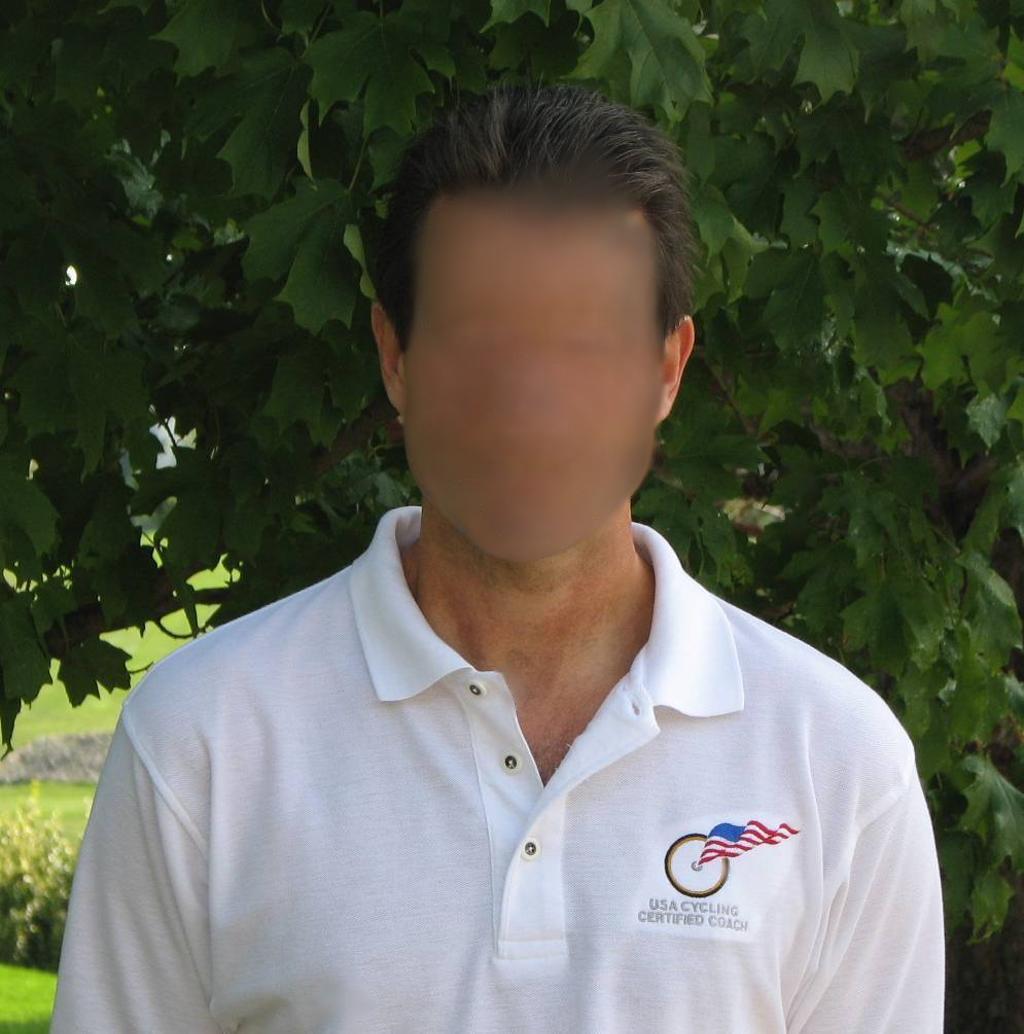}
    \includegraphics[width=0.24\linewidth]{figures/test_set/l1_512/image_25.jpg}\\
    
    \includegraphics[width=0.24\linewidth]{figures/test_set/original_images/image_1.jpg}
    \includegraphics[width=0.24\linewidth]{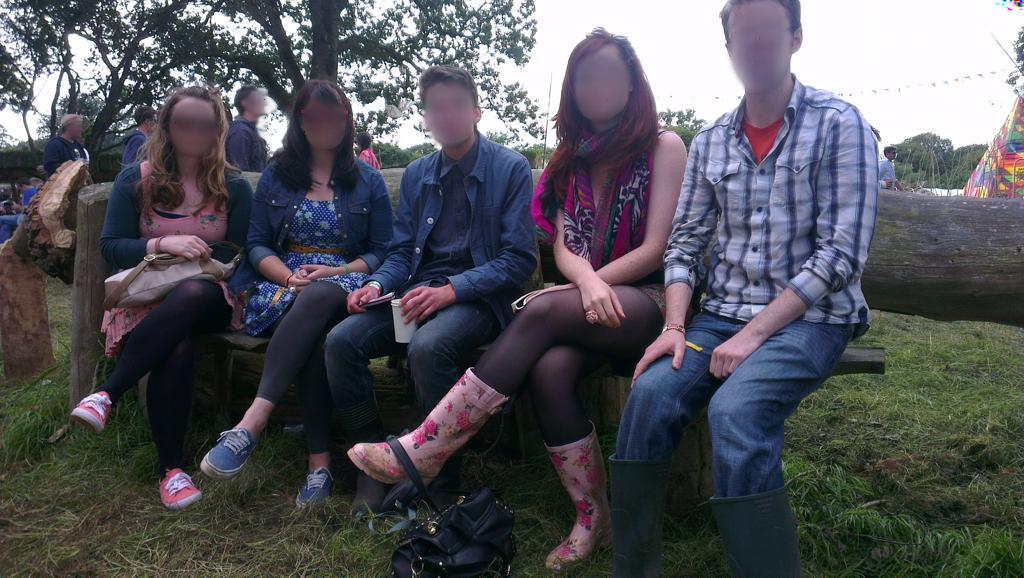}
    \includegraphics[width=0.24\linewidth]{figures/test_set/l1_512/image_1.jpg}\\

    \includegraphics[width=0.24\linewidth]{figures/test_set/original_images/image_16.jpg}
    \includegraphics[width=0.24\linewidth]{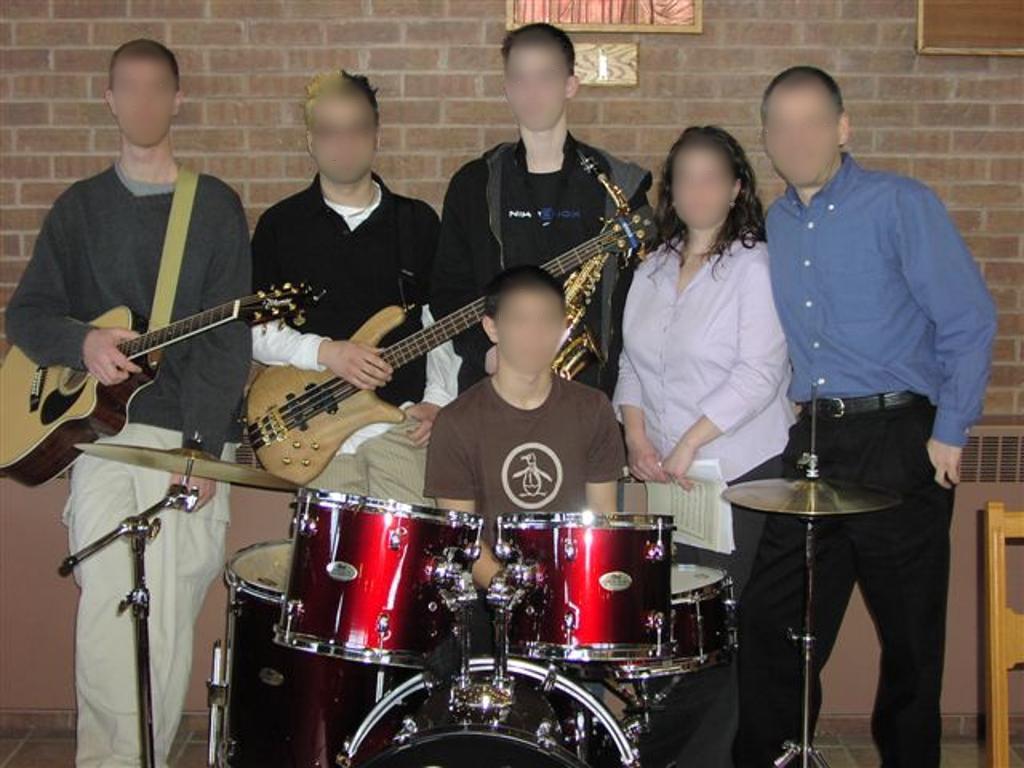}
    \includegraphics[width=0.24\linewidth]{figures/test_set/l1_512/image_16.jpg} \\

    \includegraphics[width=0.24\linewidth]{figures/test_set/original_images/image_57.jpg}
    \includegraphics[width=0.24\linewidth]{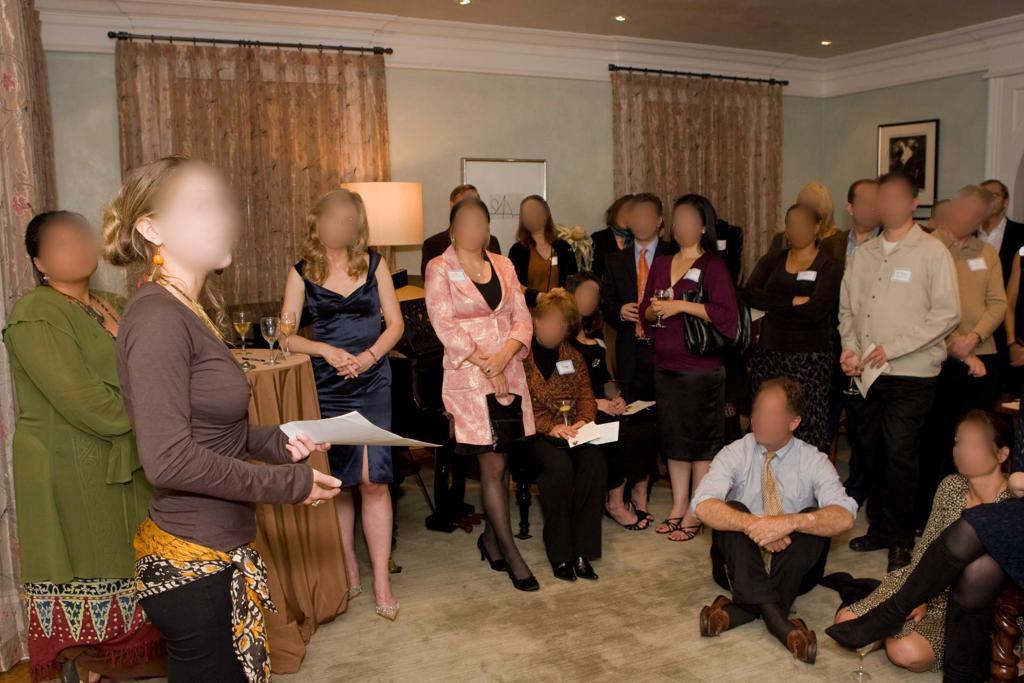}
    \includegraphics[width=0.24\linewidth]{figures/test_set/l1_512/image_57.jpg}\\
    
    \includegraphics[width=0.24\linewidth]{figures/test_set/original_images/image_24.jpg}
    \includegraphics[width=0.24\linewidth]{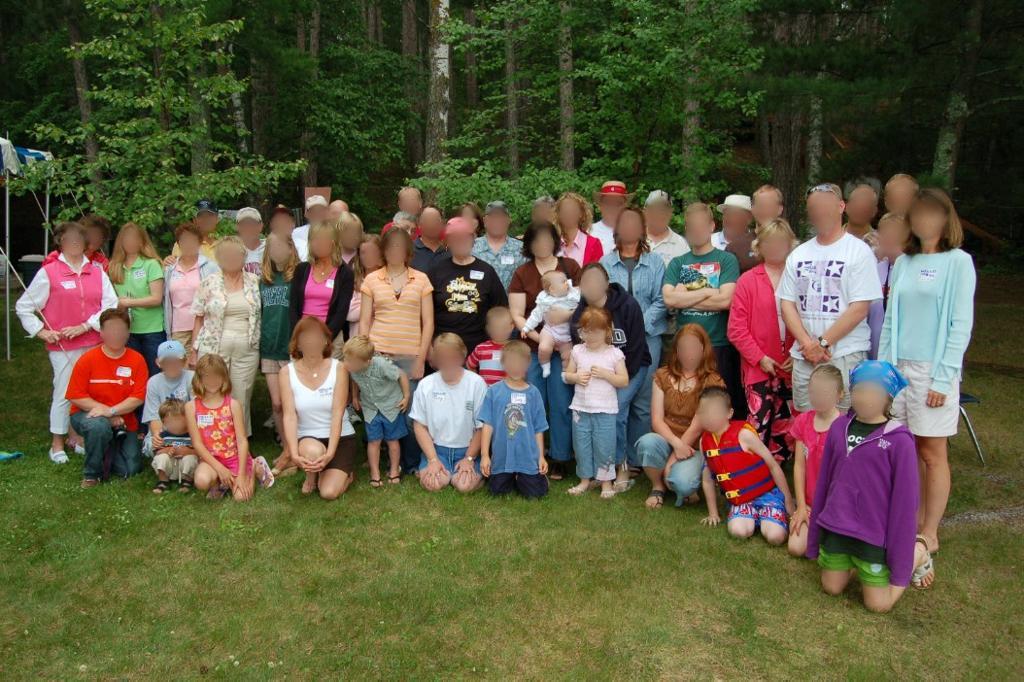}
    \includegraphics[width=0.24\linewidth]{figures/test_set/l1_512/image_24.jpg}

    \caption{Test set results. Original image (left), result for $MSE$ loss (center), result for L1 loss (right). Downsampling size $=~512\times512$}
    \label{fig:loss}
\end{figure}

\begin{figure}[h]
    \centering
    \includegraphics[width=0.24\linewidth]{figures/test_set/original_images/image_0.jpg}
    \includegraphics[width=0.24\linewidth]{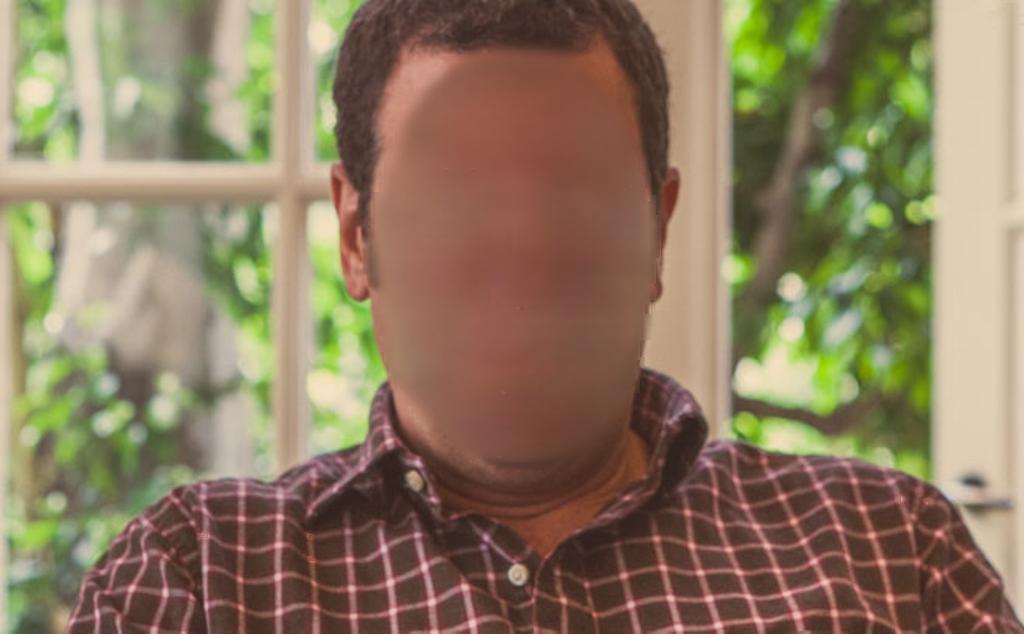}
    \includegraphics[width=0.24\linewidth]{figures/test_set/l1_192/image_0.jpg}\\
    
    \includegraphics[width=0.24\linewidth]{figures/test_set/original_images/image_25.jpg}
    \includegraphics[width=0.24\linewidth]{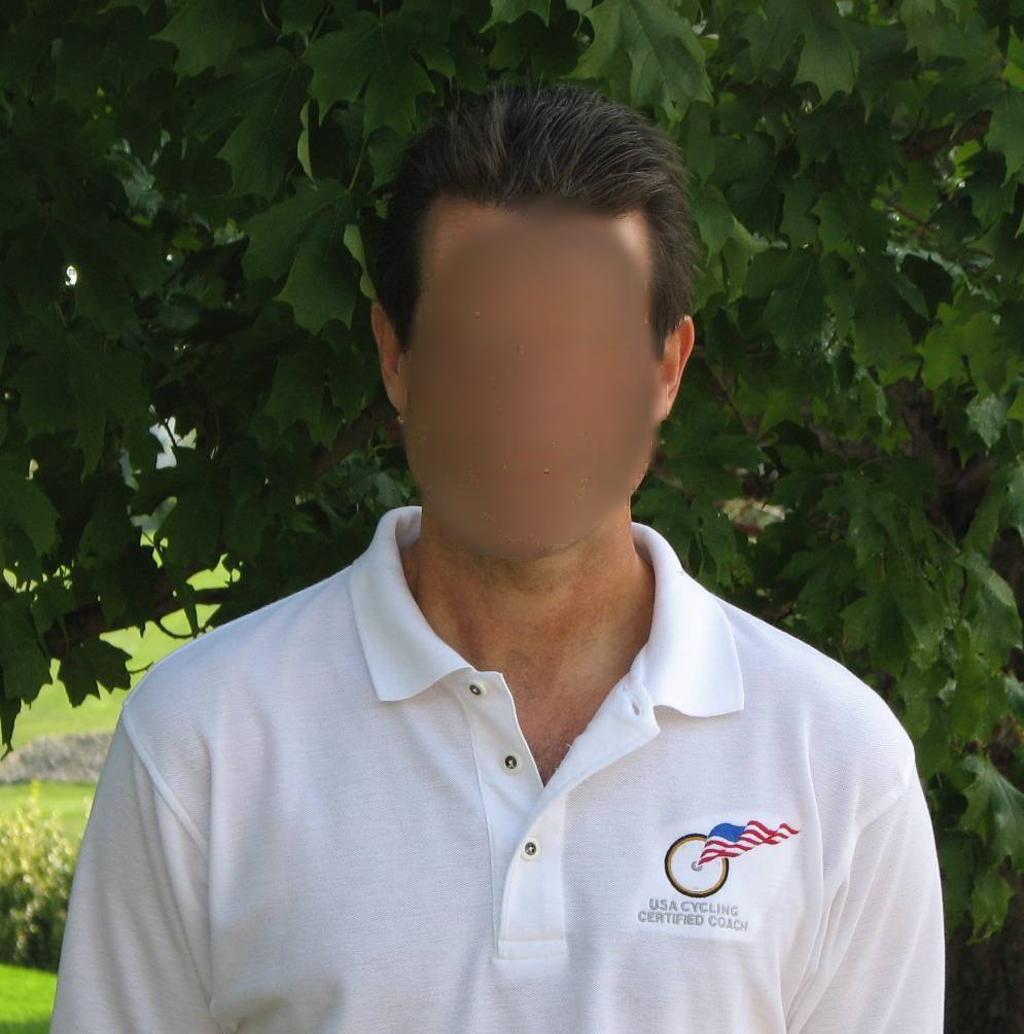}
    \includegraphics[width=0.24\linewidth]{figures/test_set/l1_192/image_25.jpg}\\
    
    \includegraphics[width=0.24\linewidth]{figures/test_set/original_images/image_1.jpg}
    \includegraphics[width=0.24\linewidth]{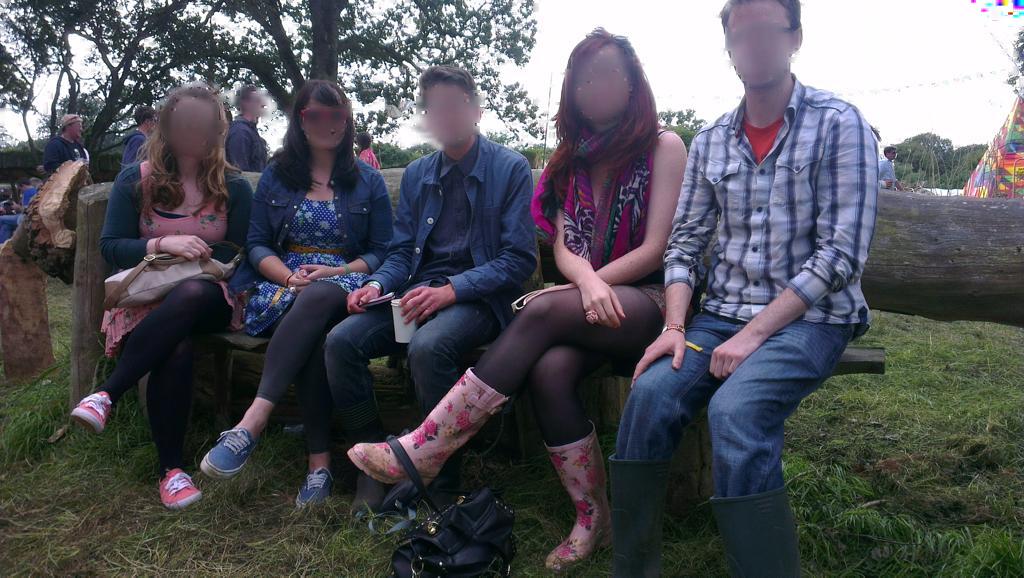}
    \includegraphics[width=0.24\linewidth]{figures/test_set/l1_192/image_1.jpg}\\

    \includegraphics[width=0.24\linewidth]{figures/test_set/original_images/image_16.jpg}
    \includegraphics[width=0.24\linewidth]{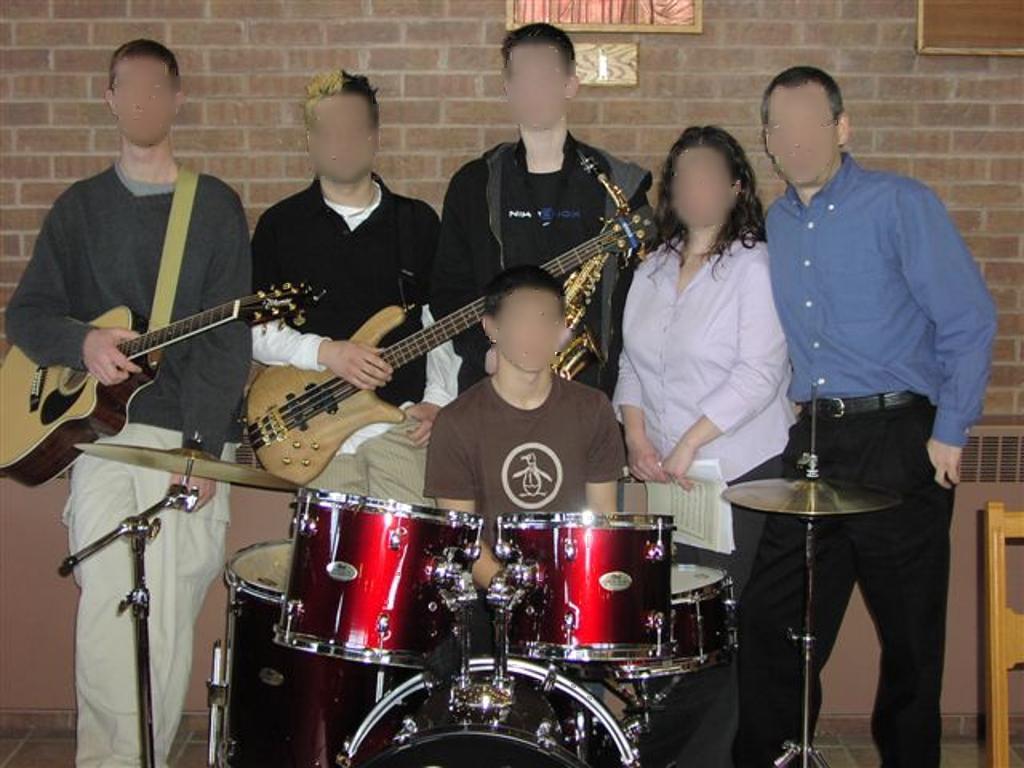}
    \includegraphics[width=0.24\linewidth]{figures/test_set/l1_192/image_16.jpg} \\

    \includegraphics[width=0.24\linewidth]{figures/test_set/original_images/image_57.jpg}
    \includegraphics[width=0.24\linewidth]{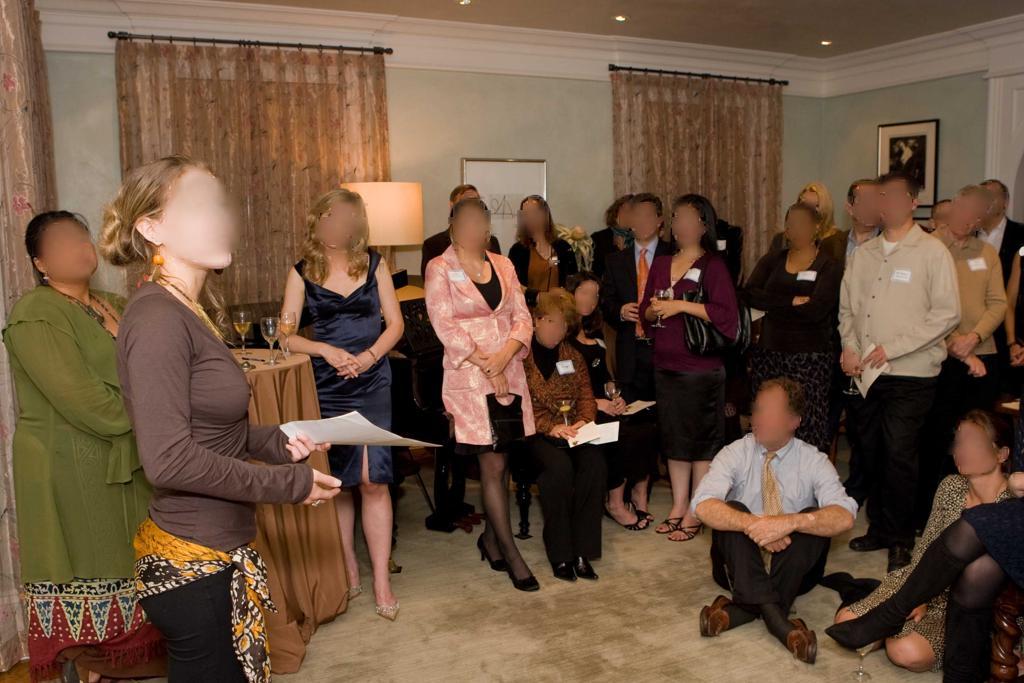}
    \includegraphics[width=0.24\linewidth]{figures/test_set/l1_192/image_57.jpg}\\
    
    \includegraphics[width=0.24\linewidth]{figures/test_set/original_images/image_24.jpg}
    \includegraphics[width=0.24\linewidth]{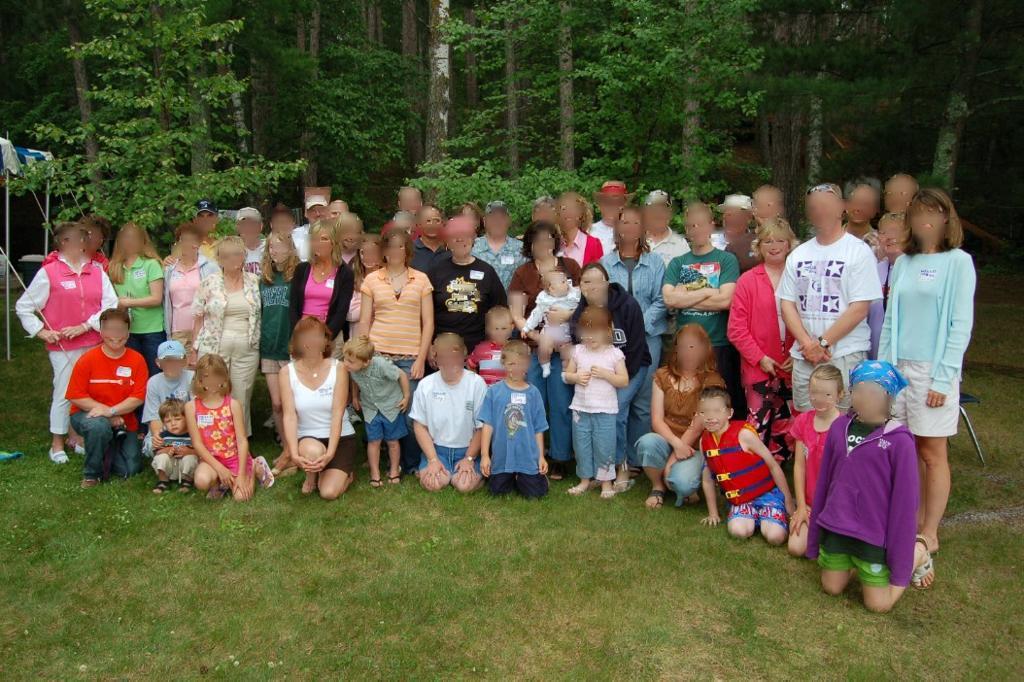}
    \includegraphics[width=0.24\linewidth]{figures/test_set/l1_192/image_24.jpg}

    \caption{Test set results. Original image (left), result for $MSE$ loss (center), result for L1 loss (right). Downsampling size $=~192\times192$}
    \label{fig:loss_192}
\end{figure}

What stands out in Figure~\ref{fig:loss} is that results are very similar. At first sight, we struggle to distinguish any difference. But in fact, there are some. First if we look closer a at ``big faces'' images (the first two ones), we can see less face features in these images: face blurring is better performed with $MSE$. Then, for images 3, 4 and 5, results are almost the same. Finally, we see in the last image, that the $MSE$ trained network succeeds in blurring the tiniest faces that the L1 trained did not blur. That is a already a strong improvement : $MSE$ tends to provide better results for small face than L1.\\
To be able to assert and verify this previous statement, we investigate the performance when considering an even smaller downsampling size. In that framework, faces appear to cover less pixels, turning them into smaller faces for the network. We consider a downsampling size of $192\times192$ and display the results for both losses in Figure~\ref{fig:loss_192}. We observe, as expected, that the $MSE$ trained network detects and blurs the great majority of faces in all the images while the L1 trained network misses some of them in images 3, 5 and 6. The most impressive results are for image 5 in which the $MSE$ trained network detects and blurs 5 more faces than the L1 trained network. Same for image 6 in which it detects and blurs 25 more faces. This shows that $MSE$ is more sensitive to small faces while performing as well as L1 loss on bigger faces. We sum up in Table~\ref{tab:loss} the blurred face counting.

\begin{table}[!h]
  \centering
\resizebox{\linewidth}{!}{
  \begin{tabular}{c|c|c|c|c|c|}
    & \multirow{3}{*}{Number of faces} & \multicolumn{4}{|c|}{Number of faces blurred}\\
    & & \multicolumn{2}{|c|}{$MSE$} & \multicolumn{2}{|c|}{L1} \\
    & & $192\times192$ & $512\times512$ & $192\times192$ & $512\times512$ \\
    \hline\hline
    image 1 & 1 & \textbf{1} & \textbf{1} & \textbf{1} & \textbf{1}\\
    \hline
    image 2 & 1 & \textbf{1} & \textbf{1} & \textbf{1} & \textbf{1}\\
    \hline\hline
    image 3 & 5 & \textbf{5} & \textbf{5} & 4 & \textbf{5} \\
    \hline
    image 4 & 6 & \textbf{6} & \textbf{6} & \textbf{6} & \textbf{6} \\
    \hline\hline
    image 5 & 18 & \textbf{18} & \textbf{18} & 13 & \textbf{18}\\
    \hline
    image 6 & 51 & 45 & \textbf{50} & 20 & 48\\
  \end{tabular}
}
\caption{Face counting results (the best results for each image is in bold)}
  \label{tab:loss}
\end{table}

We do not compare the computation times in this section as there are exactly the same as in the previous section. In fact, models have the exact same number of parameters and the pre and postprocessing steps are identical.

\subsubsection{Influence of the self-attention layer}
In this section, we study the importance of the self-attention layer and if it is relevant in our case. 
In \cite{deoldifyIPOL}, the introduction of this layer is motivated by the long-time studied question of non-locality in images. Paper \cite{nlmeans} was one of the first articles to propose a non-local algorithm in the context of denoising. This seminal work proved that to denoise a given patch of an image, all other patches of the images, provided that they are similar to the original patch, contain information relevant to denoise the patch. The self-attention mechanism is a similar process, thus more general. In the context of colorizing, this layer appears to be relevant in order to combine information, and in particular to propagate colorization to patches that are similar.\\
In our context, we hypothesize that this could permit to combine information from faces all over the image and improve the face blurring performance. \\

\begin{figure}[h]
    \centering
    \includegraphics[width=0.24\linewidth]{figures/test_set/original_images/image_0.jpg}
    \includegraphics[width=0.24\linewidth]{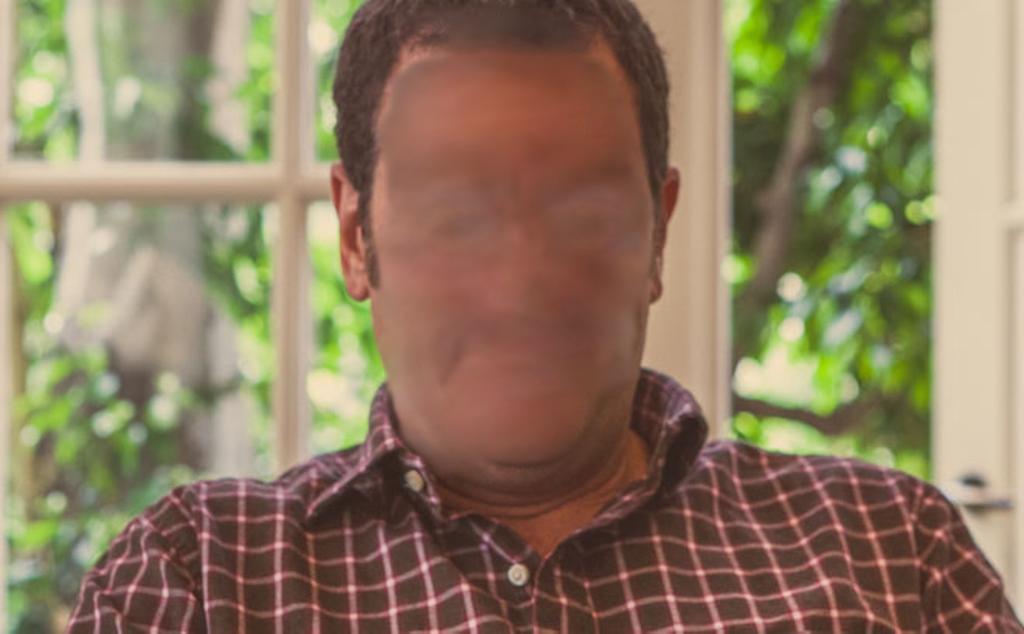}
    \includegraphics[width=0.24\linewidth]{figures/test_set/mse_512/image_0.jpg}\\
    
    \includegraphics[width=0.24\linewidth]{figures/test_set/original_images/image_25.jpg}
    \includegraphics[width=0.24\linewidth]{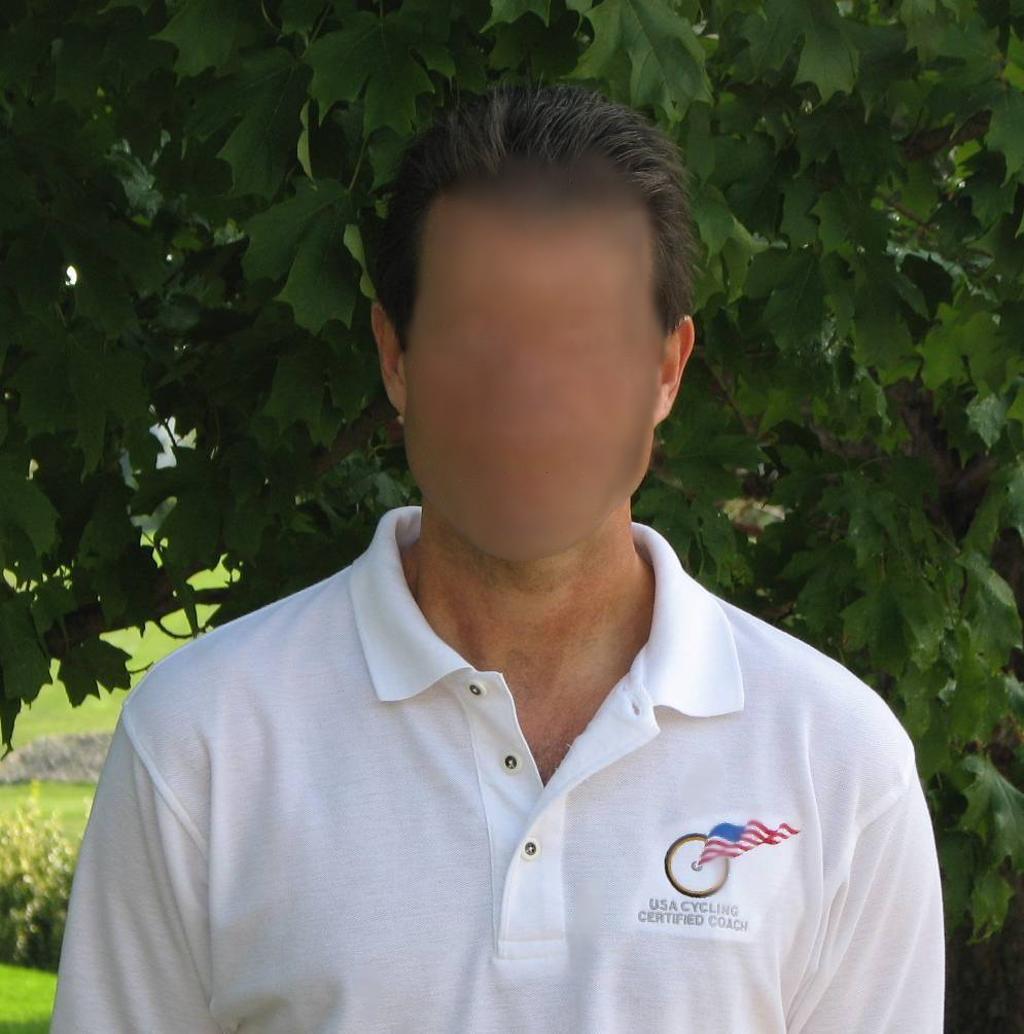}
    \includegraphics[width=0.24\linewidth]{figures/test_set/mse_512/image_25.jpg}\\
    
    \includegraphics[width=0.24\linewidth]{figures/test_set/original_images/image_1.jpg}
    \includegraphics[width=0.24\linewidth]{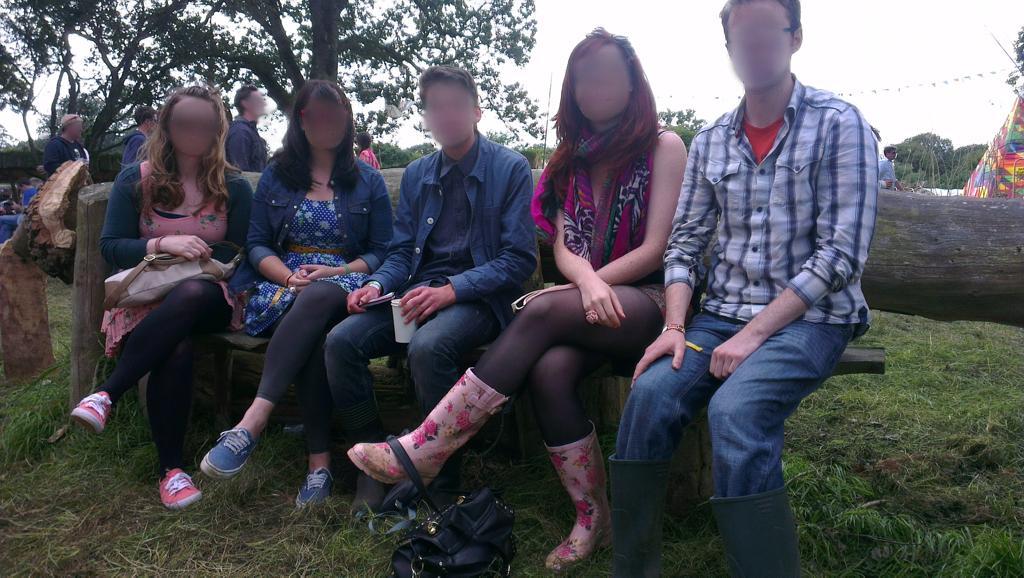}
    \includegraphics[width=0.24\linewidth]{figures/test_set/mse_512/image_1.jpg}\\

    \includegraphics[width=0.24\linewidth]{figures/test_set/original_images/image_16.jpg}
    \includegraphics[width=0.24\linewidth]{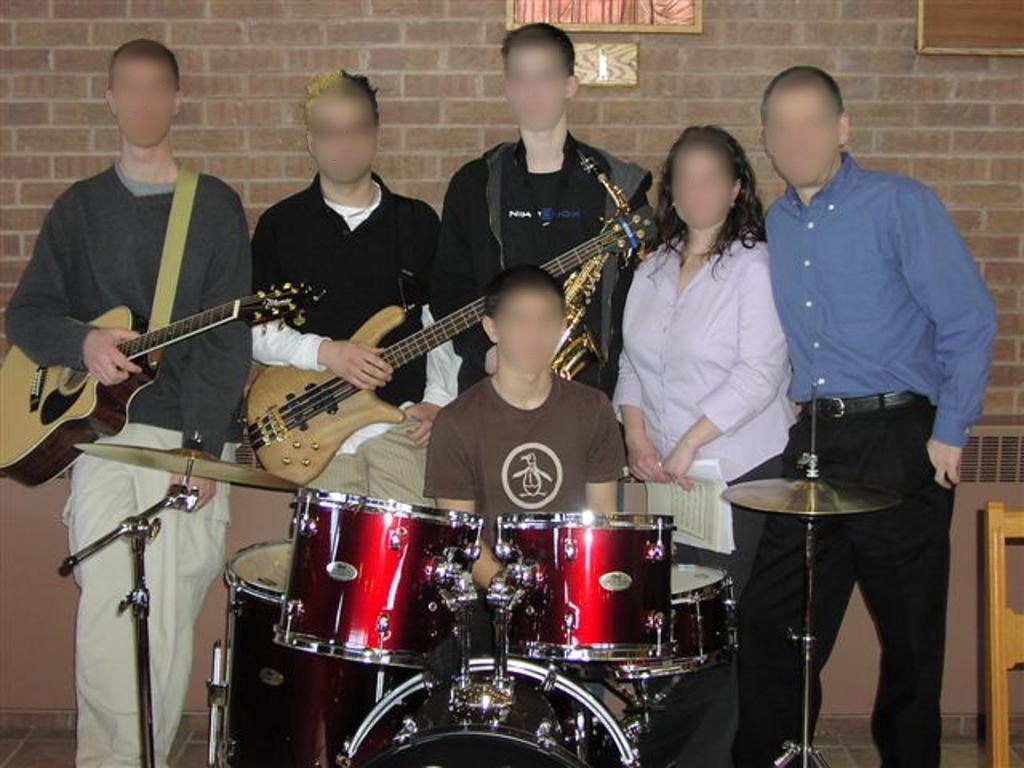}
    \includegraphics[width=0.24\linewidth]{figures/test_set/mse_512/image_16.jpg} \\

    \includegraphics[width=0.24\linewidth]{figures/test_set/original_images/image_57.jpg}
    \includegraphics[width=0.24\linewidth]{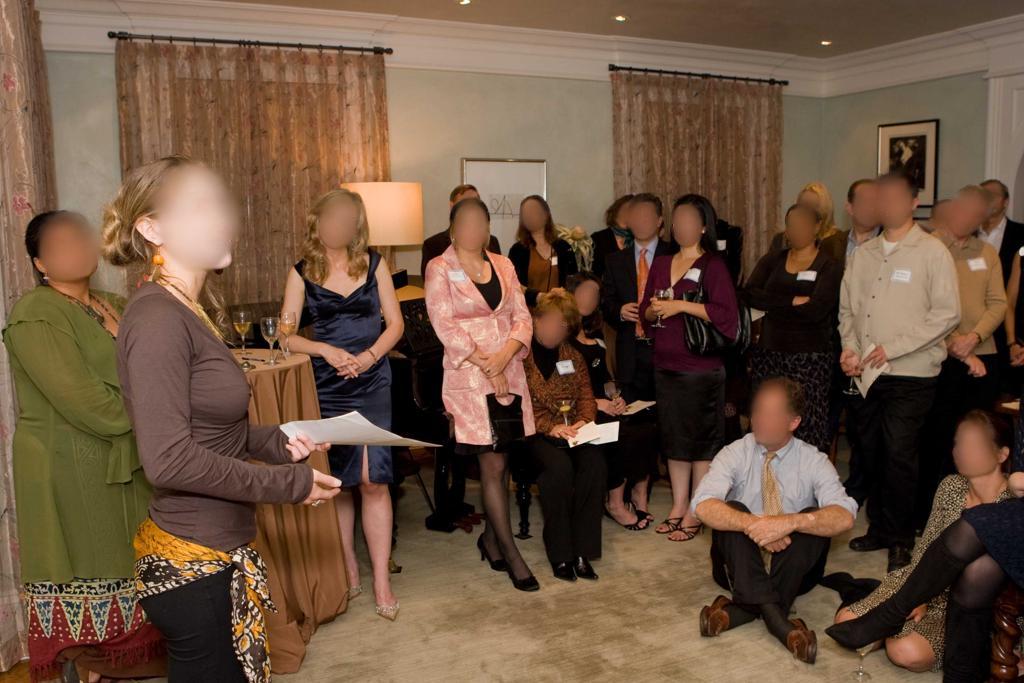}
    \includegraphics[width=0.24\linewidth]{figures/test_set/mse_512/image_57.jpg}\\
    
    \includegraphics[width=0.24\linewidth]{figures/test_set/original_images/image_24.jpg}
    \includegraphics[width=0.24\linewidth]{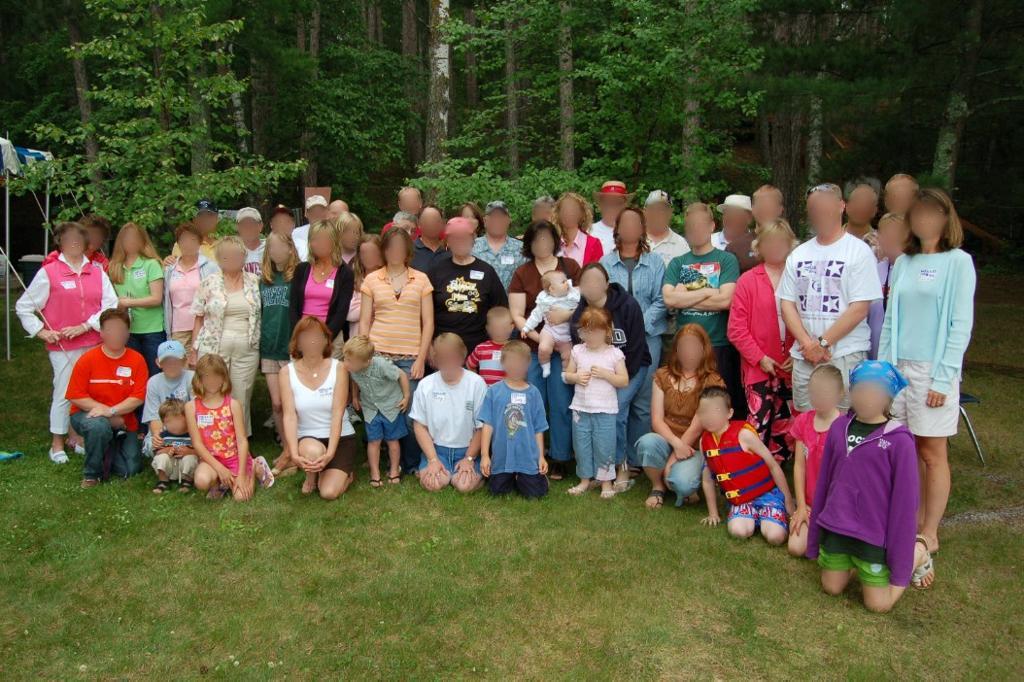}
    \includegraphics[width=0.24\linewidth]{figures/test_set/mse_512/image_24.jpg}

    \caption{Test set results. Original image (left), result without self-attention layer (center), with self-attention layer (right). Model trained with $MSE$ loss. Downsampling size $512\times 512$.}
    \label{fig:selfa}
\end{figure}

\noindent
\textbf{Visual evaluation}\\
In Figure~\ref{fig:selfa}, we compare the results of the $MSE$ trained network, with a downsampling size of $512\times512$ with and without the self-attention layer.

\begin{figure}[h]
    \centering
    \includegraphics[width=0.24\linewidth]{figures/test_set/original_images/image_0.jpg}
    \includegraphics[width=0.24\linewidth]{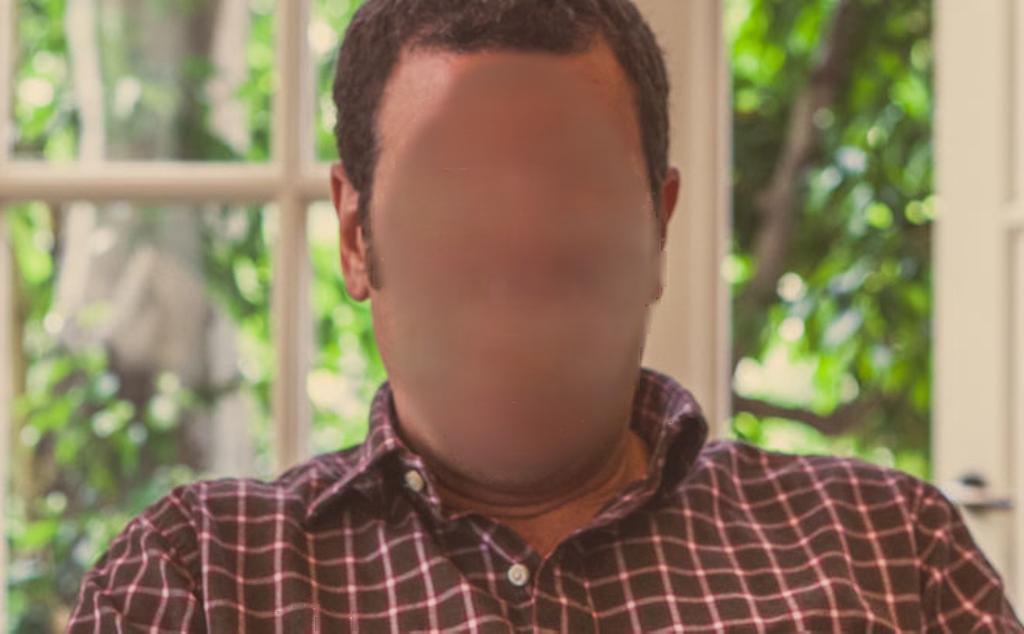}
    \includegraphics[width=0.24\linewidth]{figures/test_set/mse_192/image_0.jpg}\\
    
    \includegraphics[width=0.24\linewidth]{figures/test_set/original_images/image_25.jpg}
    \includegraphics[width=0.24\linewidth]{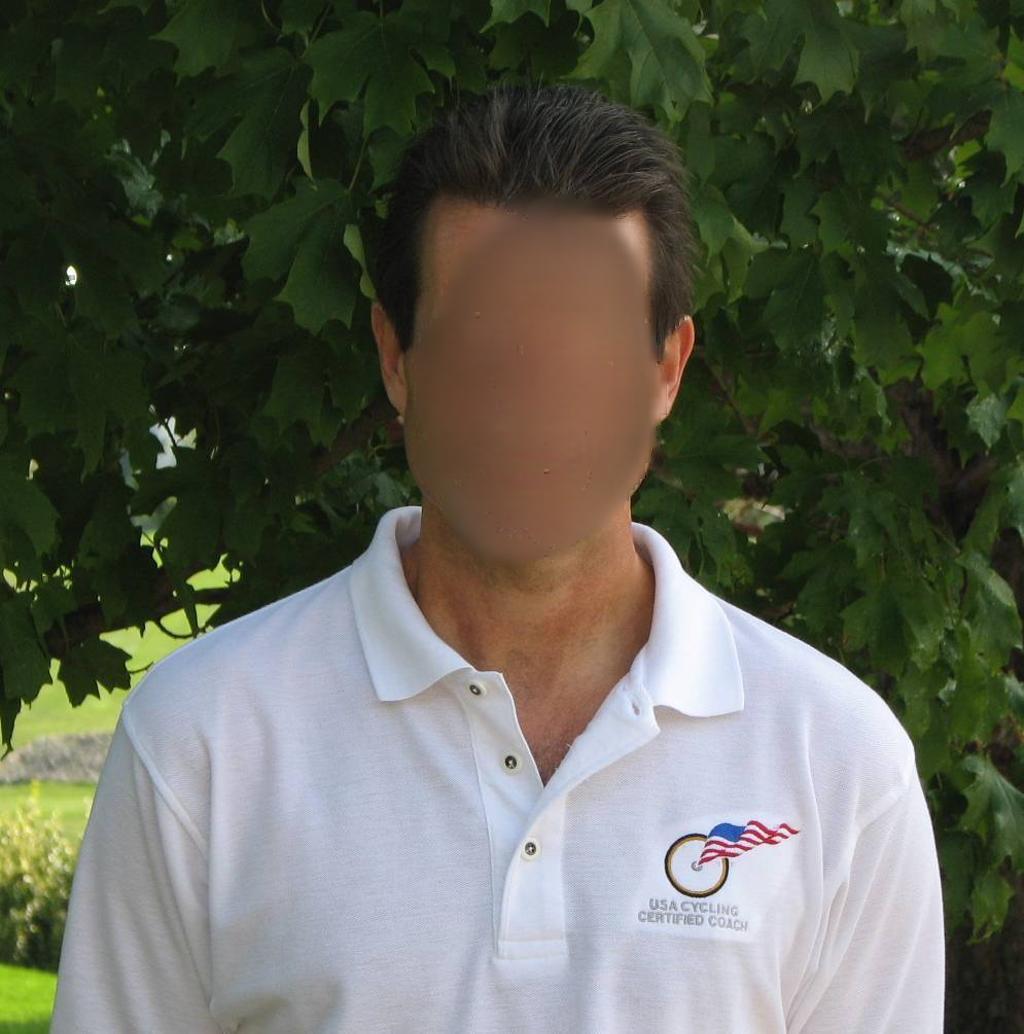}
    \includegraphics[width=0.24\linewidth]{figures/test_set/mse_192/image_25.jpg}\\
    
    \includegraphics[width=0.24\linewidth]{figures/test_set/original_images/image_1.jpg}
    \includegraphics[width=0.24\linewidth]{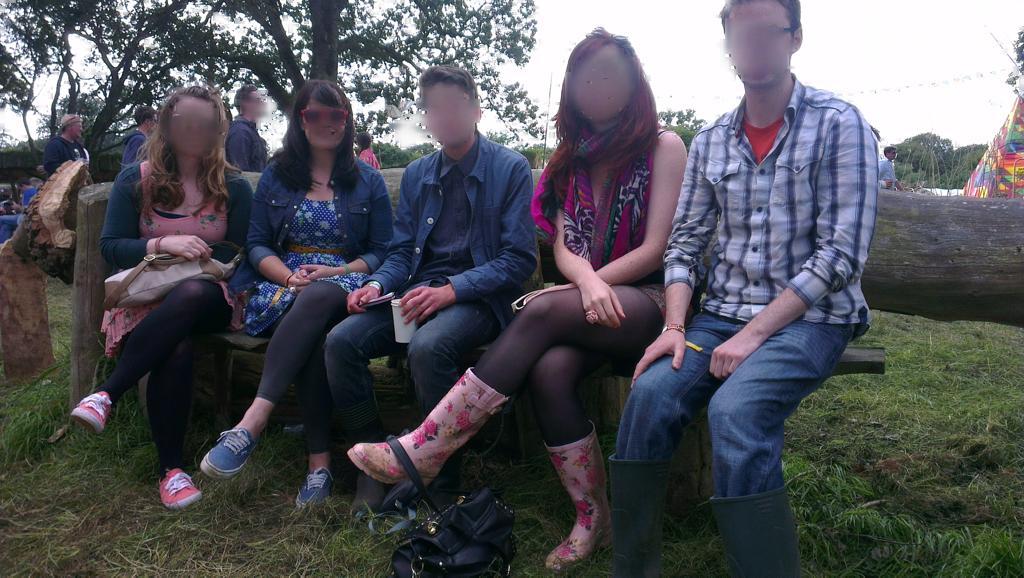}
    \includegraphics[width=0.24\linewidth]{figures/test_set/mse_192/image_1.jpg}\\

    \includegraphics[width=0.24\linewidth]{figures/test_set/original_images/image_16.jpg}
    \includegraphics[width=0.24\linewidth]{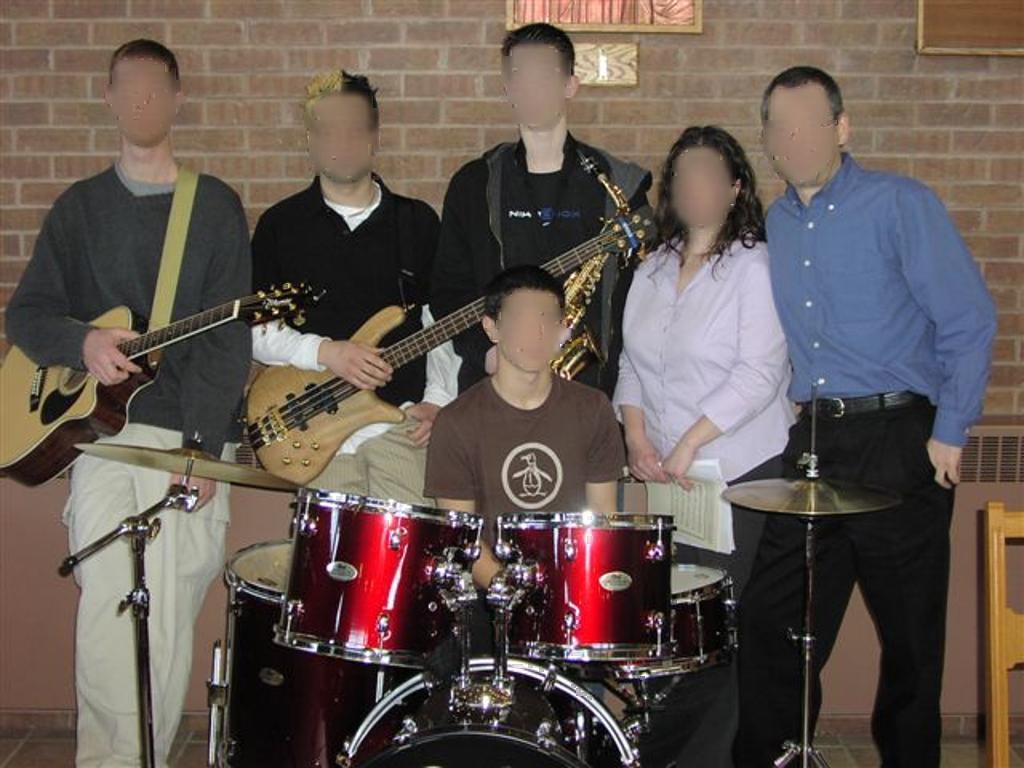}
    \includegraphics[width=0.24\linewidth]{figures/test_set/mse_192/image_16.jpg} \\

    \includegraphics[width=0.24\linewidth]{figures/test_set/original_images/image_57.jpg}
    \includegraphics[width=0.24\linewidth]{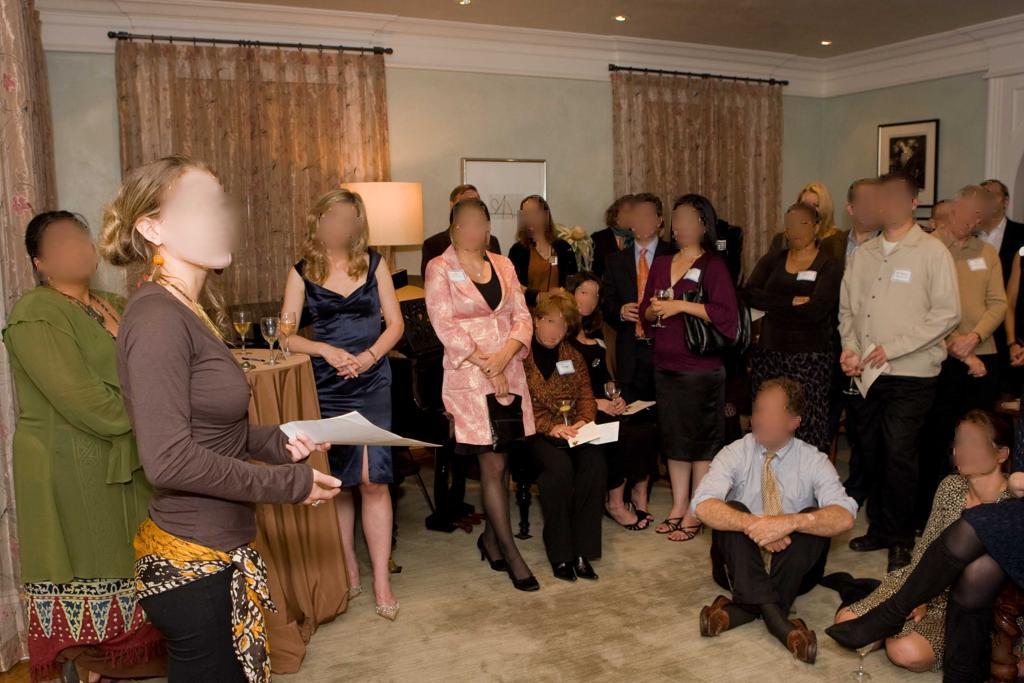}
    \includegraphics[width=0.24\linewidth]{figures/test_set/mse_192/image_57.jpg}\\
    
    \includegraphics[width=0.24\linewidth]{figures/test_set/original_images/image_24.jpg}
    \includegraphics[width=0.24\linewidth]{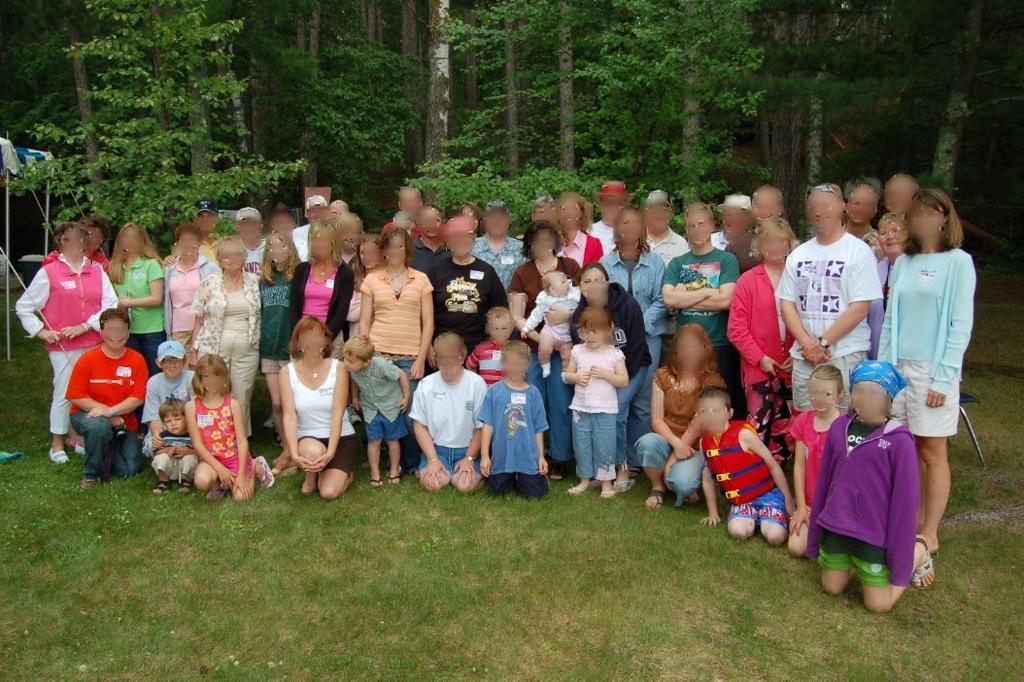}
    \includegraphics[width=0.24\linewidth]{figures/test_set/mse_192/image_24.jpg}

    \caption{Test set results. Original image (left), result without self-attention layer (center), with self-attention layer (right). Model trained with $MSE$ loss. Downsampling size $192\times192$.}
    \label{fig:selfa_192}
\end{figure}

Visually, on Figure~\ref{fig:selfa} we struggle to find differences for a downsampling size of  $512\times512$. We then display in Figure~\ref{fig:selfa_192} the results on the test set for a downsampling size of $192\times192$.

Once again, it is very hard to spot any visual difference between the two models on Figure~\ref{fig:selfa_192}. Our conclusion is that it is not relevant to add this self-attention layer to our model.\\

\noindent
\textbf{Computation time}\\
It in interesting to compare computation times with and without the self-attention layer. In fact the self-attention layer has a quadratic complexity with respect to the input dimension. To this extent, we expect that inference time with self-attention layer has a higher computation time. Here we conduct only one experiment assuming that input images are already of the right downsampling size. In fact, the only difference in computation time is when we pass through the model, pre and post processing steps are exactly the same.\\
We calculate computation times for 100 images and average them for each experiment. Results are displayed in Table~\ref{tab:fps_selfa} and expressed in frames per second.

\begin{table}[htbp]
  \centering
\resizebox{\linewidth}{!}{
  \begin{tabular}{c|c|c|c}
   & $192\times192$ & $256\times256$ & $512\times512$ \\
  \hline\hline
  With self-attention & $28.4$ & $22.0$ & $9.6$ \\
  \hline
  Without self-attention & $28.8$ & $22.6$ & $9.8$\\
  \end{tabular}
}
  \caption{Number of frames per second processed  by the algorithm with and without self-attention.}
  \label{tab:fps_selfa}
\end{table}
We conclude that the differences in computation time are negligible considering that, in practice, we have pre and post processing steps that will make the difference even tinier.
In our study, we cannot conclude that adding a self-attention layer is relevant. First, visual differences in results are imperceptible and the difference in computation time using one method or the other is negligible.

\subsubsection{Experiments using YOLOV5Face}
In this section, we evaluate the YOLO-based face blur method described in Section~\ref{sec:yolo}. We recall that the YOLOV5Face methodology consists in detecting faces and then blur them (see Section~\ref{sec:inf_meth_yolo}).\\

\begin{figure}[h]
    \centering
    \setlength{\tabcolsep}{0.1pt}    
    \begin{tabular}{ccccc}
    \includegraphics[width=0.20\linewidth]{figures/test_set/original_images/image_0.jpg} &
    \includegraphics[width=0.20\linewidth]{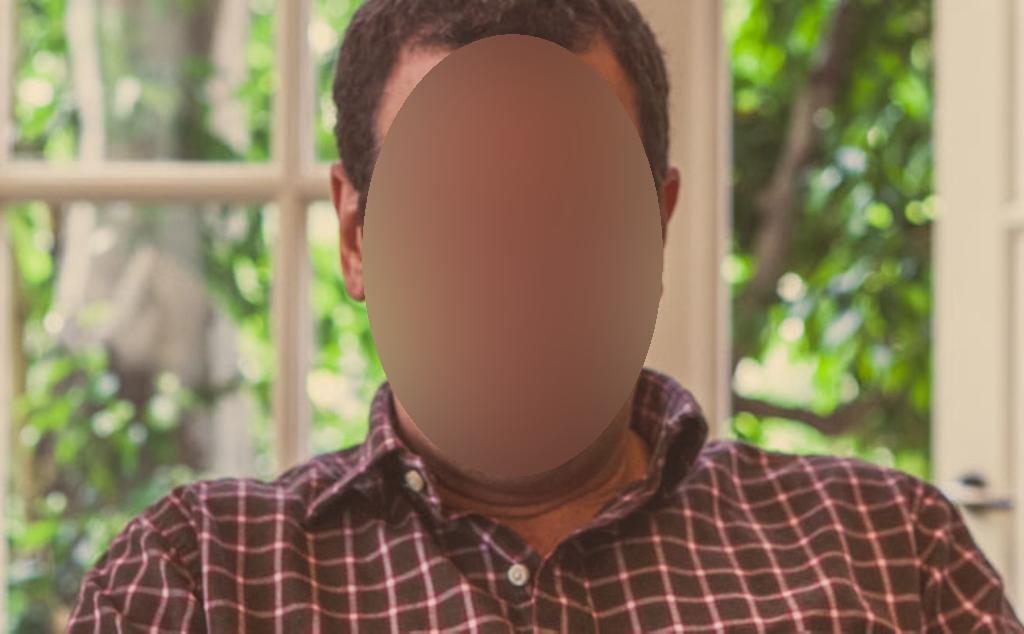} &
    \includegraphics[width=0.20\linewidth]{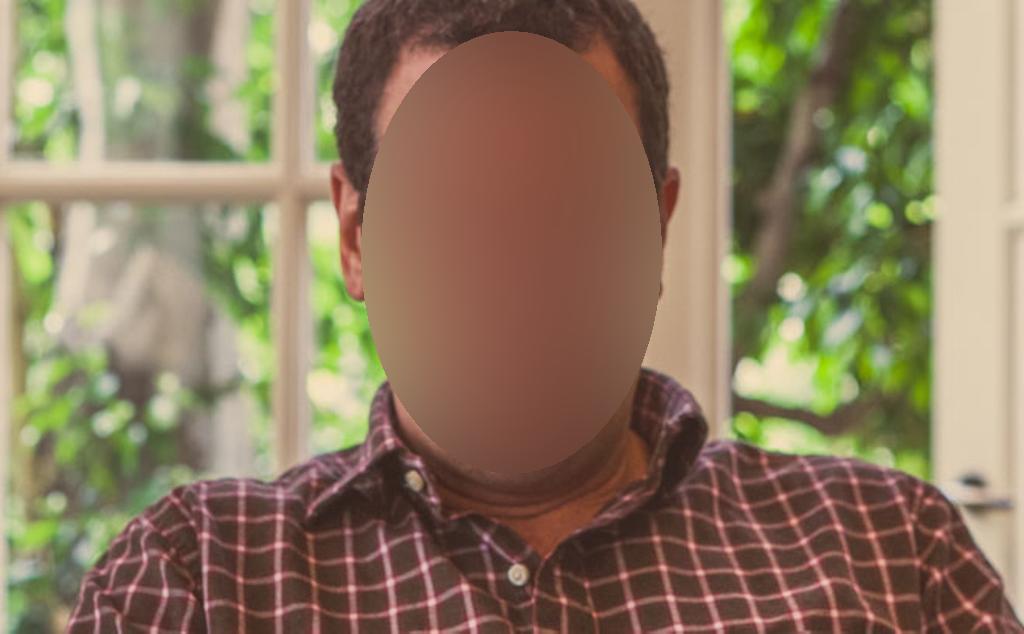} &
    \includegraphics[width=0.20\linewidth]{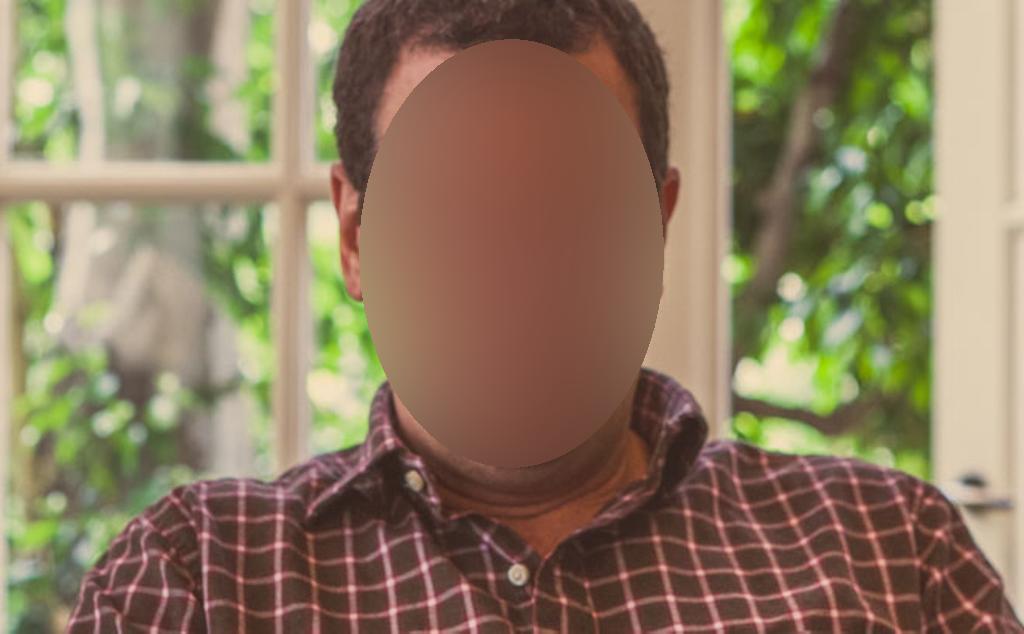} &
    \includegraphics[width=0.20\linewidth]{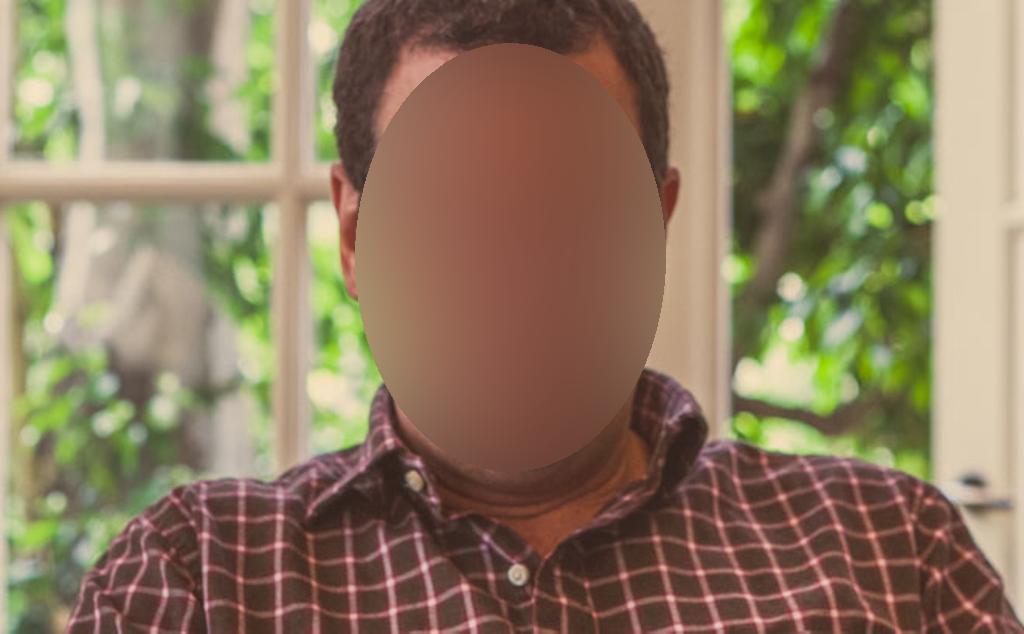}\\

    \includegraphics[width=0.20\linewidth]{figures/test_set/original_images/image_25.jpg} &
    \includegraphics[width=0.20\linewidth]{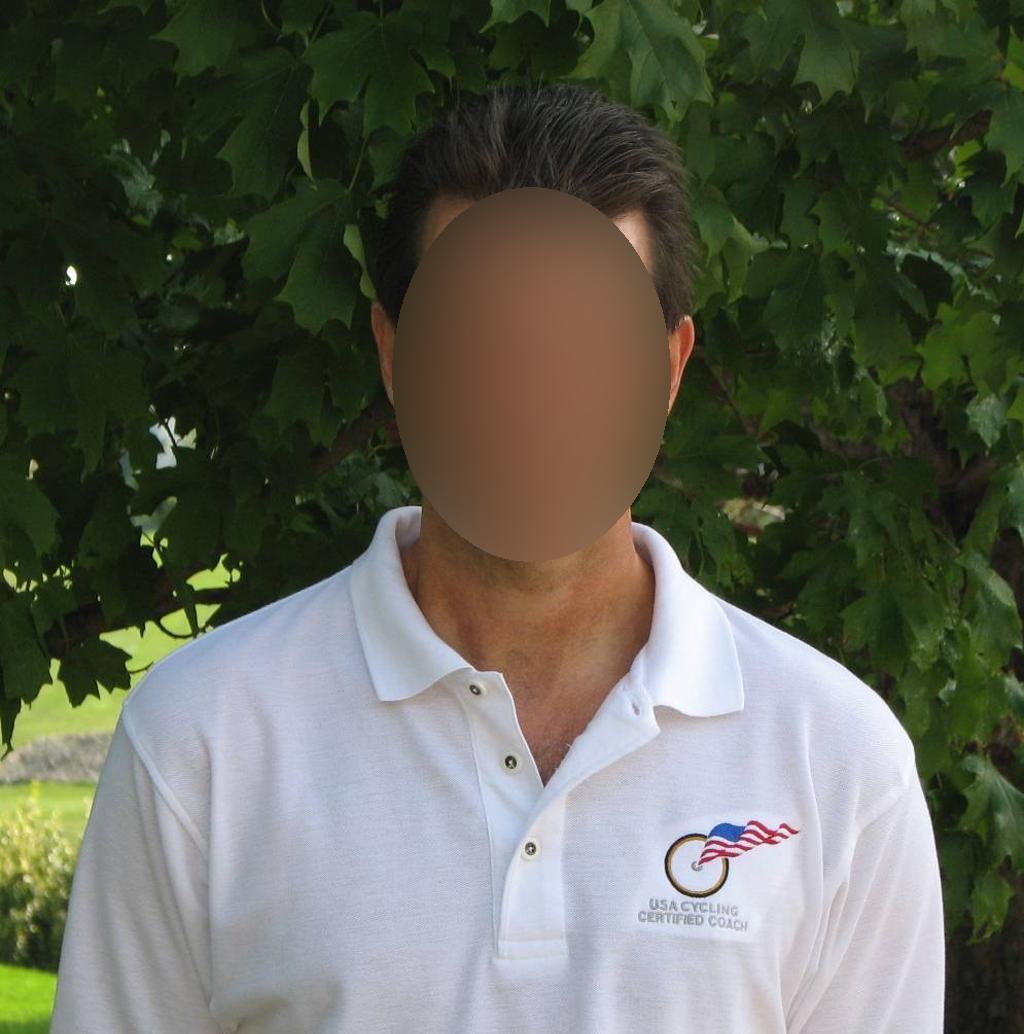} &
    \includegraphics[width=0.20\linewidth]{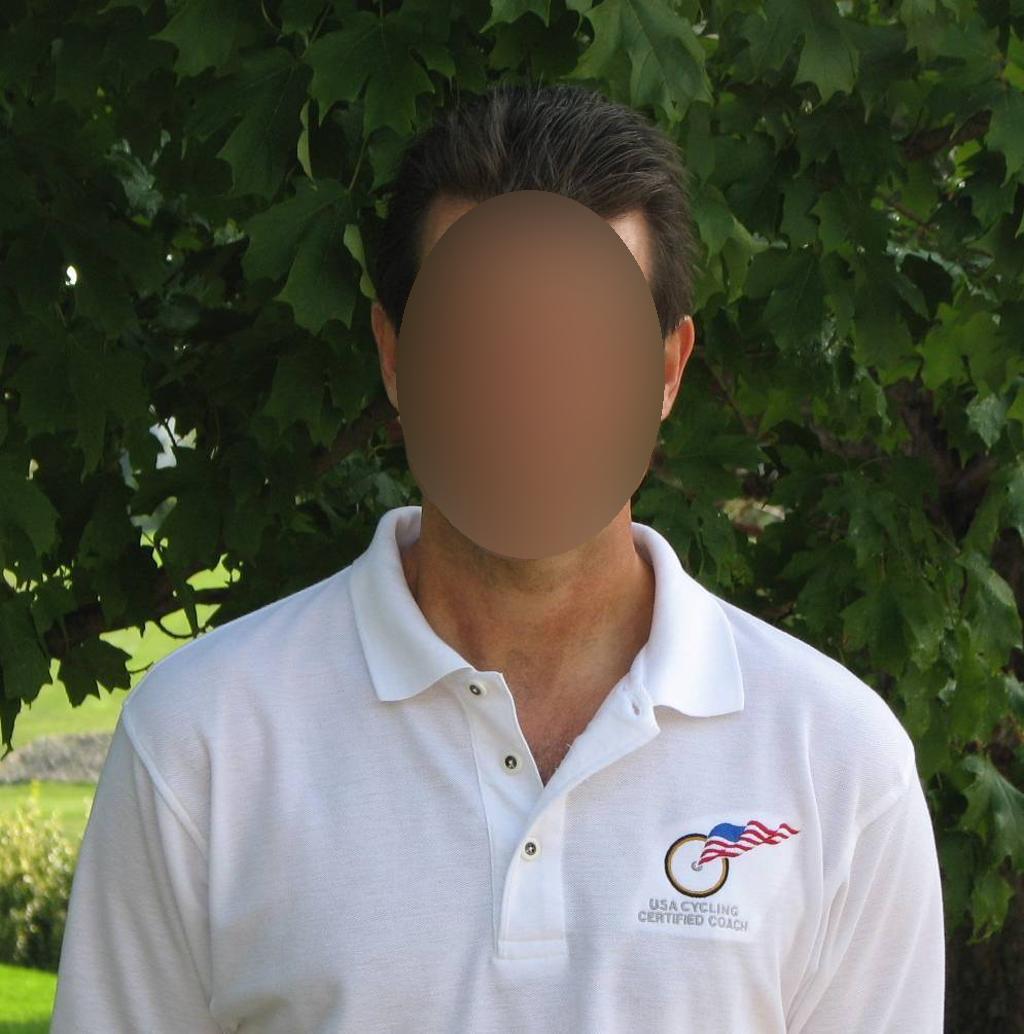} &
    \includegraphics[width=0.20\linewidth]{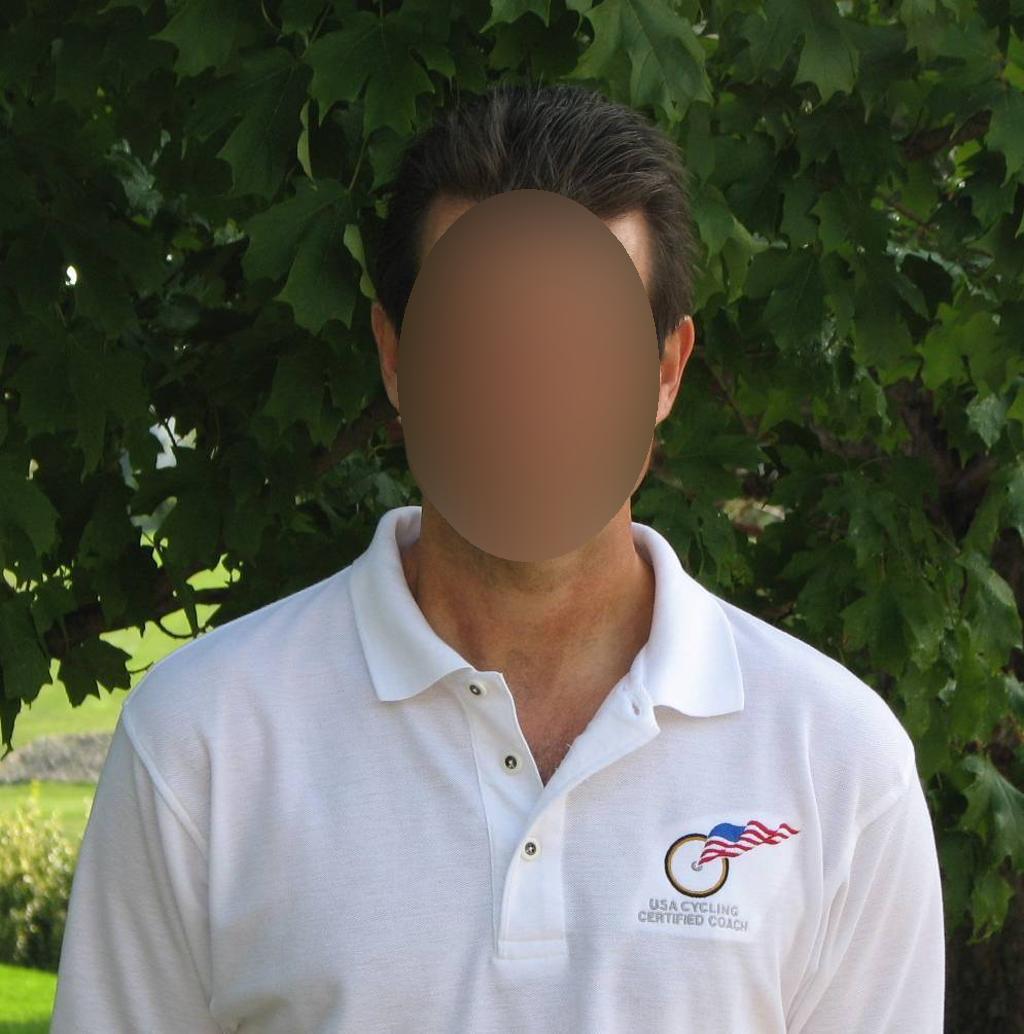} &
    \includegraphics[width=0.20\linewidth]{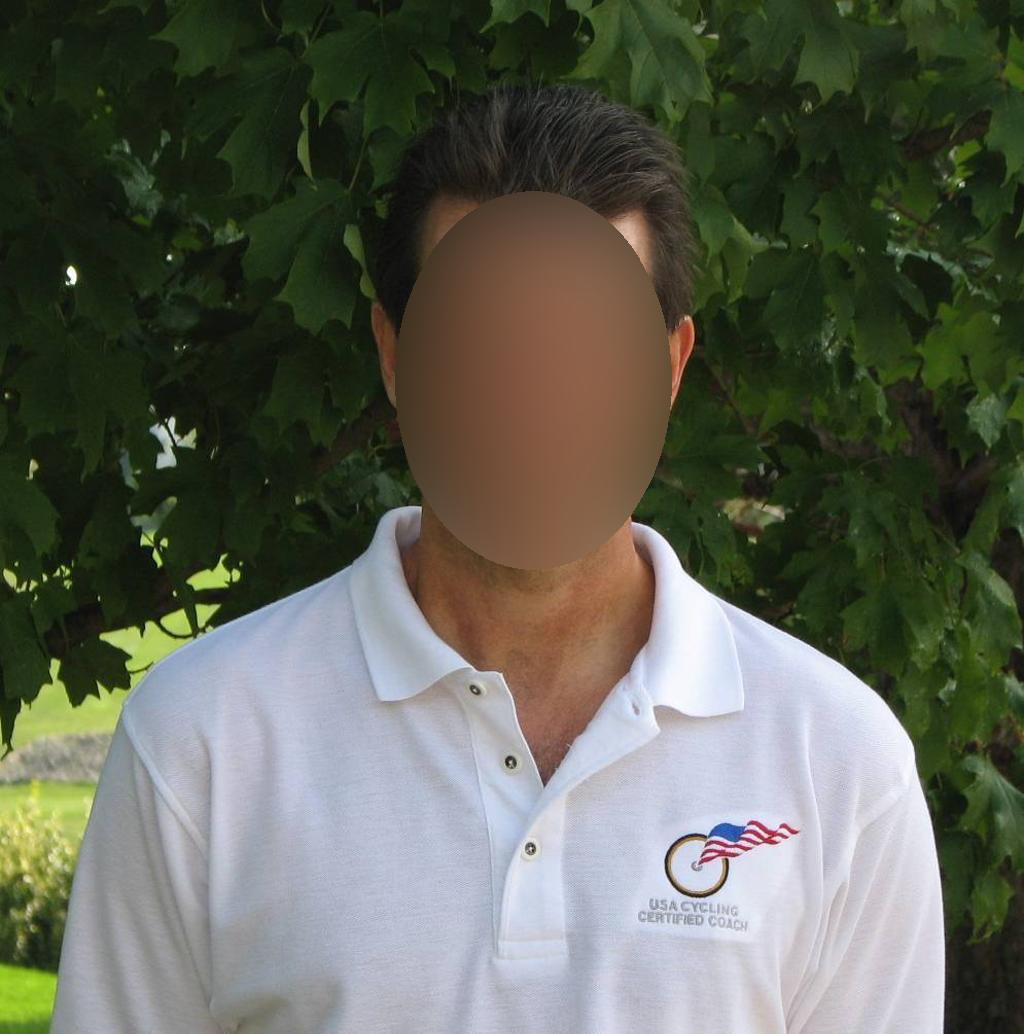}\\
    
    \includegraphics[width=0.20\linewidth]{figures/test_set/original_images/image_1.jpg} &
    \includegraphics[width=0.20\linewidth]{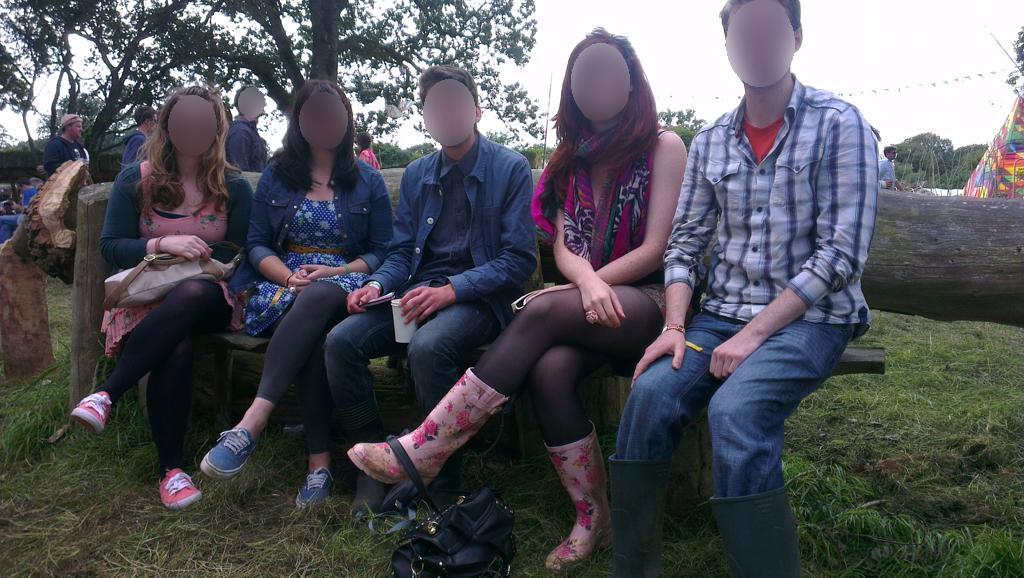} &
    \includegraphics[width=0.20\linewidth]{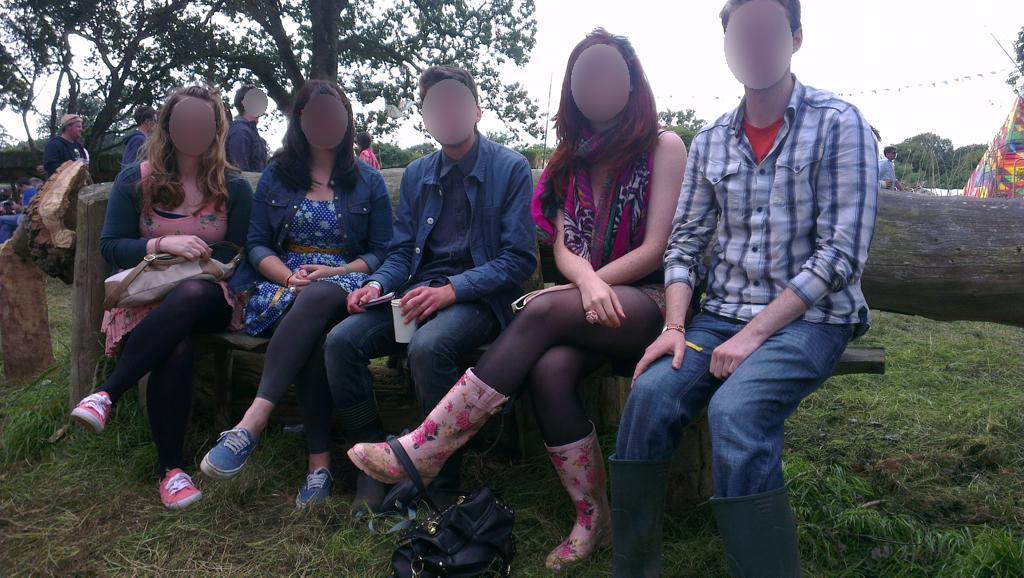} &
    \includegraphics[width=0.20\linewidth]{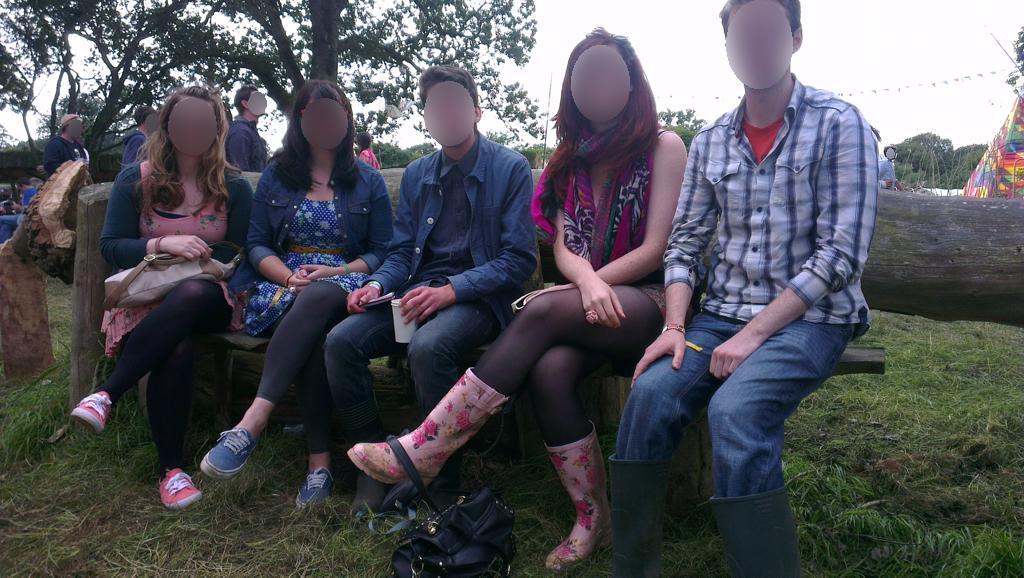} &
    \includegraphics[width=0.20\linewidth]{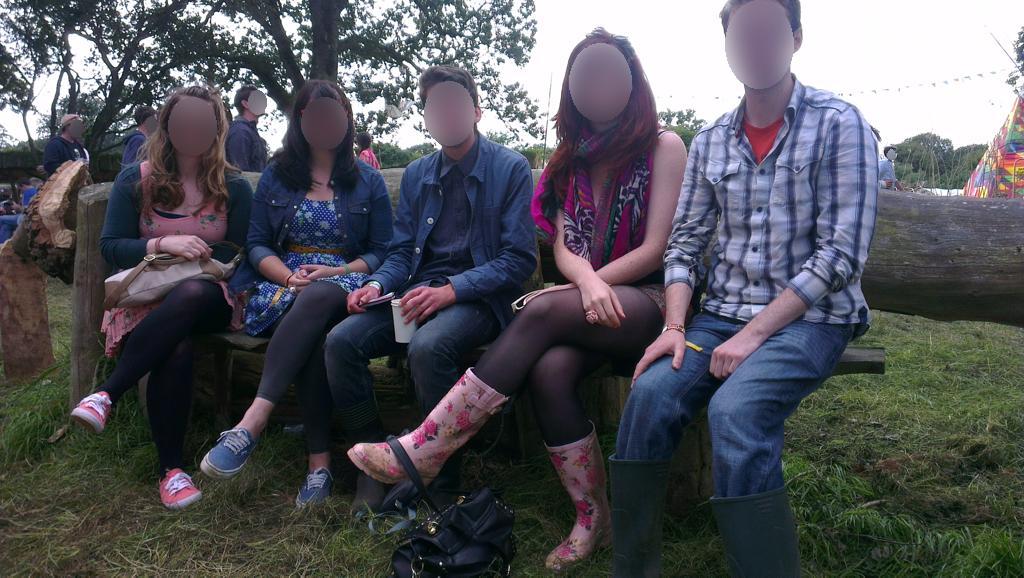}\\

    \includegraphics[width=0.20\linewidth]{figures/test_set/original_images/image_16.jpg} &
    \includegraphics[width=0.20\linewidth]{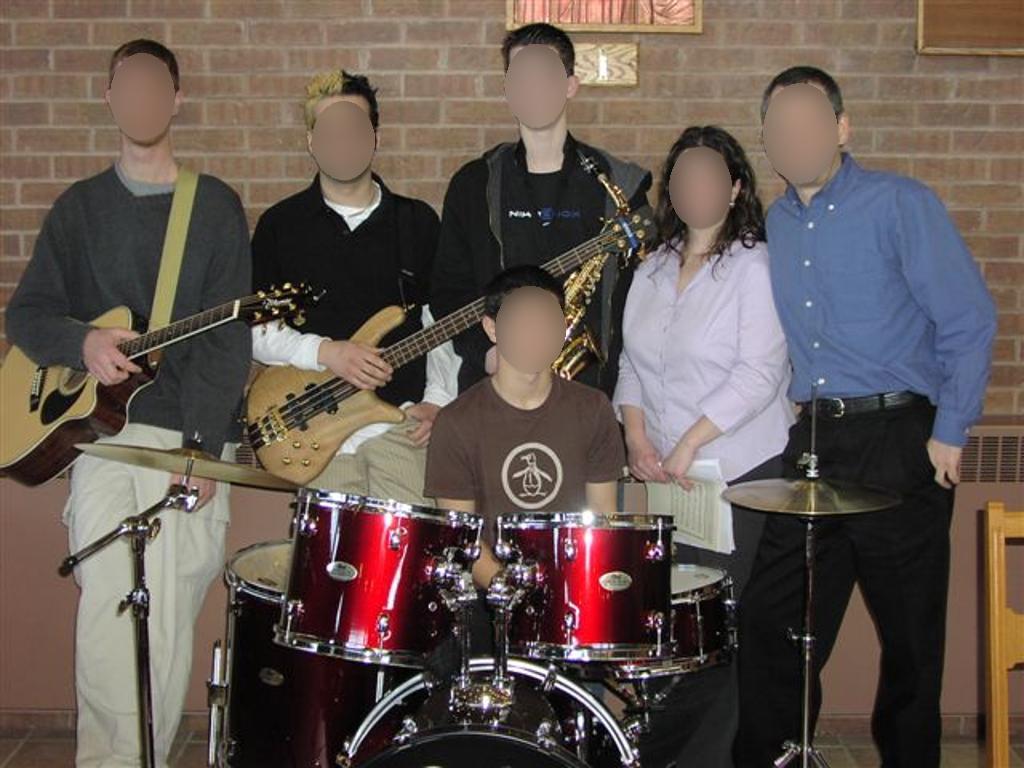} &
    \includegraphics[width=0.20\linewidth]{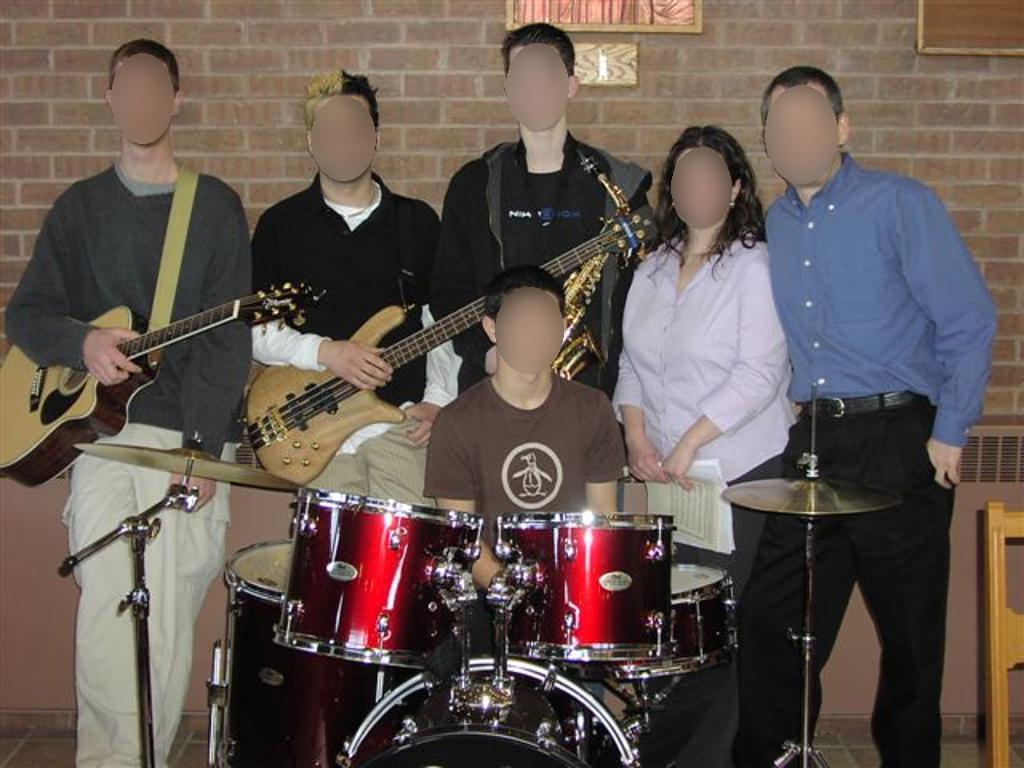} &
    \includegraphics[width=0.20\linewidth]{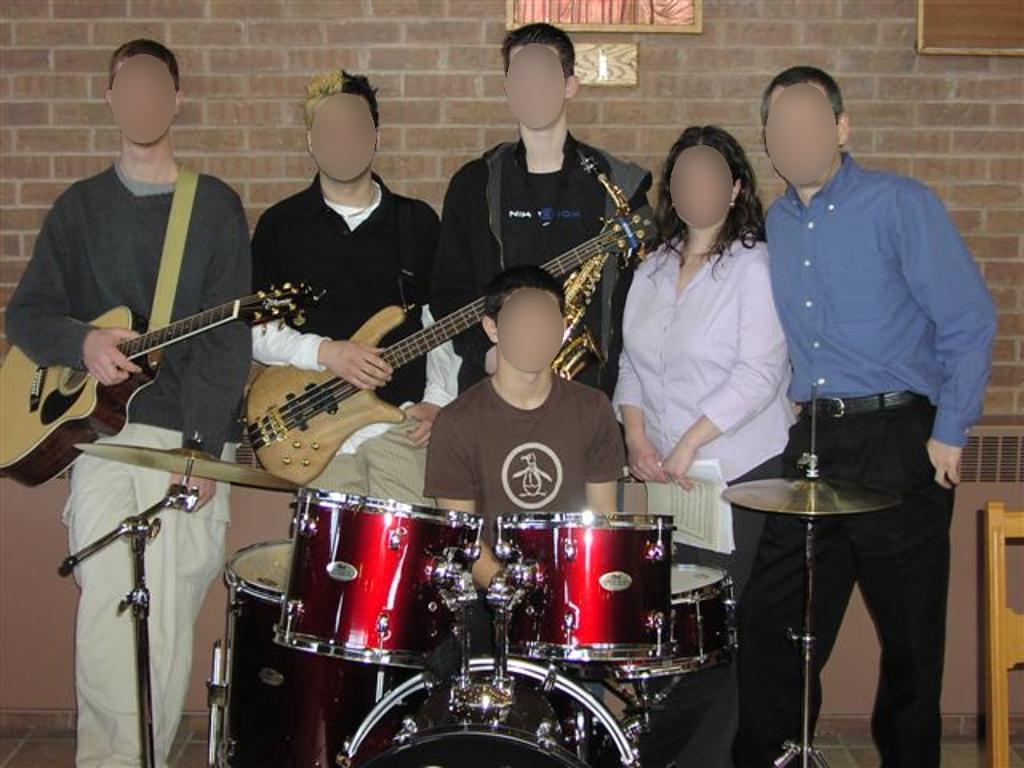} &
    \includegraphics[width=0.20\linewidth]{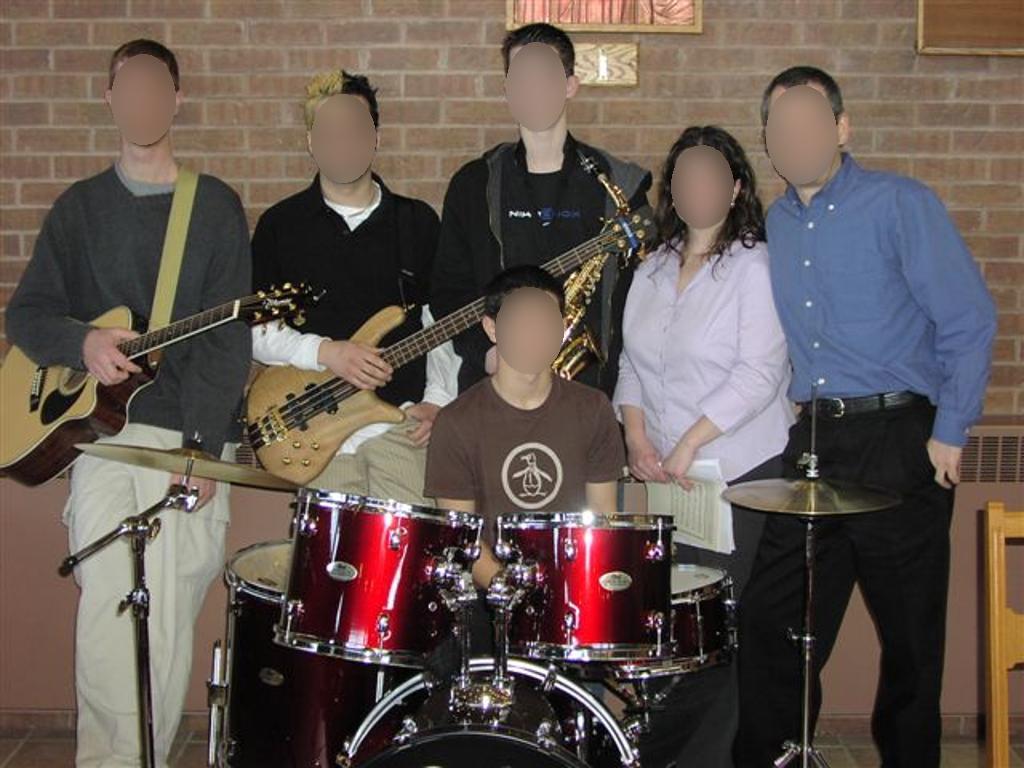}\\

    \includegraphics[width=0.20\linewidth]{figures/test_set/original_images/image_57.jpg} &
    \includegraphics[width=0.20\linewidth]{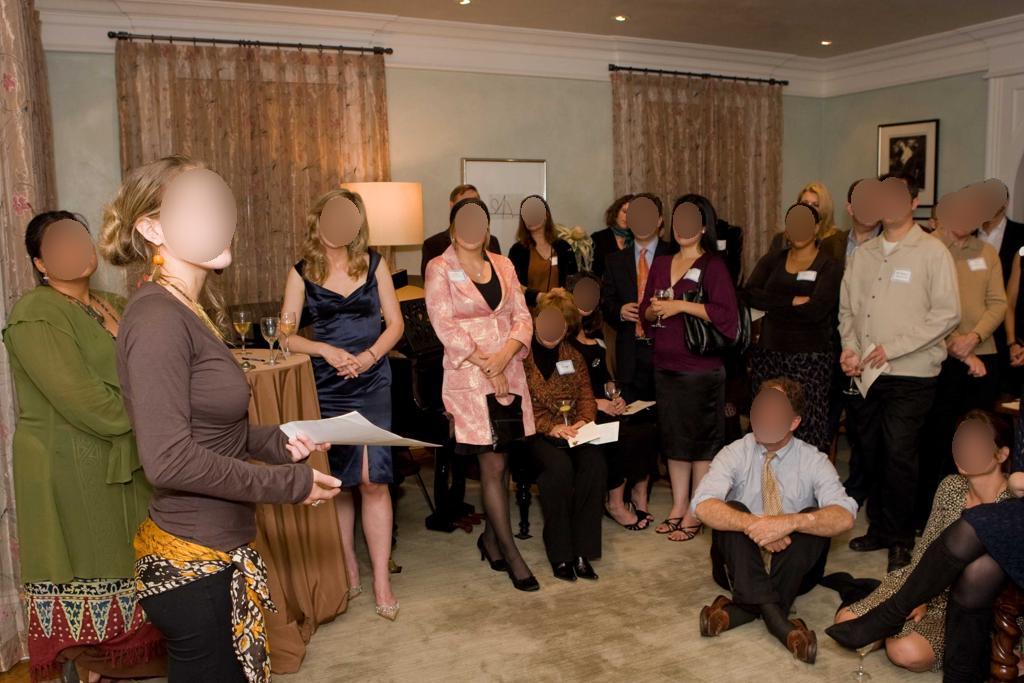} &
    \includegraphics[width=0.20\linewidth]{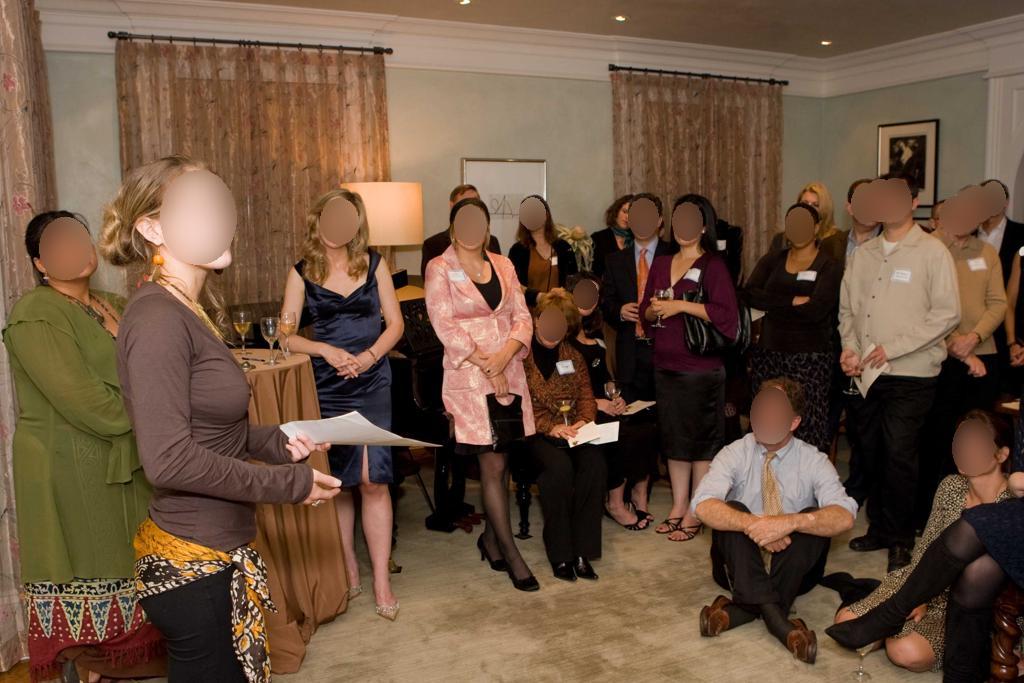} &
    \includegraphics[width=0.20\linewidth]{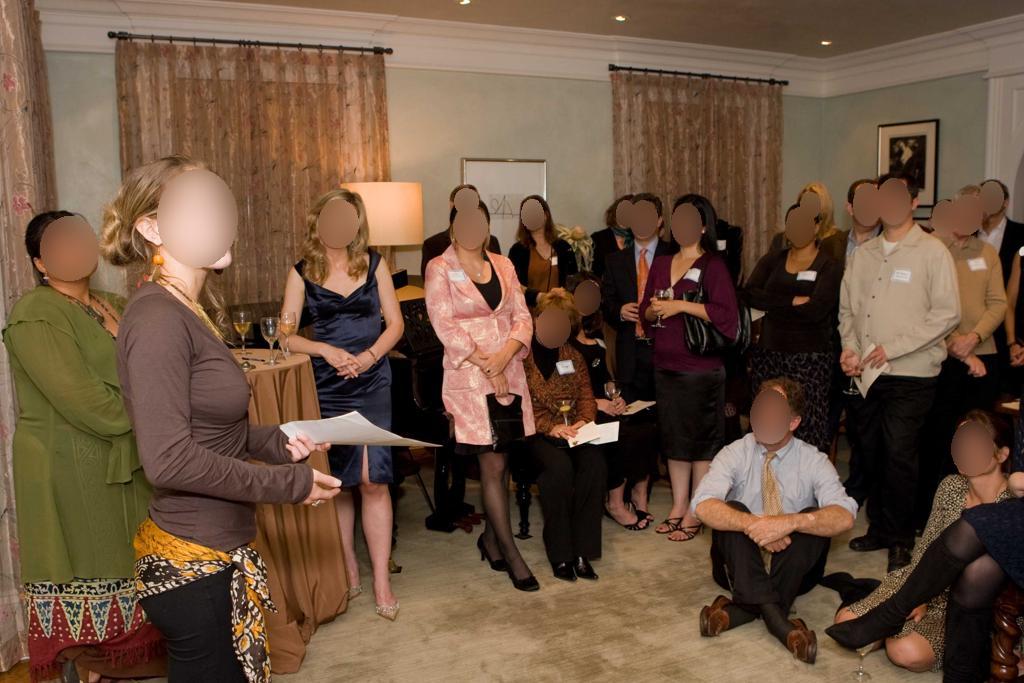} &
    \includegraphics[width=0.20\linewidth]{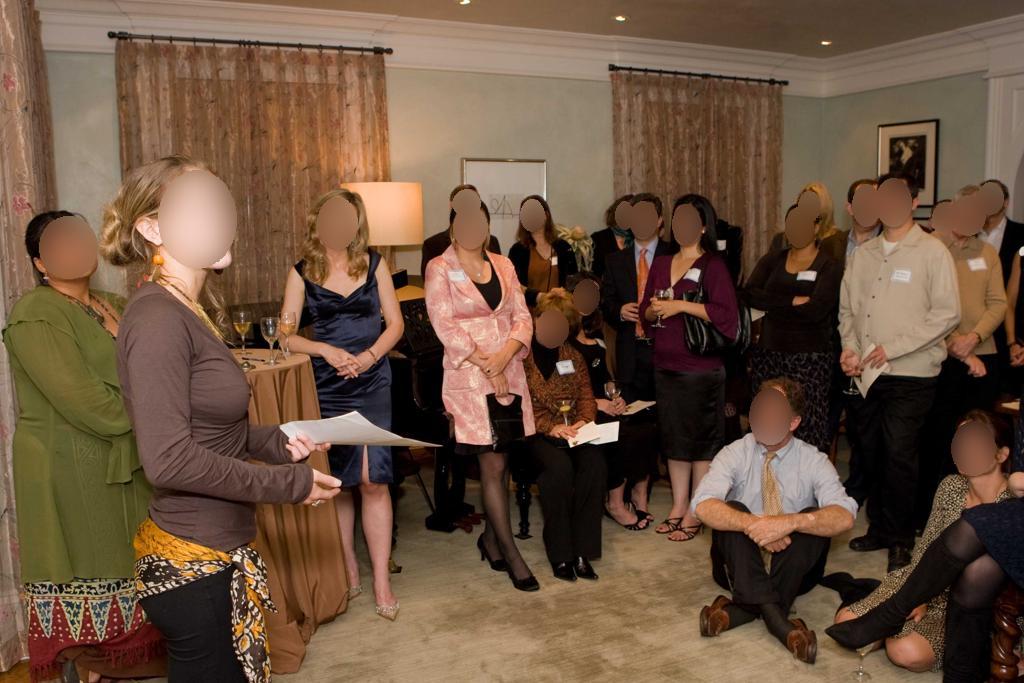}\\

    \includegraphics[width=0.20\linewidth]{figures/test_set/original_images/image_24.jpg} &
    \includegraphics[width=0.20\linewidth]{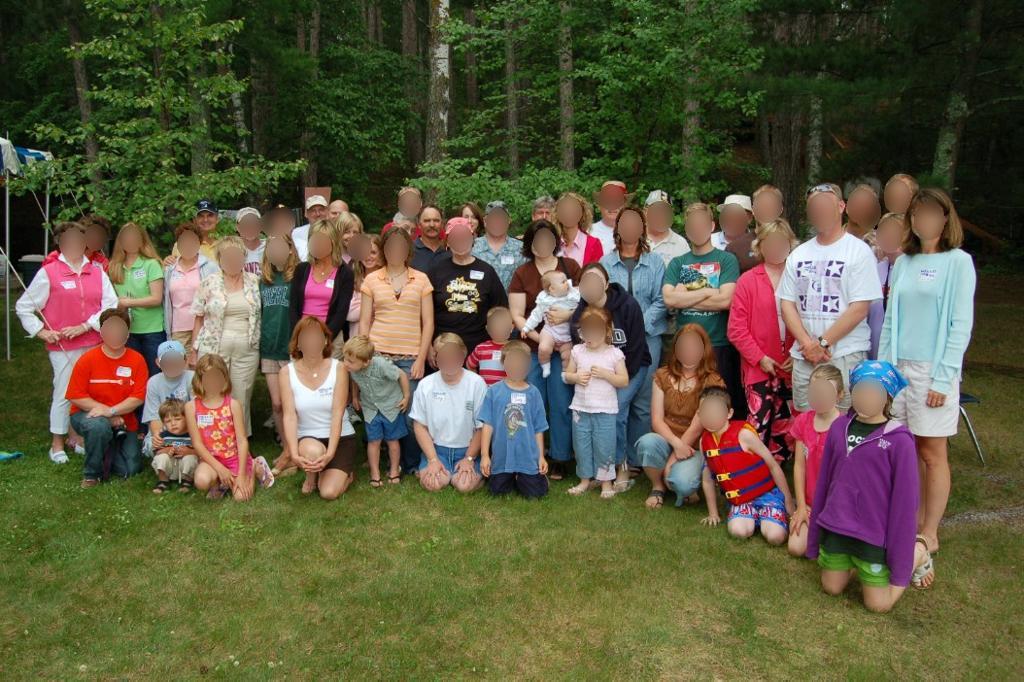} &
    \includegraphics[width=0.20\linewidth]{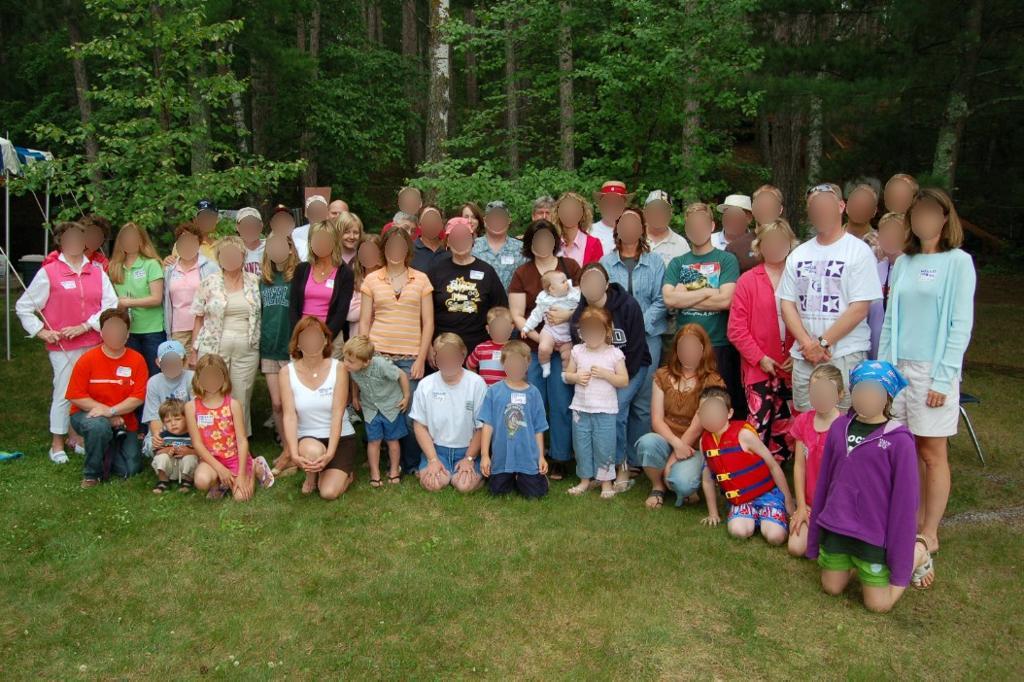} &
    \includegraphics[width=0.20\linewidth]{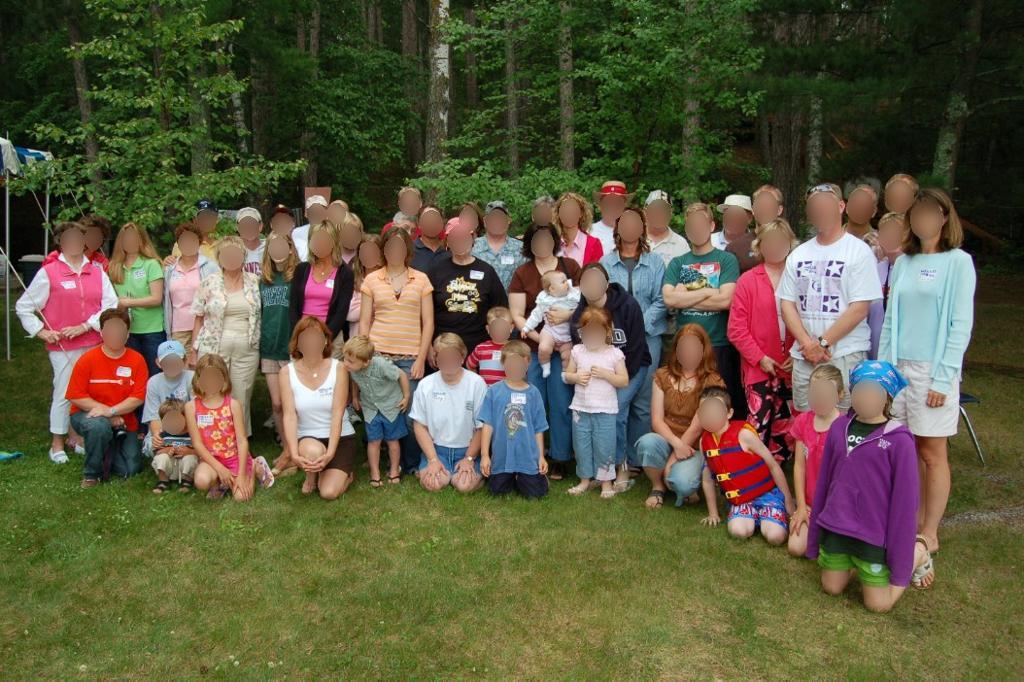} &
    \includegraphics[width=0.20\linewidth]{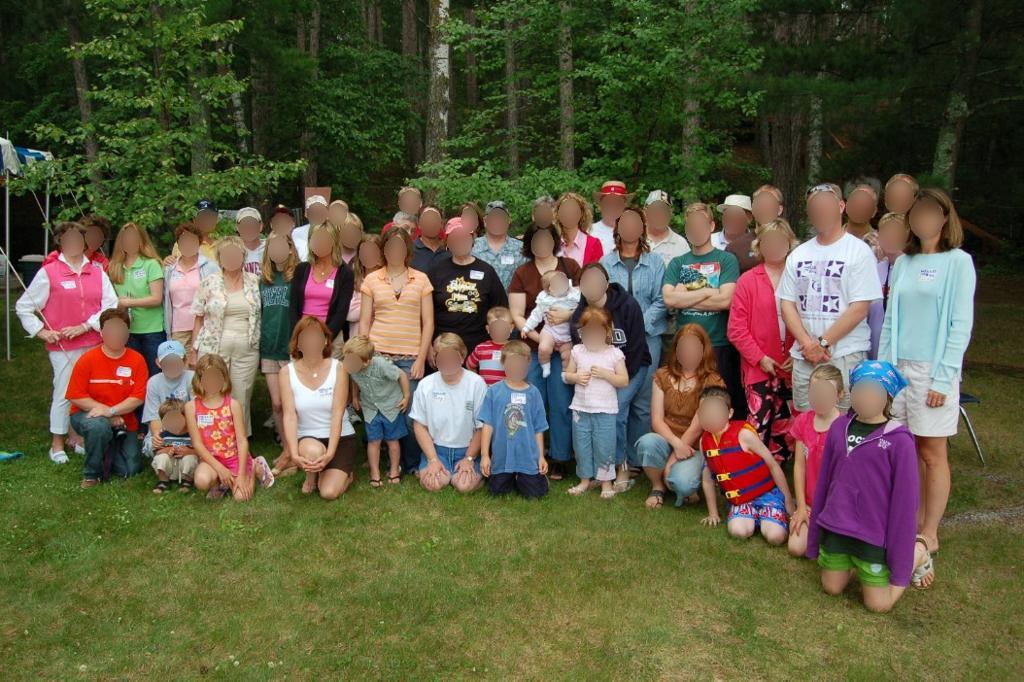}\\
    \end{tabular}
    \caption{Test set results. Original image (left), result for $192\times192$ inference dimension (center left), result for $256\times256$ inference dimension (center), result for $512\times512$ inference dimension (center right), result for original input size (right).}
    \label{fig:yolo_results}
\end{figure}

\noindent
\textbf{Visual evaluation}\\
The YOLO method is very straightforward. We display in Figure~\ref{fig:yolo_results} visual results for several input resizing ($192\times192$, $256\times256$, $512\times512$) as well as for no resizing (keeping original input size).

We conclude that the YOLO-based method is very efficient, faces are very often detected, especially when they are ``big'', meaning that the face covers a great number of pixels. However, and as the Unet before, for small downsampling sizes, YOLO struggles to detect small faces in images 5 and 6.\\
But, the higher we set the input dimension size, the higher number of faces YOLOv5Face detects. When we keep the original dimension, almost no faces are missed except one in the last image. We sum up the results in Table~\ref{tab:yolo}. 

\begin{table}[!h]
  \centering
\resizebox{\linewidth}{!}{
  \begin{tabular}{c|c|c|c|c|c|}
    & \multirow{2}{*}{Number of faces} & \multicolumn{4}{|c|}{Number of faces blurred}\\
    & & $192\times192$ & $256\times256$ & $512\times512$ & Original input size \\
    \hline\hline
    image 1 & 1 & \textbf{1} & \textbf{1} & \textbf{1} & \textbf{1}\\
    \hline
    image 2 & 1 & \textbf{1} & \textbf{1} & \textbf{1} & \textbf{1}\\
    \hline\hline
    image 3 & 5 & \textbf{5} & \textbf{5} & \textbf{5} & \textbf{5} \\
    \hline
    image 4 & 6 & \textbf{6} & \textbf{6} & \textbf{6} & \textbf{6} \\
    \hline\hline
    image 5 & 18 & 16 & 16 & \textbf{18} & \textbf{18}\\
    \hline
    image 6 & 51 & 40 & 43 & 49 & \textbf{50}\\
  \end{tabular}
}
  \caption{Face counting results (the best results for each image is in bold).}
  \label{tab:yolo}
\end{table}

\subsubsection{Computation time}
We conduct three experiments :
\begin{itemize}
    \item One experiment assuming that input images have already been downsampled to the desired input size of the network (either 192 x 192, 256 x 256 or 512 x 512). In this case the downsampling, binary mask extraction and upsampling steps described in Figure 6 do not need to be performed.
    \item One experiment assuming that input images are of size $1024\times1024$. 
    \item One experiment assuming that input images are of size $2048\times2048$.
\end{itemize}.

\begin{table}[htbp]
  \centering
\resizebox{\linewidth}{!}{
  \begin{tabular}{c|c|c|c|c|c}
   & $192\times192$ & $256\times256$ & $512\times512$ & $1024\times1024$ & $2048\times2048$\\  
  \hline\hline
  No resizing needed & $52.5$ & $48.9$ & $26.7$ & $5.7$ & $1.9$ \\
  \hline
  Input size of $1024\times1024$ & $7.2$ & $6.9$ & $6.3$ & $5.7$ & \\
  \hline
  Input size of $2048\times2048$ & $4.4$ & $4.2$ & $2.4$ & $2.0$ & $1.9$
  \end{tabular}
}
  \caption{Number of frames per second processed  by the algorithm in function of the downsampling size.}
  \label{tab:fps_yolo}
\end{table}
What stands out in Table~\ref{tab:fps_yolo} is that the number of frames processed by second by the YOLOv5Face methodology is always higher than with the Unet based method. In fact, we use here the Nano model, which has the smaller number of parameters which speeds up the inference. We observe, as for the Unet-based method, that all the resizing steps greatly deteriorate the algorithm performance. 

\section{Conclusion}\label{sec:conclusions}
In this paper, we have investigated the automatic blurring of faces in images and videos. We have presented two methods, one one-step direct method based on the Unet architecture (DeOldify~\cite{deoldifyIPOL}), and another two-step method which relies on the well-known YOLO object detector \cite{yolo_original, yolo}. 

The YOLO-based method first detects faces which are then blurred using a Gaussian kernel. The Unet-based method directly outputs images in which faces are blurred. We have constructed a dataset of pairs of original and face-blurred images to train this network. The original images come from two popular face-detection datasets (FDDB \cite{fddb} and WIDER \cite{wider}).

The experiments show that both methods are able to correctly blur faces in images and that they are robust to variations in size and pose. 
In terms of computation time, the YOLO-based method is faster, since it benefits from all the optimizations introduced in the YOLO architecture 
to increase its speed. However, the ability of the Unet-based network to blur the faces without detecting them explicitly is
an interesting property that is worth exploring and, in the future, we will investigate how to optimize its speed.

\section*{On Line Demo and Code}
An online demo is available for the interested readers that want to test the performance of both methods in their own videos: \url{https://ipolcore.ipol.im/demo/clientApp/demo.html?id=77777000406}.

Moreover, the code used to obtain the results displayed in the previous sections is available here:
\url{https://github.com/RomanPlaud/script-face-blurring-ipol}. 
The original source codes are borrowed from \url{http://www.ipol.im/pub/art/2022/403/},
\url{https://github.com/jantic/DeOldify} 
and \url{https://github.com/elyha7/yoloface}.

\section*{Acknowledgment}
For the second author the publication is part of the project PID2021-125711OB-I00, financed by MCIN/AEI/10.13039/501100011033/FEDER, EU.

\section*{Image Credits}
{\small
All the original images displayed in the paper come from the  FDDB \cite{fddb} and WIDER \cite{wider} datasets.
}


\bibliographystyle{siam}
\bibliography{article}

\end{document}